\documentclass{article}

\usepackage{arxiv}

\usepackage[utf8]{inputenc} % allow utf-8 input
\usepackage[T1]{fontenc}    % use 8-bit T1 fonts
\usepackage{hyperref}       % hyperlinks
\usepackage{url}            % simple URL typesetting
\usepackage{booktabs}       % professional-quality tables
\usepackage{amsfonts}       % blackboard math symbols
\usepackage{nicefrac}       % compact symbols for 1/2, etc.
\usepackage{microtype}      % microtypography
\usepackage{graphicx}
\usepackage{doi}
\usepackage{subcaption}
\usepackage{amssymb, amsmath, amsthm}
\usepackage{algorithm}
\usepackage{algpseudocode}
\usepackage{appendix}
\usepackage{enumitem}
\usepackage[round]{natbib}

\title{Stochastic Cluster Embedding}

\author{ Zhirong Yang$^{1,2}$\thanks{\texttt{zhirong.yang@ntnu.no}}, Yuwei Chen$^{3}$, Denis Sedov$^{2}$, Samuel Kaski$^{2}$, and Jukka Corander$^{4,5,6}$\\
$^1$Norwegian University of Science and Technology\\
$^2$Aalto University\\
$^3$Finnish Geospatial Research Institute\\
$^4$University of Oslo\\
$^5$Wellcome Sanger Institute\\
$^6$University of Helsinki}

\renewcommand{\shorttitle}{Stochastic Cluster Embedding}

\hypersetup{
	pdftitle={Stochastic Cluster Embedding},
	pdfsubject={},
	pdfauthor={Zhirong Yang, Yuwei Chen, Denis Sedov, Samuel Kaski, and Jukka Corander},
	pdfkeywords={Clustering, Information Divergence, Neighbor Embedding, Nonlinear Dimensionality Reduction, Stochastic Optimization, Visualization},
}

\begin{document}
	\maketitle

\newcommand{\matA}{\mathbf{A}}
\newcommand{\matB}{\mathbf{B}}
\newcommand{\matC}{\mathbf{C}}
\newcommand{\matD}{\mathbf{D}}
\newcommand{\matE}{\mathbf{E}}
\newcommand{\matF}{\mathbf{F}}
\newcommand{\matG}{\mathbf{G}}
\newcommand{\matH}{\mathbf{H}}
\newcommand{\matI}{\mathbf{I}}
\newcommand{\matK}{\mathbf{K}}
\newcommand{\matL}{\mathbf{L}}
\newcommand{\matM}{\mathbf{M}}
\newcommand{\matN}{\mathbf{N}}
\newcommand{\matO}{\mathbf{O}}
\newcommand{\matP}{\mathbf{P}}
\newcommand{\matQ}{\mathbf{Q}}
\newcommand{\matR}{\mathbf{R}}
\newcommand{\matS}{\mathbf{S}}
\newcommand{\matT}{\mathbf{T}}
\newcommand{\matU}{\mathbf{U}}
\newcommand{\matV}{\mathbf{V}}
\newcommand{\matW}{\mathbf{W}}
\newcommand{\matX}{\mathbf{X}}
\newcommand{\matY}{\mathbf{Y}}
\newcommand{\matZ}{\mathbf{Z}}
\newcommand{\matg}{\mathbf{g}}

\newcommand{\calA}{\mathcal{A}}
\newcommand{\calB}{\mathcal{B}}
\newcommand{\calC}{\mathcal{C}}
\newcommand{\calD}{\mathcal{D}}
\newcommand{\calE}{\mathcal{E}}
\newcommand{\calF}{\mathcal{F}}
\newcommand{\calG}{\mathcal{G}}
\newcommand{\calH}{\mathcal{H}}
\newcommand{\calI}{\mathcal{I}}
\newcommand{\calJ}{\mathcal{J}}
\newcommand{\calK}{\mathcal{K}}
\newcommand{\calL}{\mathcal{L}}
\newcommand{\calM}{\mathcal{M}}
\newcommand{\calN}{\mathcal{N}}
\newcommand{\calO}{\mathcal{O}}
\newcommand{\calP}{\mathcal{P}}
\newcommand{\calQ}{\mathcal{Q}}
\newcommand{\calR}{\mathcal{R}}
\newcommand{\calS}{\mathcal{S}}
\newcommand{\calT}{\mathcal{T}}
\newcommand{\calU}{\mathcal{U}}
\newcommand{\calV}{\mathcal{V}}
\newcommand{\calW}{\mathcal{W}}
\newcommand{\calX}{\mathcal{X}}
\newcommand{\calY}{\mathcal{Y}}
\newcommand{\calZ}{\mathcal{Z}}

\newcommand{\bbA}{\mathbb{A}}
\newcommand{\bbB}{\mathbb{B}}
\newcommand{\bbR}{\mathbb{R}}
\newcommand{\bbZ}{\mathbb{Z}}
\newcommand{\bbE}{\mathbb{E}}
\newcommand{\bbH}{\mathbb{H}}

\newcommand{\veca}{\mathbf{a}}
\newcommand{\vecb}{\mathbf{b}}
\newcommand{\vecc}{\mathbf{c}}
\newcommand{\vecd}{\mathbf{d}}
\newcommand{\vece}{\mathbf{e}}
\newcommand{\vecf}{\mathbf{f}}
\newcommand{\vecg}{\mathbf{g}}
\newcommand{\vech}{\mathbf{h}}
\newcommand{\veci}{\mathbf{i}}
\newcommand{\vecj}{\mathbf{j}}
\newcommand{\veck}{\mathbf{k}}
\newcommand{\vecl}{\mathbf{l}}
\newcommand{\vecm}{\mathbf{m}}
\newcommand{\vecn}{\mathbf{n}}
\newcommand{\veco}{\mathbf{o}}
\newcommand{\vecp}{\mathbf{p}}
\newcommand{\vecq}{\mathbf{q}}
\newcommand{\vecr}{\mathbf{r}}
\newcommand{\vecs}{\mathbf{s}}
\newcommand{\vect}{\mathbf{t}}
\newcommand{\vecu}{\mathbf{u}}
\newcommand{\vecv}{\mathbf{v}}
\newcommand{\vecw}{\mathbf{w}}
\newcommand{\vecx}{\mathbf{x}}
\newcommand{\vecy}{\mathbf{y}}
\newcommand{\vecz}{\mathbf{z}}

\newcommand{\vecalpha}{\boldsymbol{\alpha}}
\newcommand{\vecbeta}{\boldsymbol{\beta}}
\newcommand{\veceta}{\boldsymbol{\eta}}
\newcommand{\vectheta}{\boldsymbol{\theta}}
\newcommand{\vecphi}{\boldsymbol{\phi}}
\newcommand{\vecpsi}{\boldsymbol{\psi}}
\newcommand{\vecrho}{\boldsymbol{\rho}}
\newcommand{\vectau}{\boldsymbol{\tau}}
\newcommand{\vecmu}{\boldsymbol{\mu}}
\newcommand{\veceps}{\boldsymbol{\epsilon}}
\newcommand{\vecxi}{\boldsymbol{\xi}}
\newcommand{\vecPhi}{\boldsymbol{\Phi}}
\newcommand{\vecDelta}{\boldsymbol{\Delta}}

\newcommand{\matDelta}{\boldsymbol{\Delta}}
\newcommand{\matEta}{\boldsymbol{\eta}}
\newcommand{\matOmega}{\boldsymbol{\Omega}}
\newcommand{\matPhi}{\boldsymbol{\Phi}}
\newcommand{\matPsi}{\boldsymbol{\Psi}}
\newcommand{\matTheta}{\boldsymbol{\Theta}}
\newcommand{\matLambda}{\boldsymbol{\Lambda}}
\newcommand{\matSigma}{\boldsymbol{\Sigma}}
\newcommand{\matzero}{\mathbf{0}}
\newcommand{\IndexSetI}{\mathcal{I}}
\newcommand{\grad}{\mathcal{\nabla}}

\newcommand{\vecone}{\mathbf{1}}
\newcommand{\veczero}{\mathbf{0}}

\def\maximize{\mathop{{\mathgroup\symoperators maximize}}}
\def\Maximize{\mathop{{\mathgroup\symoperators Maximize}}}
\def\minimize{\mathop{{\mathgroup\symoperators minimize}}}

\def\approach{\mathop{{\mathgroup\symoperators \longrightarrow}}}
\def\defineoperator{\mathop{{\mathgroup\symoperators =}}}
\newcommand{\define}{\defineoperator^{\text{def}}}

\newcommand{\Tr}{\text{Tr}}
\newcommand{\trace}{\text{trace}}
\newcommand{\diag}{\text{diag}}
\newcommand{\gradWJ}{\nabla_{\scriptscriptstyle{\matW}}\calJ}
\newcommand{\const}{\text{constant}}
\newcommand{\fracpartial}[2]{\frac{\partial #1}{\partial  #2}}

\newcommand{\defeq}{\stackrel{\text{def}}{=}}

	\begin{abstract}
Neighbor Embedding (NE) aims to preserve pairwise similarities between data items and has been shown to yield an effective principle for data visualization. However, even the best existing NE methods such as Stochastic Neighbor Embedding (SNE) may leave large-scale patterns hidden, for example clusters, despite strong signals being present in the data. To address this, we propose a new cluster visualization method based on the Neighbor Embedding principle. We first present a family of Neighbor Embedding methods that generalizes SNE by using non-normalized Kullback-Leibler divergence with a scale parameter. In this family, much better cluster visualizations often appear with a parameter value different from the one corresponding to SNE. We also develop an efficient software that employs asynchronous stochastic block coordinate descent to optimize the new family of objective functions. Our experimental results demonstrate that the method consistently and substantially improves the visualization of data clusters compared with the state-of-the-art NE approaches. The code of our method is publicly available\footnote{\url{https://github.com/rozyangno/sce}}.
	\end{abstract}

	% keywords can be removed

\section{Introduction}
The rapid growth in the amount of data processed by analysts demands more efficient information digestion and communication methods. Data visualization by dimensionality reduction facilitates a viewer to digest information in massive data sets quickly. Therefore, it is increasingly applied as a critical component in scientific research, digital libraries, data mining, financial data analysis, market studies, manufacturing production control, drug discovery, etc.
	
Stochastic Neighbor Embedding \citep[SNE;][]{hinton2003sne} is a widely used nonlinear dimensionality reduction (NLDR) method, which approximately preserves the pairwise probabilities of being neighbors (neighboring probabilities for short) in the input space. In particular, the Student t-Distributed Stochastic Neighbor Embedding \citep[t-SNE;][]{vandermaaten2008tsne} has become one of the most popular nonlinear dimensionality reduction methods for data visualization. The t-SNE method employs a heavy-tailed distribution for the neighboring probabilities in the embedding and minimizes their Kullback-Leibler divergence against precomputed input probabilities.
	
Discovery of large-scale patterns such as clusters is an important task of NLDR. It is often believed that t-SNE can show clusters for well clusterable data, with a smaller Kullback-Leibler divergence corresponding to a better quality. However, recently we found many counter-examples where t-SNE may not correctly visualize the clusters even if the input neighborhoods are well clusterable (see Section \ref{sec:exps} for details). Tuning hyperparameters in t-SNE does not help in solving this issue because a better t-SNE learning objective can correspond to a worse cluster embedding. The root cause of the failure is that t-SNE was designed to preserve the neighborhood or local information, which is not necessarily directly connected with finding large-scale patterns such as clusters.
	
To address the problem, we propose a new method called Stochastic Cluster Embedding (SCE) for better cluster visualizations. We generalize SNE using a family of I-divergences parameterized by a scaling factor $s$, between the non-normalized similarities in the input and output space. We show that SNE is a special case in the family with $s$ chosen to be the normalizing factor of output similarities. However, through a user study, we find that the best $s$ value for cluster visualization is often different from the one chosen by SNE.
	
To overcome the t-SNE drawback, SCE employs another choice that mixes the input similarities in calculating $s$. The scale factor is adaptively adjusted when optimizing the new learning objective, and the data points are thus better grouped. We have also developed an efficient optimization algorithm that employs asynchronous stochastic block coordinate descent. The new algorithm can utilize parallel computing devices such as CPUs or GPUs and is suitable for mega-scale problems with massive data items.

Our new method is tested on six real-world data sets and compared with the state-of-the-art nonlinear dimensionality reduction approaches, including t-SNE \citep{vandermaaten2008tsne}. The results show that our method consistently performs better than t-SNE for cluster visualization. In all tested cases where t-SNE fails, SCE can show the clusters clearly and accurately. Among all compared methods, our new method is the only one that gives good cluster visualizations for all tested data sets.
	
The remainder of the paper is organized as follows. The popular visualization method SNE is briefly reviewed in Section \ref{sec:tsne}. In Section \ref{sec:cluvis} we give our notations about clustering and cluster visualization. In Section \ref{sec:ne4cluvis}, we present the new method based on Neighbor Embedding for cluster visualization, including its learning objective function and optimization algorithms. We present the experiment settings and results in Section \ref{sec:exps} and conclude the paper by presenting possible directions for future research in Section \ref{sec:conclusions}. 

\section{Stochastic Neighbor Embedding}
\label{sec:tsne}
Neighbor Embedding (NE) is a collection of methods that seek vectorial representation of data items according to their neighborhoods or similarities in the original space \citep{yang2013scalable,yang2014optequiv}. Given a set of multivariate data points $\{x_1,x_2,\dots,x_N\}$, where $x_i\in\bbR^D$, their pairwise similarities are encoded in a nonnegative square matrix $p$. Neighbor Embedding finds a mapping $x_i\mapsto y_i\in\bbR^d$ for $i=1,\dots,N$ ($d=2$ or $d=3$ for visualization) such that the similarities in the mapped space, encoded by $q_{ij}$, approximate those in $p$. The approximation is realized by minimizing a certain divergence between $p$ and $q$. For abbreviation we also write $Y=[y_1,\dots,y_N]$.

SNE is a family of Neigbhor Embedding methods, which minimizes the Kullback-Leibler divergence $D_\text{KL}(P||Q)$ between the normalized input and output similarities $P$ and $Q$. The matrix $P$ can be precomputed, for example, $P_{ij}=(P_{i|j}+P_{j|i})/(2N)$ with $P_{i|j}=p_{ij}/\sum_lp_{il}$. In this paper we focus on the matrix-wise normalization\footnote{In this paper the matrixwise summation is over off-diagonal elements, i.e.~$\sum_{ij}A_{ij}\defeq\sum_{ij:i\neq j}A_{ij}$.} of $Q$ using $Q_{ij}=q_{ij}/\sum_{ab}q_{ab}$. In SNE, the output similarities are usually given by a Gaussian kernel $q_{ij}=\exp\left(-\|y_i-y_j\|^2\right)$ or a Student $t$-kernel $q_{ij}=(1+\|y_i-y_j\|^2)^{-1}$. SNE equipped with the latter kernel is called Student t-distributed Stochastic Neighbor Embedding \citep[t-SNE;][]{vandermaaten2008tsne}.

\section{Cluster Visualization}
\label{sec:cluvis}
We focus on Neighbor Embedding for cluster visualization. 
A clustering divides the data objects into a number of groups, such that objects in the same group (called a cluster) are more similar to each other than to those in other groups (clusters) \citep{tandatamining}. The pairwise similarities can be encoded in a similarity matrix $p$, defined as in NE.

A similarity matrix is considered to be (well) clusterable if there is a clustering such that high similarities appear (much) more probably within clusters than between clusters. If we sort the rows and columns in such a similarity matrix according to the cluster labels, we should observe a diagonal blockwise pattern, where high similarities are denser within the cluster blocks. The pattern is much clearer for a well clusterable similarity matrix.

A good cluster visualization is a display where the user can easily see the groups of data points. In scatter plots, there should be clear space separating the groups such that points in the same group are closer to each other than to those in other groups. Besides sufficient separation, a good cluster visualization should also respect the within-cluster information by approximating the corresponding neighborhoods.

It is often believed that t-SNE can achieve a good cluster visualization \citep{vandermaaten2008tsne} and some successful examples have been presented in the literature. However, as any data analysis method, t-SNE may also produce false-negative results for challenging data sets. That is, t-SNE can result in no apparent clustering for some $p$ matrices, despite them being well clusterable. Such errors are demonstrated with several examples in Section \ref{sec:exps}. The problem cannot be solved by using better optimization algorithms because minimizing the t-SNE objective aims at a better matching of local information and this may not correspond to a better large-scale pattern. A smaller t-SNE cost can in fact correspond to a worse cluster visualization.

\section{Neighbor Embedding for Cluster Visualization}
\label{sec:ne4cluvis}
In this section, we show in closer detail why t-SNE can fail despite a $p$ matrix being well clusterable and how to correct the problem to obtain good cluster visualizations.

\subsection{Generalized Stochastic Neighbor Embedding}
We begin with a variant of Kullback-Leibler divergence \citep{amari1985differential} which is defined over non-normalized output similarities with a parameter $s>0$:
\begin{align}
\label{eq:idiv}
D_\text{I}(P||sq)=\sum_{ij}\left[P_{ij}\ln\frac{P_{ij}}{sq_{ij}}-P_{ij}+sq_{ij}\right].
\end{align}
The divergence is called non-normalized KL-divergence or I-divergence, which measures the discrepancy between $P$ and $q$ at a certain scale $s$. In this work we focus on the student $t$-kernel $q_{ij}=(1+\|y_i-y_j\|^2)^{-1}$.

We call the NE variant that minimizes $D_\text{I}(P||sq)$ Generalized Stochastic Neighbor Embedding (GSNE) as it generalizes SNE. To see this, we eliminate $s$ by setting $\partial D_\text{I}(P||sq)/\partial s=0$, which gives \citep[see][ or supplemental document Section 1]{yang2014optequiv}
\begin{align}
\label{eq:optequiv}
\arg\min_YD_\text{KL}(P||Q)=\arg\min_Y\min_{s>0}D_\text{I}(P||sq),
\end{align}
where the equality holds when
\begin{align}
\label{eq:snechoice}
s=\frac{1}{\sum_{ij}q_{ij}}.
\end{align}
That is, SNE is a special case of GSNE with $s$ set to the normalizing factor of $q$.

The SNE choice of $s$ in Eq.~\ref{eq:snechoice} is not necessarily an optimal choice for cluster visualization. It aims at matching two neighboring probability matrices. Such a locality preserving objective can prevent the discovery of large-scale patterns, such as clusters present among the input items. Fortunately, below we show that using another choice of $s$ can correct the problem.

\subsection{Selecting $s$ for better cluster visualization}
The GSNE objective can be rewritten as
\begin{align}
\label{eq:idivex}
D_\text{I}(P||sq)=\calJ_\text{attraction} + \calJ_\text{repulsion} + C
\end{align}
where  $\calJ_\text{attraction}=-\sum_{ij}P_{ij}\ln q_{ij}$ and $\calJ_\text{repulsion}=s\sum_{ij}q_{ij}$  respectively correspond to attractive and repulsive forces, and $C=-\ln s-1 +\sum_{ij}P_{ij}\ln P_{ij}$ is a constant w.r.t. $q$ for a given $s$. Here the scale parameter $s$ also controls the tradeoff between the two forces.

It is known that increasing attraction, e.g., by replacing $P$ with $\beta P$ ($\beta>1$) can encourage the mapped data points to form tighter clumps, with more empty space in the visualization \citep[see e.g.,][]{vandermaaten2008tsne,vandermaaten2014bhtsne,optsne}. The trick is called ``early exaggeration'' and has been used in t-SNE initialization, where $\beta=4$ by \citet{vandermaaten2008tsne} or $\beta=12$ by \citet{vandermaaten2014bhtsne} in the first 250 iterations\footnote{Other t-SNE optimization methods \citep[e.g.][]{optsne} may use a different number of iterations spent in early exaggeration}. The ``early exaggeration'' trick is equivalent to setting
\begin{align}
s = \frac{1}{\beta\sum_{ij}q_{ij}}
\end{align}
during the initialization. Note that after the initialization, t-SNE still uses $\beta=1$, i.e. the original choice in Eq.~\ref{eq:snechoice}. As we will see in Section \ref{sec:exps} and the supplemental document Section 4, the ``early-exaggeration'' still cannot produce good cluster visualizations, even if $\beta>1$ is used throughout the iterations (see the supplemental document Section 3). 

In this work, we propose to choose
\begin{align}
\label{eq:scechoice}
s = \frac{1}{\sum_{ij}w_{ij}q_{ij}},
\end{align}
where $w_{ij}=\alpha N(N-1)P_{ij} + (1-\alpha)$ with $\alpha\in[0,1]$ for better cluster visualizations. When $\alpha=0$, it reduces to the SNE choice. When $\alpha>0$, it can adaptively bring extra repulsion to improve cluster visualization. To see this, we rewrite the denominator in Eq.~\ref{eq:scechoice}:
\begin{align}
\sum_{ij}w_{ij}q_{ij} = N(N-1)\left[\alpha\bbE_{(i,j)\sim\text{Categorical}(P)}\{q_{ij}\} + (1-\alpha)\bbE_{(i,j)\sim \text{Uniform}\left([1,\dots,N]^2\right)}\{q_{ij}\}\right].
\end{align}
In the brackets, there are two expectations over different distributions. Because $Y$ is random initially, the two expectations do not differ much, and overall the optimization is similar to the original SNE. After minimizing the discrepancy $D_I(P||sq)$ for a while, data pairs that correspond to large $P_{ij}$'s will be mapped closer. As a result, $\bbE_{(i,j)\sim\text{Categorical}(P)}\{q_{ij}\}$ becomes larger than $\bbE_{(i,j)\sim \text{Uniform}\left([1,\dots,N]^2\right)}\{q_{ij}\}$, which leads to a smaller $s$ and thus smaller repulsion than in SNE. In this work, we simply set $\alpha=0.5$, and we find it already suffices for many data sets. Because our method often produces clearer cluster visualizations, we have named it \emph{Stochastic Cluster Embedding} (SCE).

\subsection{Optimization}
The GSNE or SCE objective function is smooth over $Y$. Therefore it can be optimized by any existing unconstrained smooth optimization techniques such as gradient descent with momentum in t-SNE. However, the original t-SNE algorithm runs in a centralized manner and is thus slow for large-scale data sets. Moreover, the tree-based approximation \citep{yang2013scalable,vandermaaten2014bhtsne,vladymyrov2014fmm} to the objective function and gradient calculation requires rather complex programming \citep[see e.g.,][]{chan2019gpu}.

Here we develop a simple tree-free parallel algorithm to optimize SCE. It repeats the following steps until maximum iterations or the $Y$ change is smaller than a given tolerance:
\begin{enumerate}
	\item update $Y$ given the current $s$;
	\label{step:update_Y}
	\item estimate $s$ given the current $Y$,
	\label{step:estimate_s}
\end{enumerate}
Because we usually initialize $Y$ with small numbers around zero, all $q_{ij}$'s are close to 1 at the beginning. Therefore it is reasonable to initialize $s=N(N-1)$.

In both steps, the computation is distributed to a number of computing units called workers. In Step \ref{step:update_Y}, we first rewrite $D_I(P||sq$ in a stochastic form:
\begin{align}
\calJ_\text{attraction}&= \bbE_{(i,j)\sim\text{Categorical}(P)}\left\{\calJ_\text{attraction}^{(i,j)}\right\},\\
\calJ_\text{repulsion}&=\bbE_{(i,j)\sim\text{Uniform}\left([1,\dots,N]^2\right)}\left\{\calJ_\text{repulsion}^{(i,j)}\right\},
\end{align}
where $\calJ_\text{attraction}^{i,j}=-\ln q_{ij}$ and $\calJ_\text{repulsion}^{(i,j)}=sN(N-1)q_{ij}$.
According to this form, each worker randomly draws a pair $(i,j)$ for attraction and another pair for repulsion, calculates their partial stochastic gradients w.r.t. $y_i$ and $y_j$:
\begin{align}
\fracpartial{\calJ_\text{attraction}^{(i,j)}}{y_i} &= -\fracpartial{\calJ_\text{attraction}^{(i,j)}}{y_j} = -2q_{ij}(y_i-y_j),\\
\fracpartial{\calJ_\text{repulsion}^{(i,j)}}{y_i} &= -\fracpartial{\calJ_\text{repulsion}^{(i,j)}}{y_j} = 2sN(N-1)q_{ij}^2(y_i-y_j),
\end{align}
and updates the corresponding mapped points by stochastic partial gradient descent. In this way, each worker requires only $O(1)$ cost for updating a pair of mapped points $y_i$ and $y_j$.

Next we consider how to estimate $s$ in an asynchronously stochastic and distributed manner. The $(i,j)$ samples and the corresponding $q_{ij}$'s in the denominator of Eq.~\ref{eq:scechoice}, that is $s^{-1}$), have already been obtained from Step \ref{step:update_Y}. 
Denote by $\xi$ and $\omega$, respectively, the weighted sum and count of newly calculated $q_{ij}$'s.
We get $\displaystyle\frac{N(N-1)\xi}{\omega}$ as a stochastic approximation of $s^{-1}$ and mix it with the current estimate as the new one:
\begin{align}
s^{-1}\leftarrow\rho s^{-1}+(1-\rho)\frac{N(N-1)\xi}{\omega},
\end{align} 
where $\rho\in(0,1)$ is the forgetting rate. We find $\rho=\frac{N(N-1)}{N(N-1)+\omega}$ working well in practice.

The pseudo-code of the SCE algorithm is given in Algorithm \ref{alg:sce}. Because the algorithm is almost lock-free, it is straightforward to implement it efficiently on multi-core CPUs and GPUs. Our algorithm belongs to the family of stochastic block coordinate descent optimization, for which \citet{sbcdcomplexity} gave the convergence guarantee and convergence rate.

\algnewcommand\algorithmicparforeach{\textbf{parallel for each}}
\algdef{S}[FOR]{ParForEach}[1]{\algorithmicparforeach\ #1\ \algorithmicdo}
\algnewcommand{\LineComment}[1]{\noindent \(\triangleright\) #1}
\begin{algorithm}[t]
	\caption{Stochastic Cluster Embedding}
	\label{alg:sce}
	\begin{algorithmic}[1]
		\State {\bfseries Input:} similarity matrix $P$, number of iterations $T$.
		\State Initialize $Y=\{y_i\}_{i=1}^N$ with small numbers, e.g.~$y_{id}\sim\calN(0,10^{-4})$ for all $i,d$.
		\State Initialize $s^{-1}\leftarrow N(N-1)$
		\For{t=0\text{ to }T}
		\State $\eta_t\leftarrow1-t/T$
		\State $\xi\leftarrow0$ %\Comment sum of calculated $q_{ij}$
		\State $\omega\leftarrow0$ %\Comment count of calculated $q_{ij}$
		%		\ParForEach{worker (with reduction on $\xi$ and $\omega$)} %\Comment{update $Y$}
		\ParForEach{worker} %\Comment{update $Y$}
		\State Draw $(i,j)\sim$Categorical($P$) \Comment{handle attraction}
		\State $q_{ij}\leftarrow (1+\|y_i-y_j\|^2)^{-1}$
		\State $\grad\leftarrow\displaystyle-2q_{ij}(y_i-y_j)$
		\State $y_i\leftarrow y_i + \eta_t \grad$~~~~~~$y_j\leftarrow y_j - \eta_t \grad$
		\State $\xi\leftarrow \xi + \alpha q_{ij}$~~~~~~ $\omega\leftarrow \omega + \alpha$
		
		\State Draw $(i,j)\sim$Uniform($[1,\dots,N]^2$) \Comment{handle repulsion}
		\State $q_{ij}\leftarrow (1+\|y_i-y_j\|^2)^{-1}$
		\State $\grad\leftarrow\displaystyle2sN(N-1)q_{ij}^2(y_i-y_j)$
		\State $y_i\leftarrow y_i + \eta_t \grad$~~~~~~$y_j\leftarrow y_j - \eta_t \grad$
		\State $\xi\leftarrow \xi + (1-\alpha)q_{ij}$~~~~~~ $\omega\leftarrow \omega + (1-\alpha)$
		\EndFor
		\State $\rho=\frac{N(N-1)}{N(N-1)+\omega}$
		\State $\displaystyle s^{-1}\leftarrow\rho s^{-1}+(1-\rho)\frac{N(N-1)\xi}{\omega}$;  %\Comment update $s$
		\EndFor
		
		\State {\bfseries Output:} low-dimensional representations $Y$
	\end{algorithmic}
\end{algorithm}

\section{Experiments}
\label{sec:exps}
We have compared our method with t-SNE, as well as two other more recent methods called LargeVis and UMAP:
\begin{itemize}
	\item t-SNE: We have used the implementation\footnote{available at \url{https://github.com/lvdmaaten/bhtsne}} by \citet{vandermaaten2014bhtsne}, where the maximum iterations in t-SNE was set to 10000 (ten times as the default) to get closer to convergence.
	
	\item LargeVis \citep{tang2016largevis}: we have used its official implementation in GitHub\footnote{available at \url{https://github.com/lferry007/LargeVis}}.
	\item UMAP \citep[Uniform Manifold Approximation and Projection;][]{umap}: we have used the \texttt{umap-learn} package in Python.
\end{itemize}

We have used six real-world data sets in our experiments:
\begin{itemize}
	\item \texttt{SHUTTLE}: the Statlog (Shuttle) Data Set in the UCI repository\footnote{available at \url{https://archive.ics.uci.edu/ml/datasets/Statlog+(Shuttle)}}. There are 58000 samples of 9 dimensions in three large and four small classes.
	
	\item \texttt{MNIST}: the MNIST database of handwritten digits\footnote{available at \url{http://yann.lecun.com/exdb/mnist/}}. There are 70,000 samples of 784 dimensions in ten digit classes.
	
	\item \texttt{IJCNN}: the IJCNN 2001 neural network competition data\footnote{available at \url{https://www.csie.ntu.edu.tw/~cjlin/libsvmtools/datasets/binary.html}}. There are 126,701 samples of 22 dimensions and from ten engines (classes).
	
	\item \texttt{TOMORADAR}: The data was collected via a helicopter-borne microwave profiling radar \citep{chen2017uav} termed FGI-Tomoradar to investigate the vertical topography structure of forests.  After preprocessing, the data set contains 120,024 samples of 8192 dimensions from three classes.
	
	\item \texttt{FLOW-CYTOMETRY}: the single-cell biology data set collected from Flow Repository\footnote{available at \url{https://flowrepository.org/id/FR-FCM-ZZ36}}. After preprocessing, the data set contains 1,000,000 samples of 17 dimensions.
	
	\item \texttt{HIGGS}: the HIGGS Data Set in the UCI repository\footnote{available at \url{https://archive.ics.uci.edu/ml/datasets/HIGGS}}. The data was produced using Monte Carlo simulations of the particles in a physics experiment. There are 11,000,000 data points of 28 dimensions. Previously the data were used for classification between the bosons and the background particles, whereas there is little research on unsupervised learning on the data. Here we compared visualizations to discover the particle clusters.
\end{itemize}

To our knowledge, it is a largely unsolved problem how to convert vectorial data to a similarity matrix optimally, and there is no universally best solution. In practice, popular choices are Entropic Affinity \citep[EA;][]{hinton2003sne} and $k$-Nearest Neighbor ($k$-NN) with tuning of the perplexity (or $k$) parameter.

This work focuses on cluster embedding of a given similarity matrix $P$. We have constructed the $P$ matrix by using EA with perplexity 30 for \texttt{SHUTTLE} and \texttt{IJCNN}. We have used symmetrized $k$-NN graph adjacency matrix as $P$ for \texttt{MNIST}, \texttt{TOMORADAR}, \texttt{FLOW-CYTOMETRY} and \texttt{HIGGS} with $k=10$, $k=50$, $k=15$ and $k=5$, respectively. In this way, the constructed similarity matrices are well clusterable because they comprise diagonal blockwise pattern, as we shall see in Figure \ref{fig:blockvis}. A good cluster visualization method should be able to clearly show these clusters.

We have performed three groups of empirical studies to verify the advantages of the proposed SCE method. Below we first demonstrate its visualizations compared with those by t-SNE, LargeVis, and UMAP. Second, we compare the $s$ choices in SCE and t-SNE to our user study. Third, we verify the clustering quality of the compared methods by seeing how well they can group the input similarities.

\subsection{Visualization comparison}
\label{sec:viscomp}
The SCE visualizations compared with the other methods are shown in Figures \ref{fig:vis_shuttle} to \ref{fig:vis_higgs}. It is important to notice that cluster visualizations are unsupervised. Therefore, we should focus on the colorless scatter plots and check whether large-scale patterns (i.e.~clusters) appear or not. The colors in the sub-figures are only used for reference to verify the alignment between clusters and ground truth classes.

We can see that SCE works well for all six data sets. Compared with other approaches, SCE shows several typically compact clusters with clear separation among them. Therefore it is easy to see the major clusters in all even if color is removed (i.e.~without supervised labels).

t-SNE fails for five of the six test data sets. For \texttt{SHUTTLE}, \texttt{IJCNN}, \texttt{TOMORADAR}, the t-SNE layouts overall look like a single diamond with too many small groups, where no major clusters can be identified. For \texttt{FLOW-CYTOMETRY} the t-SNE visualization is nearly a single ball, while for \texttt{HIGGS} it is just a hairball. The only barely successful case is \texttt{MNIST}, but there are still many data points that scatter between the clusters, leaving the boundaries unclear.

LargeVis is slightly better than t-SNE. It correctly shows ten clusters for \texttt{MNIST}. In the LargeVis visualizations of \texttt{SHUTTLE}, \texttt{IJCNN}, \texttt{TOMORADAR} and \texttt{FLOW-CYTOMETRY}, we can also see some groups of data points. However, there are still too many small groups, and it is hard to identify the major clusters. LargeVis fails for \texttt{HIGGS}, where no clear cluster is shown.

UMAP works better than t-SNE and LargeVis for some data sets. It also correctly shows ten clusters for \texttt{MNIST}. UMAP can separate the ten engine clusters for \texttt{IJCNN}. We can barely see several major clusters in its \texttt{FLOW-CYTOMETRY} visualization. The method also successfully identifies several clusters for \texttt{HIGGS}. However, UMAP does not work well for \texttt{SHUTTLE} because there are many small groups without clear separation of major clusters. UMAP fails for \texttt{TOMORADAR}, where no clustering pattern is found.

\newcommand{\comparedfigwidth}{7cm}
\begin{figure}[p]
	\begin{center}
		\subcaptionbox{SCE (69 seconds)}
		{\includegraphics[width=\comparedfigwidth]{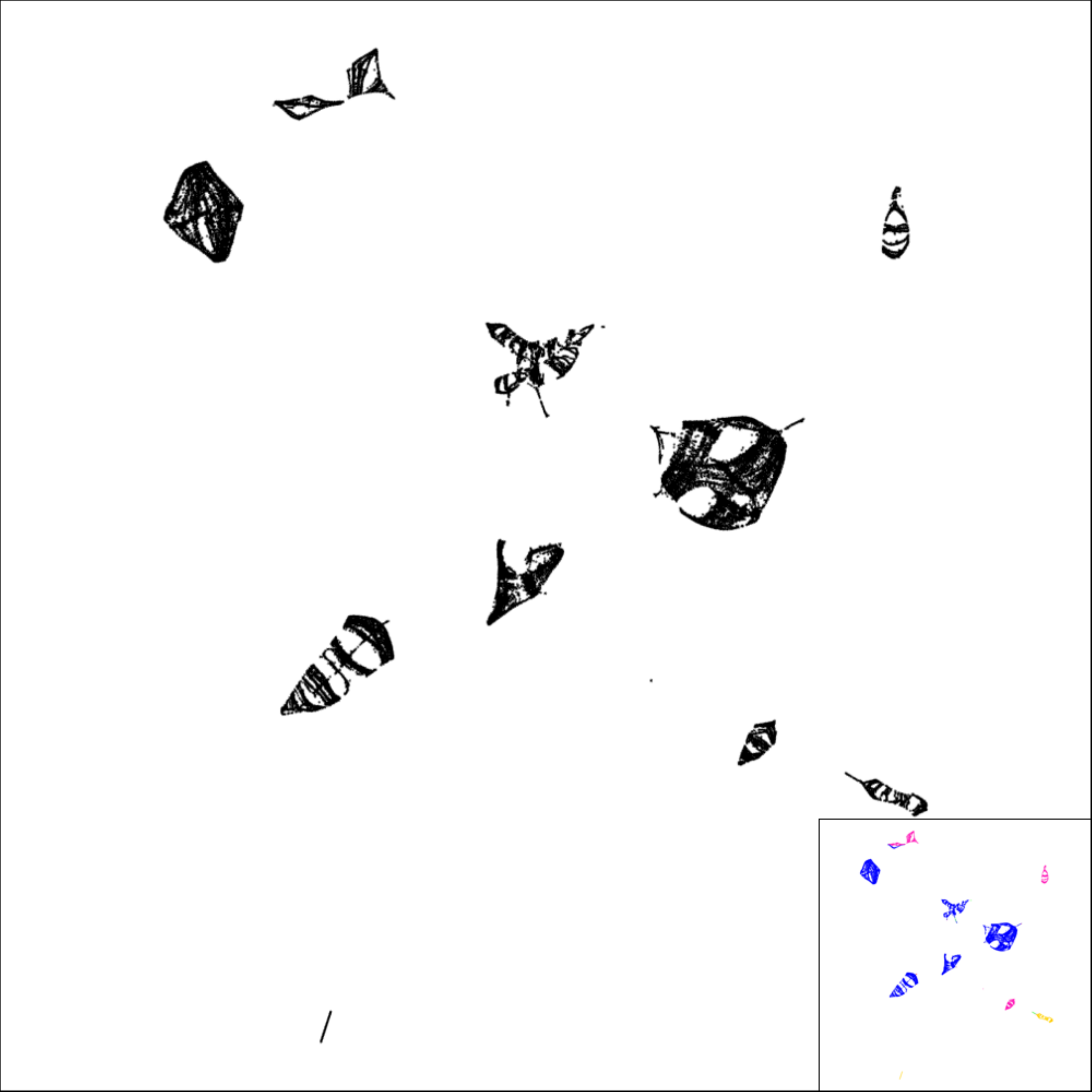}}
		\subcaptionbox{t-SNE}
		{\includegraphics[width=\comparedfigwidth]{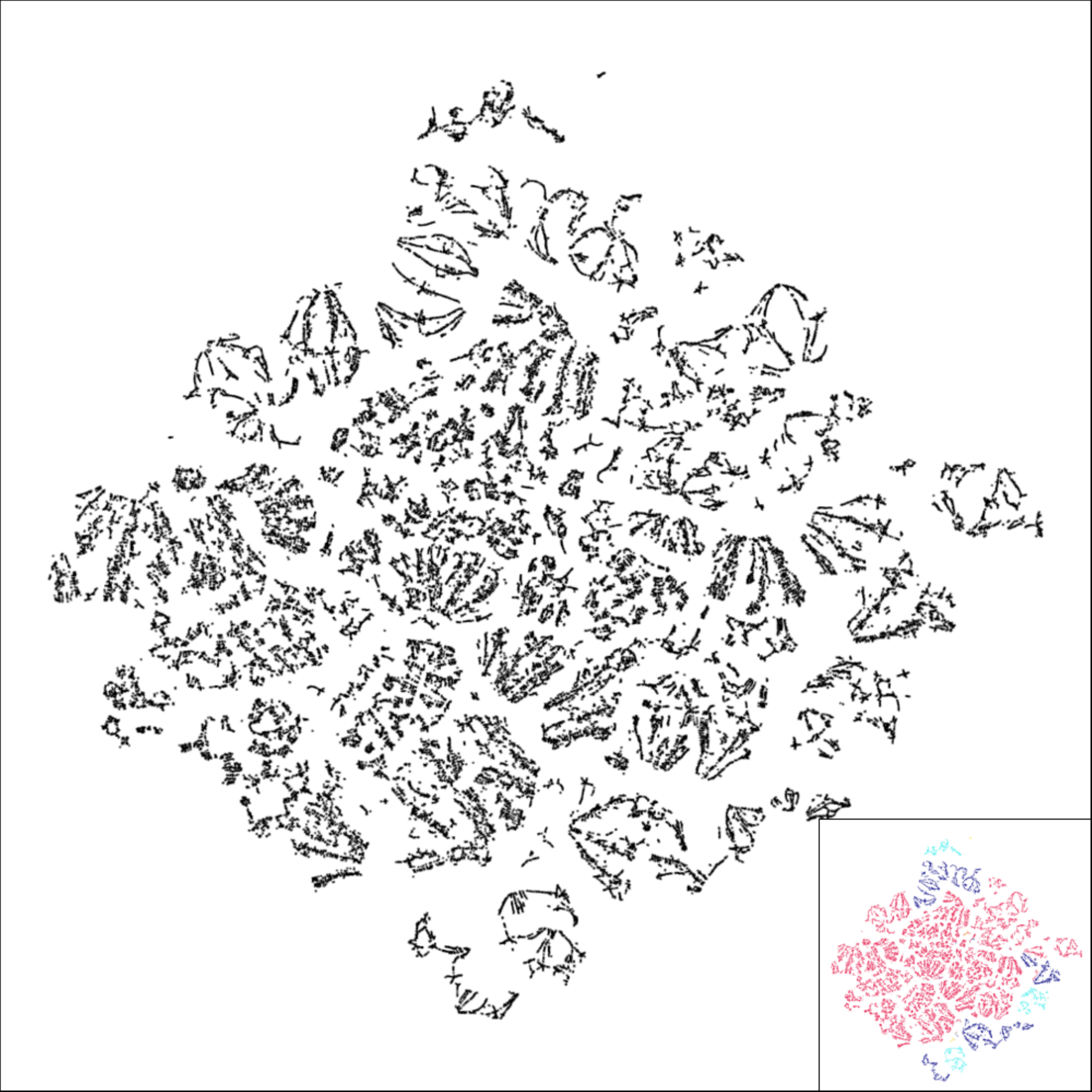}}
		
		\vspace{\baselineskip}
		
		\subcaptionbox{LargeVis}
		{\includegraphics[width=\comparedfigwidth]{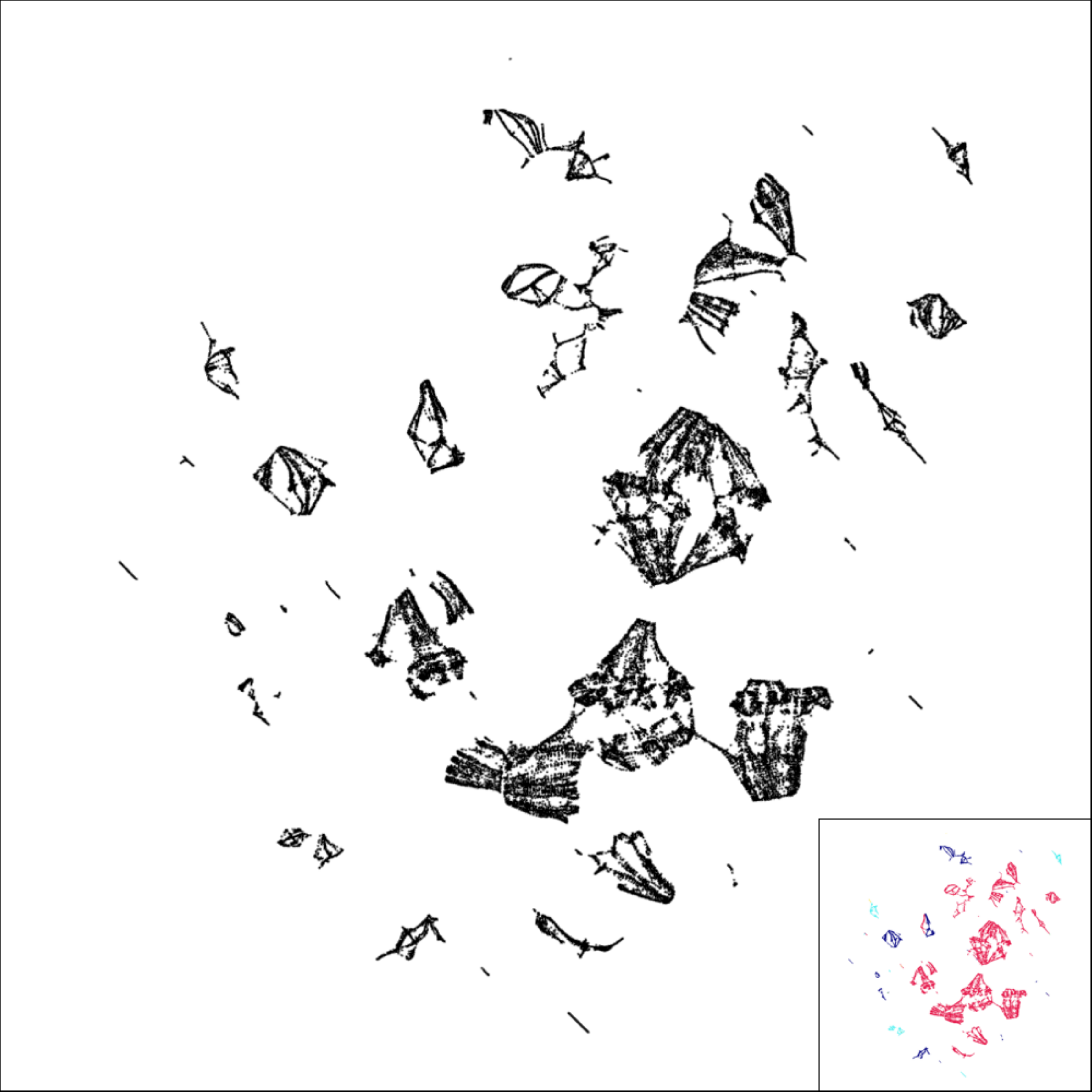}}
		\subcaptionbox{UMAP}
		{\includegraphics[width=\comparedfigwidth]{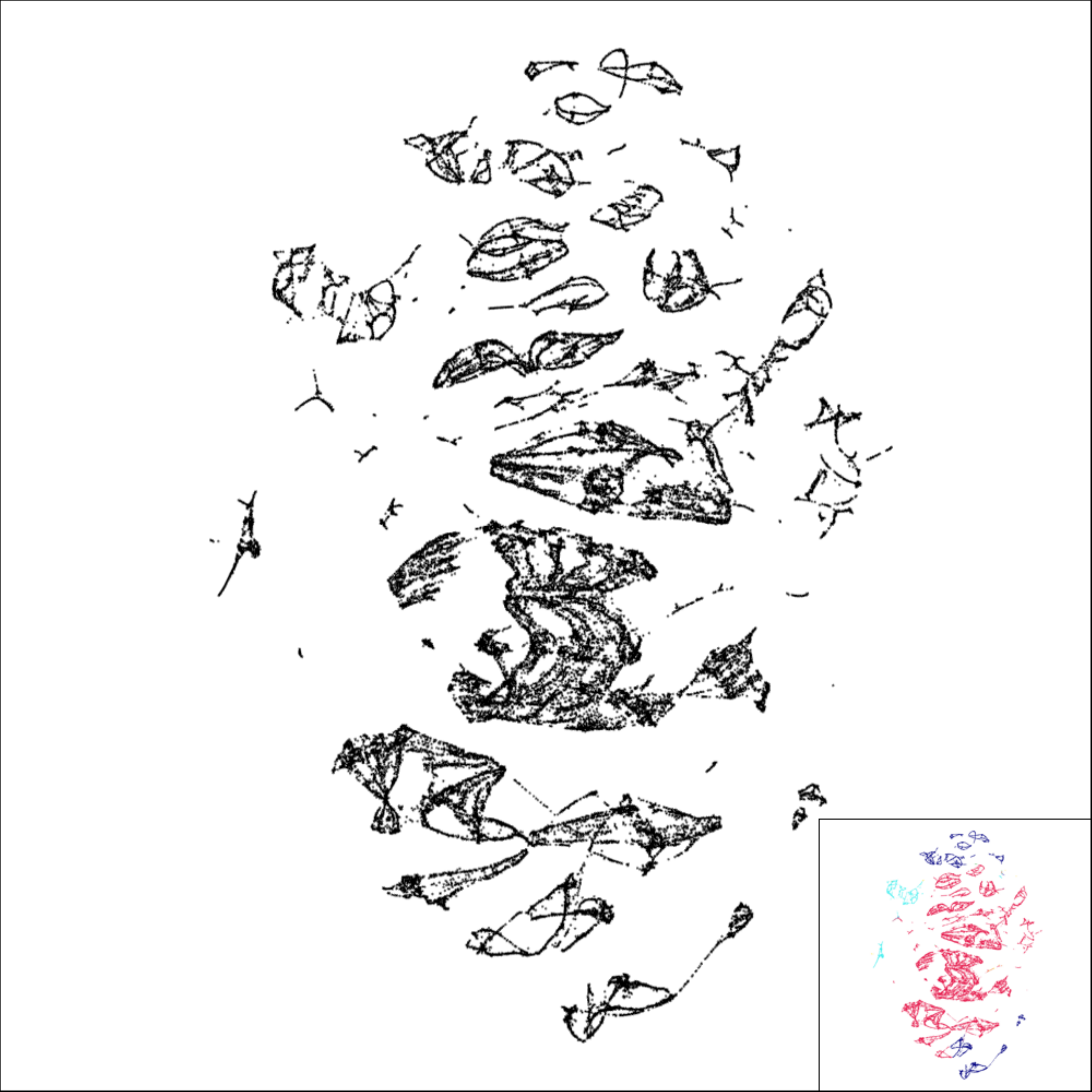}}
	\end{center}
	\caption{Visualizations of the \texttt{SHUTTLE} data set by using the compared methods. The classes are shown by colors in the small sub-figures.}
	\label{fig:vis_shuttle}
	\vspace{-3mm}
\end{figure}

\begin{figure}[p]
	\begin{center}
		\subcaptionbox{SCE (47 seconds)}
		{\includegraphics[width=\comparedfigwidth]{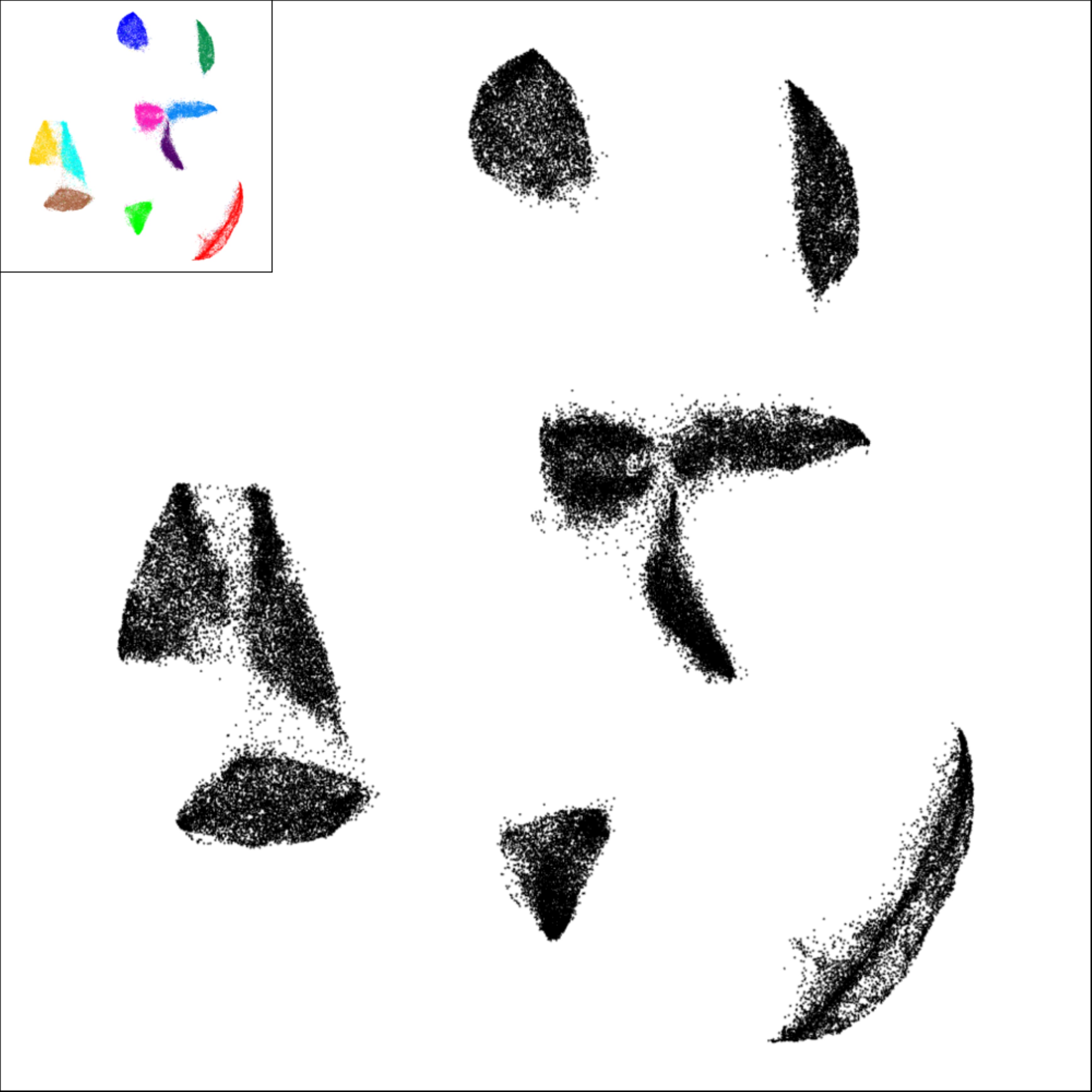}}
		\subcaptionbox{t-SNE}
		{\includegraphics[width=\comparedfigwidth]{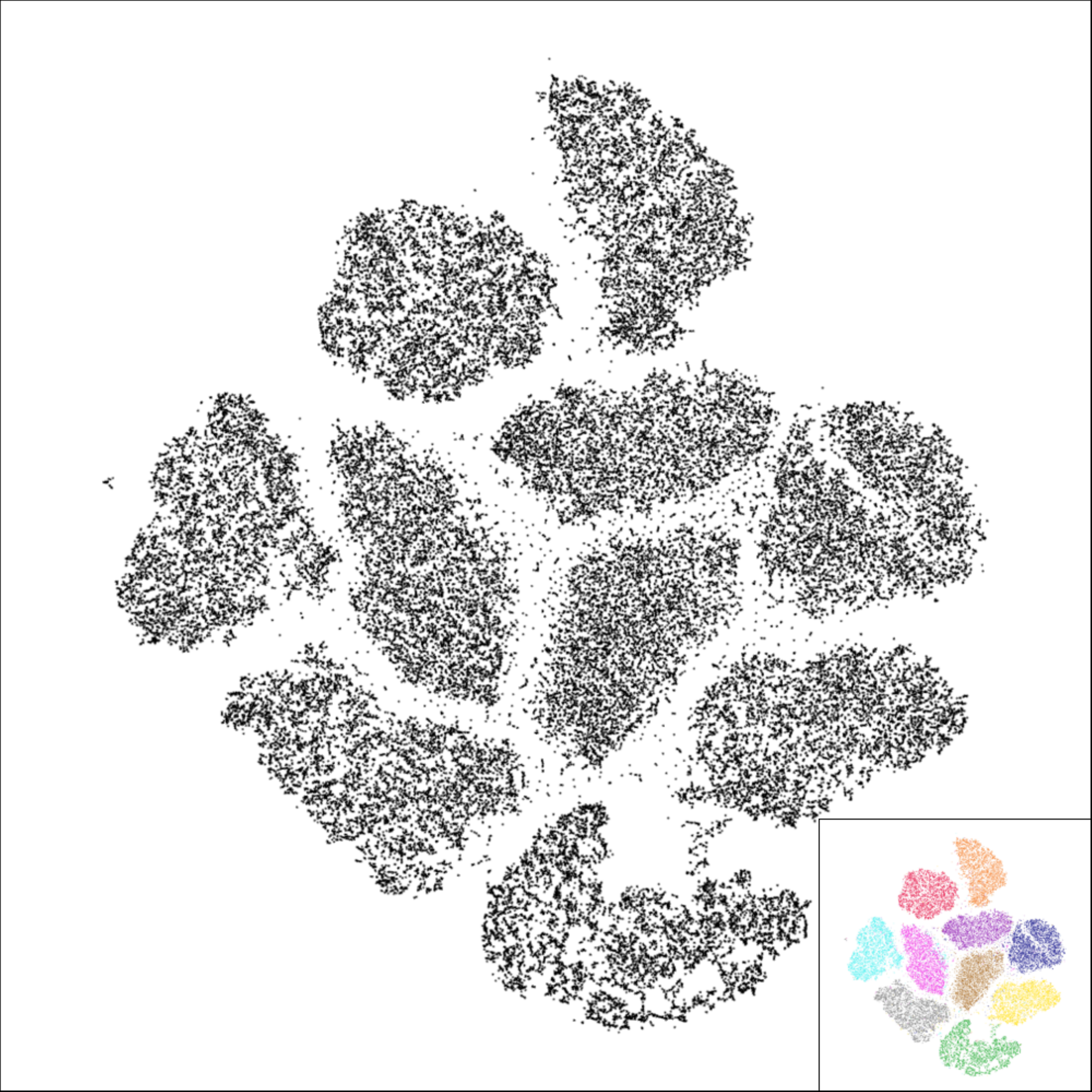}}
		
		\vspace{\baselineskip}
		
		\subcaptionbox{LargeVis}
		{\includegraphics[width=\comparedfigwidth]{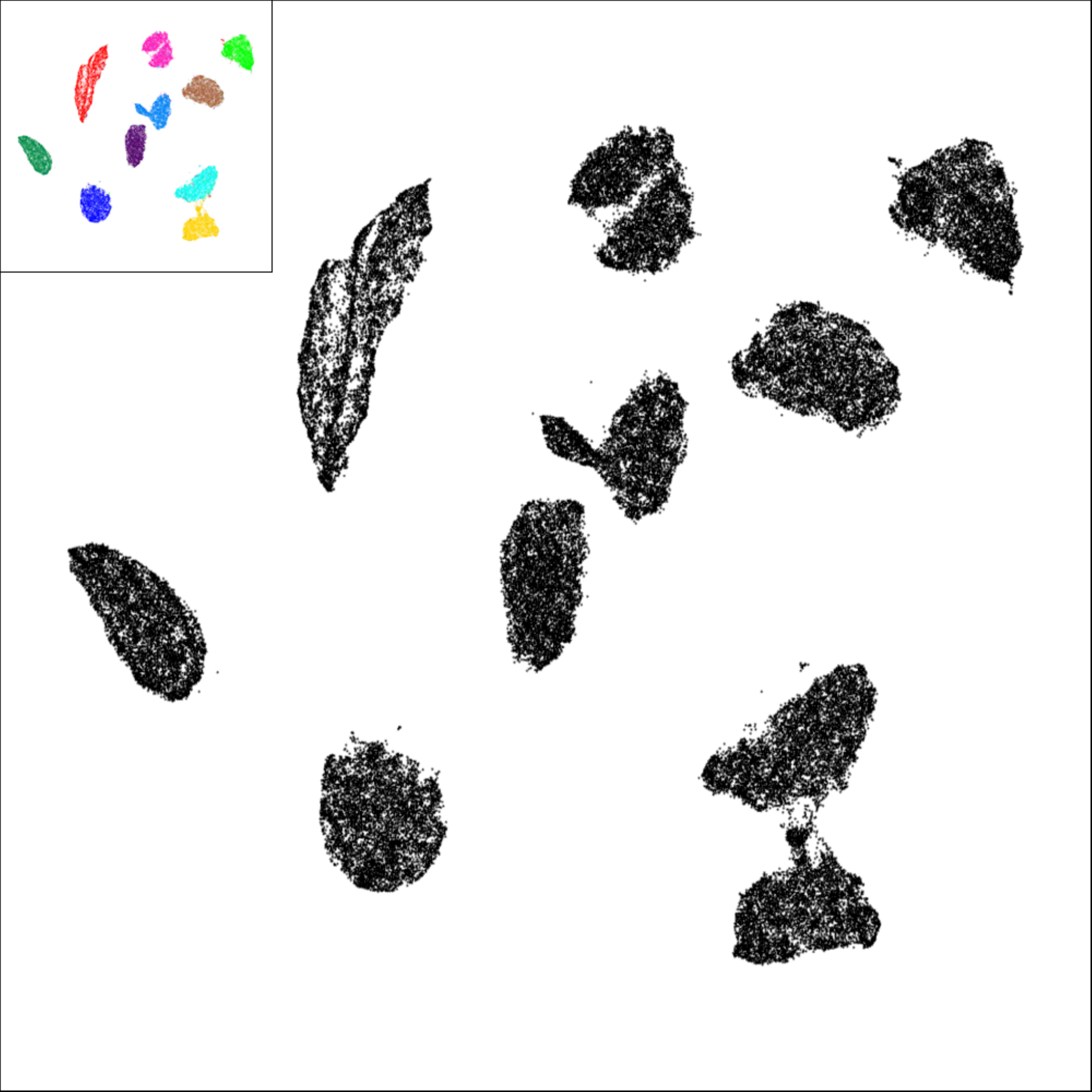}}
		\subcaptionbox{UMAP}
		{\includegraphics[width=\comparedfigwidth]{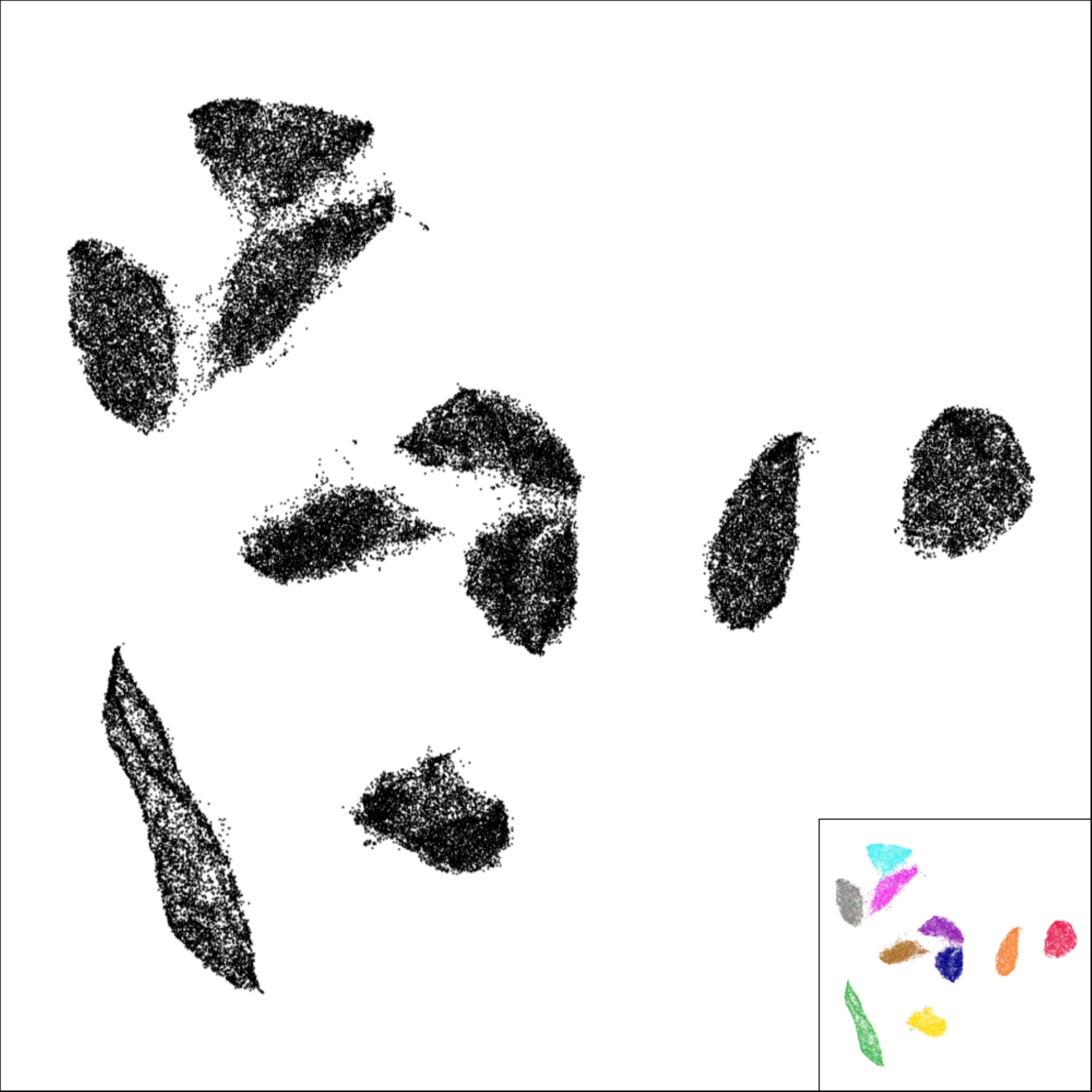}}
	\end{center}
	\caption{Visualizations of the \texttt{MNIST} data set by using the compared methods. The classes are shown by colors in the small sub-figures.}
	\label{fig:vis_mnist}
	\vspace{-3mm}
\end{figure}

\begin{figure}[p]
	\begin{center}
		\subcaptionbox{SCE (75 seconds)}
		{\includegraphics[width=\comparedfigwidth]{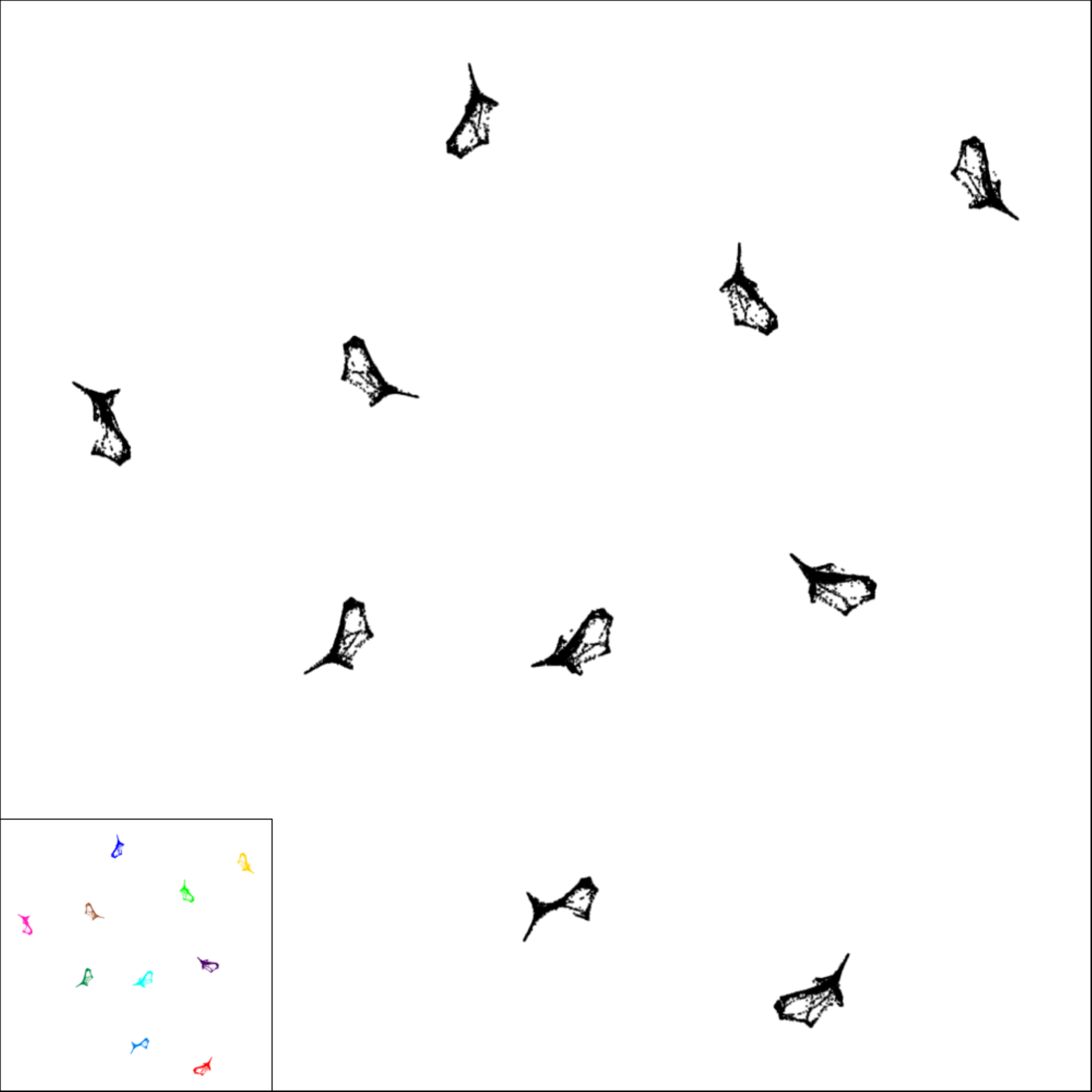}}
		\subcaptionbox{t-SNE}
		{\includegraphics[width=\comparedfigwidth]{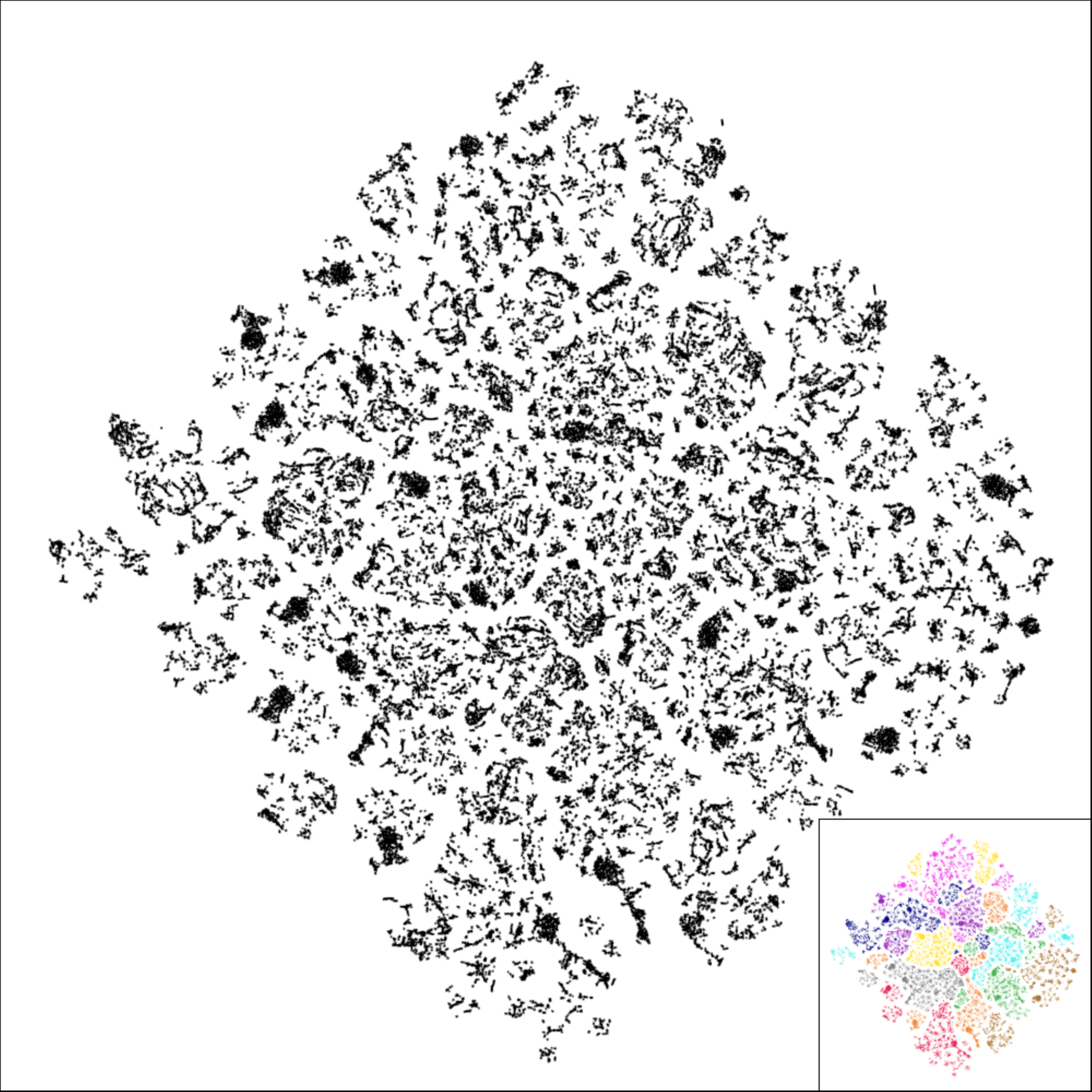}}
		
		\vspace{\baselineskip}
		
		\subcaptionbox{LargeVis}
		{\includegraphics[width=\comparedfigwidth]{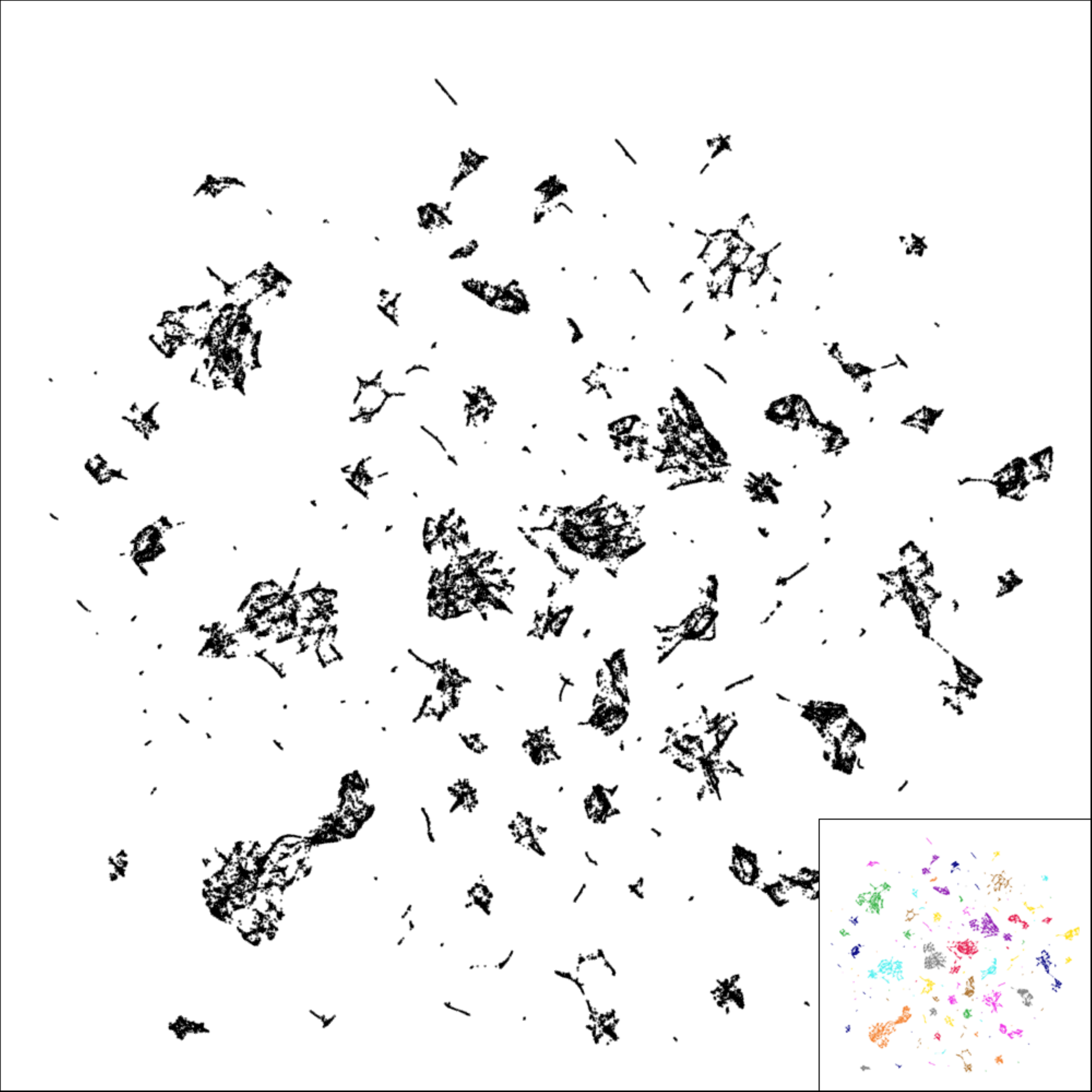}}
		\subcaptionbox{UMAP}
		{\includegraphics[width=\comparedfigwidth]{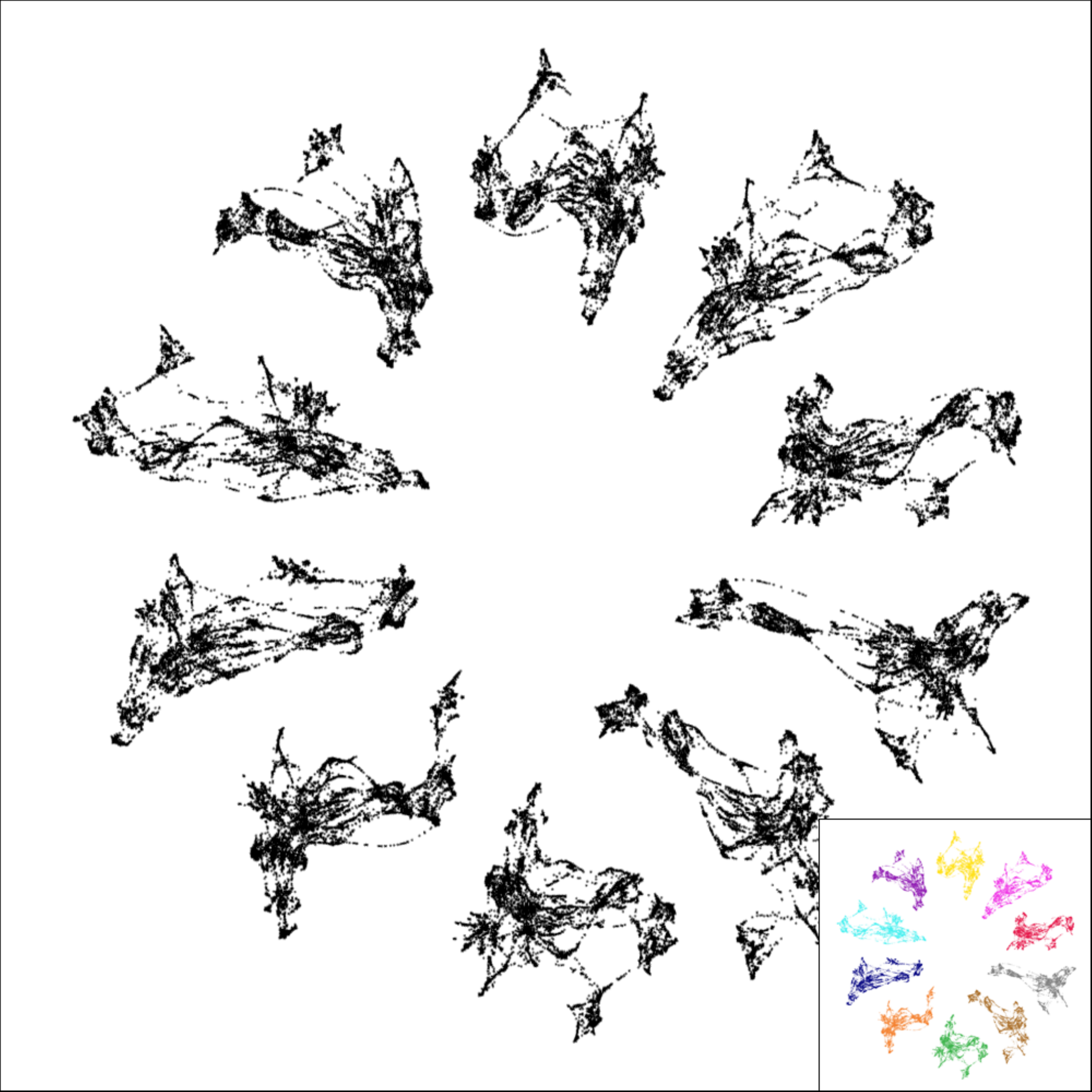}}
	\end{center}
	\caption{Visualizations of the \texttt{IJCNN} data set by using the compared methods. The classes are shown by colors in the small sub-figures.}
	\label{fig:vis_ijcnn}
	\vspace{-3mm}
\end{figure}

\begin{figure}[p]
	\begin{center}
		\subcaptionbox{SCE (660 seconds)}
		{\includegraphics[width=\comparedfigwidth]{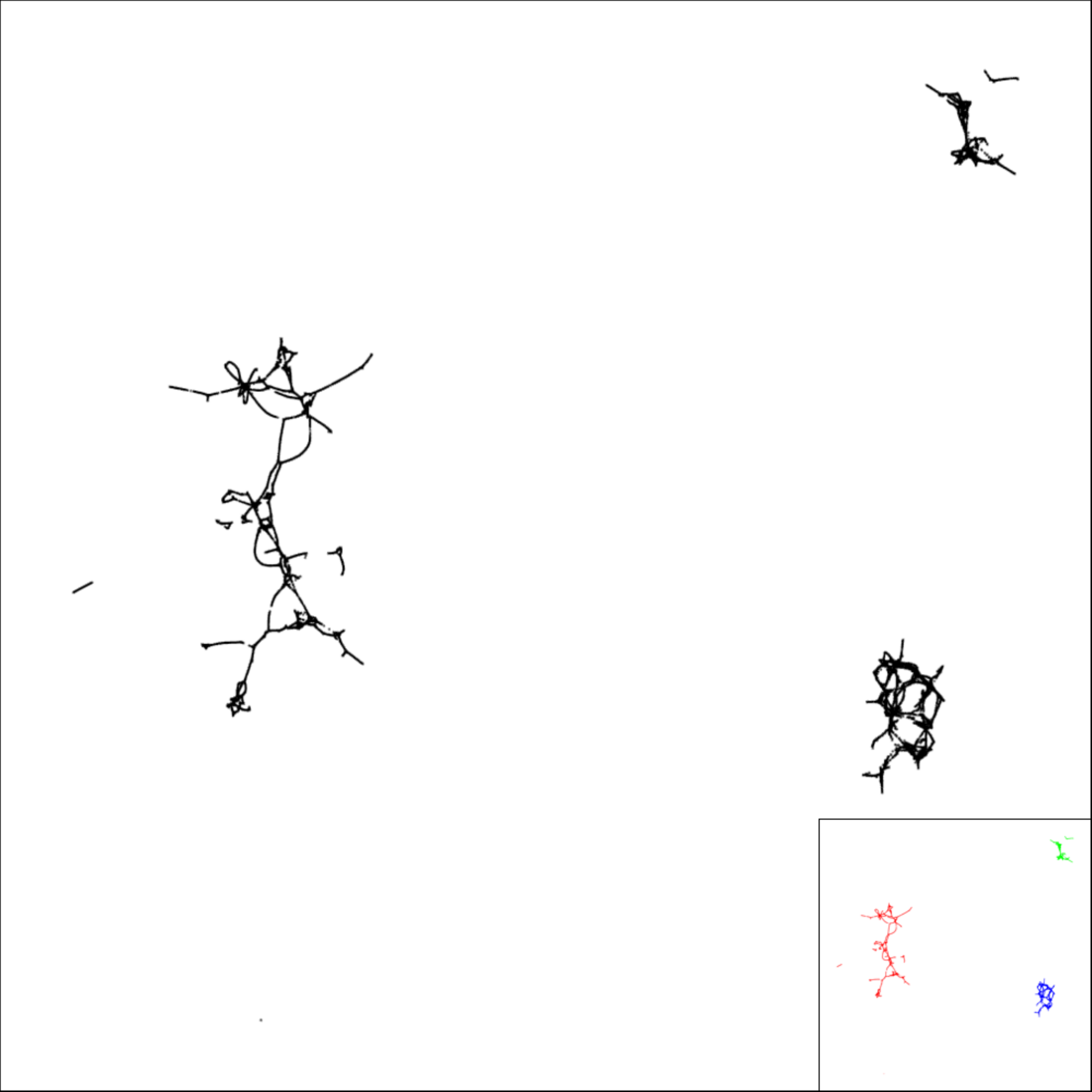}}
		\subcaptionbox{t-SNE}
		{\includegraphics[width=\comparedfigwidth]{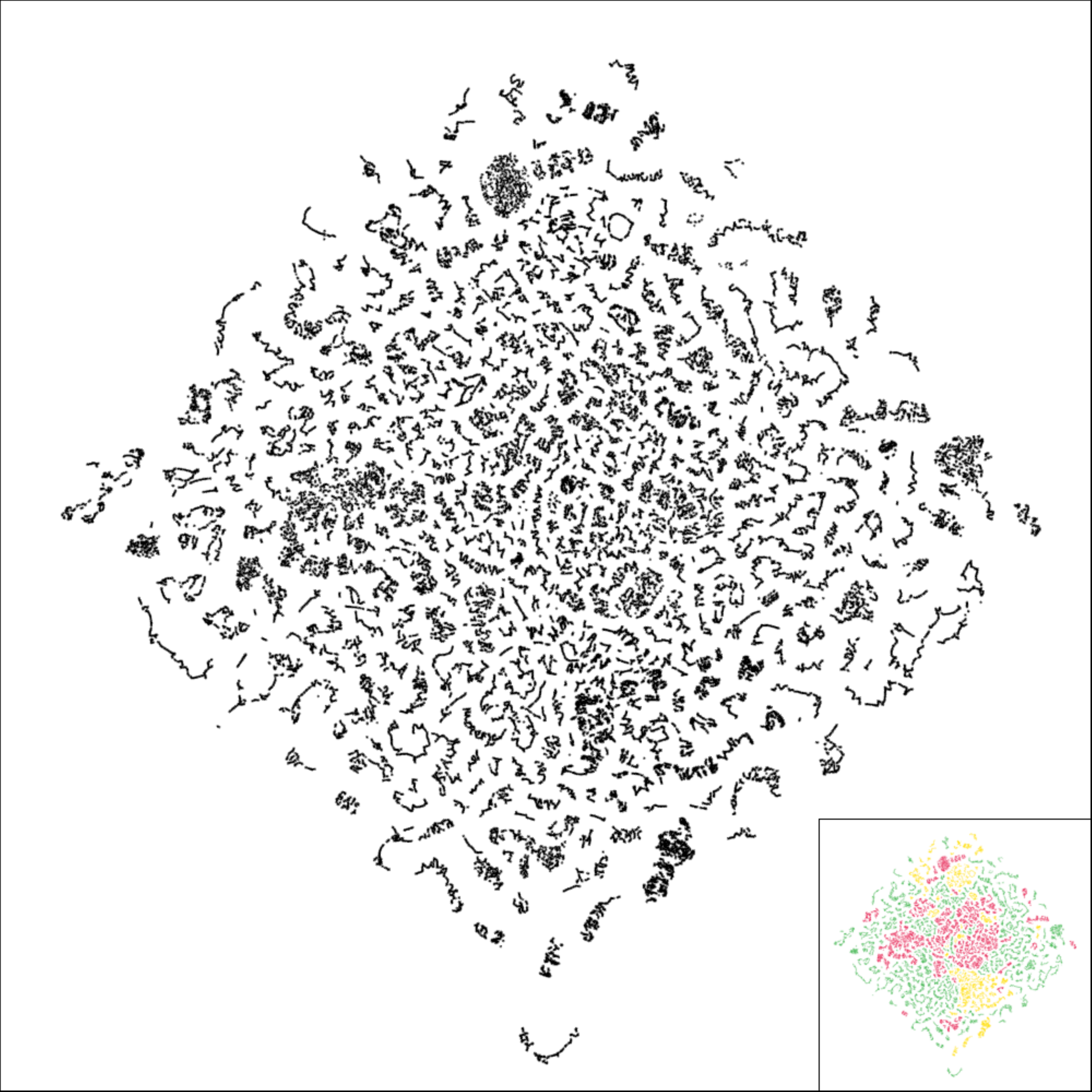}}
		
		\vspace{\baselineskip}
		
		\subcaptionbox{LargeVis}
		{\includegraphics[width=\comparedfigwidth]{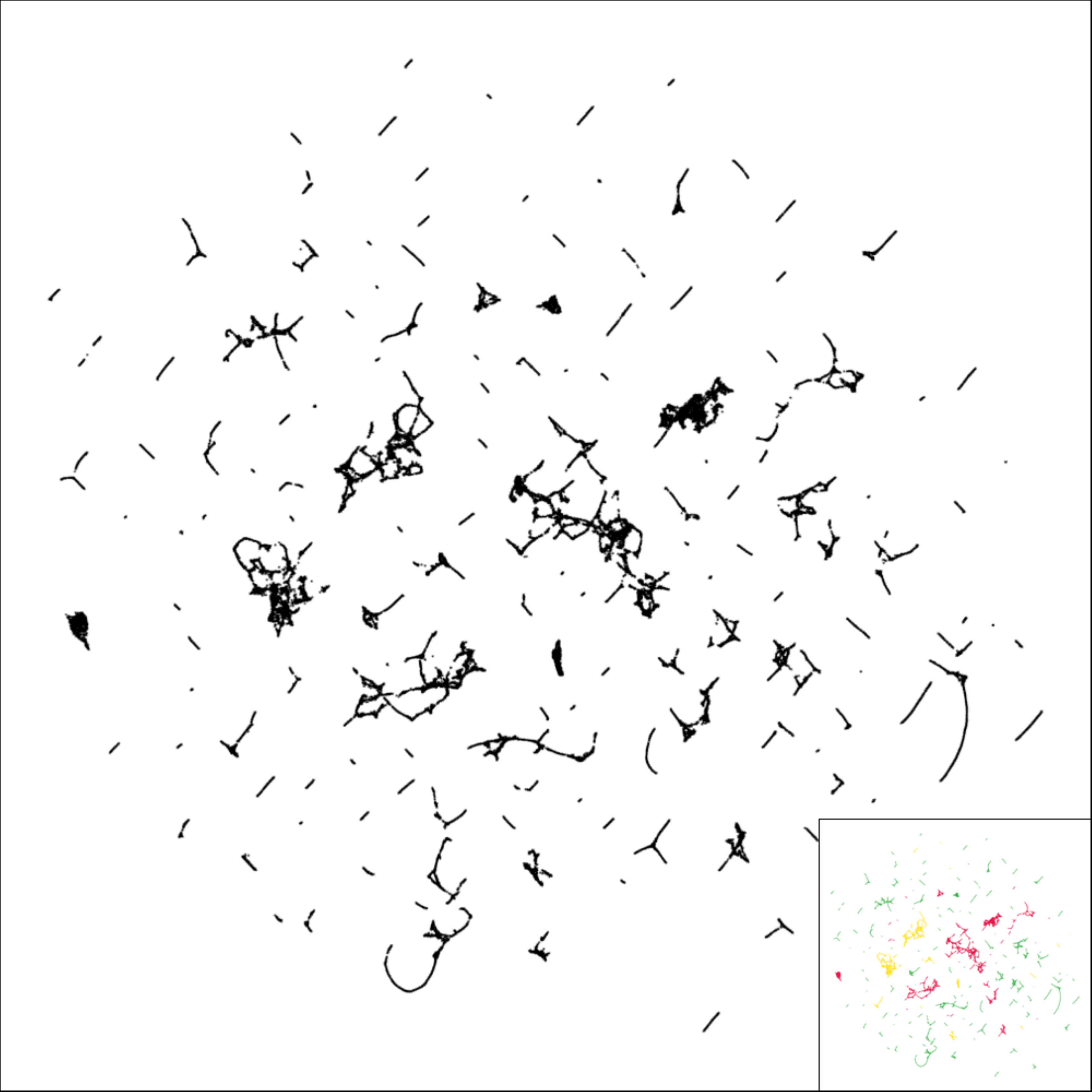}}
		\subcaptionbox{UMAP}
		{\includegraphics[width=\comparedfigwidth]{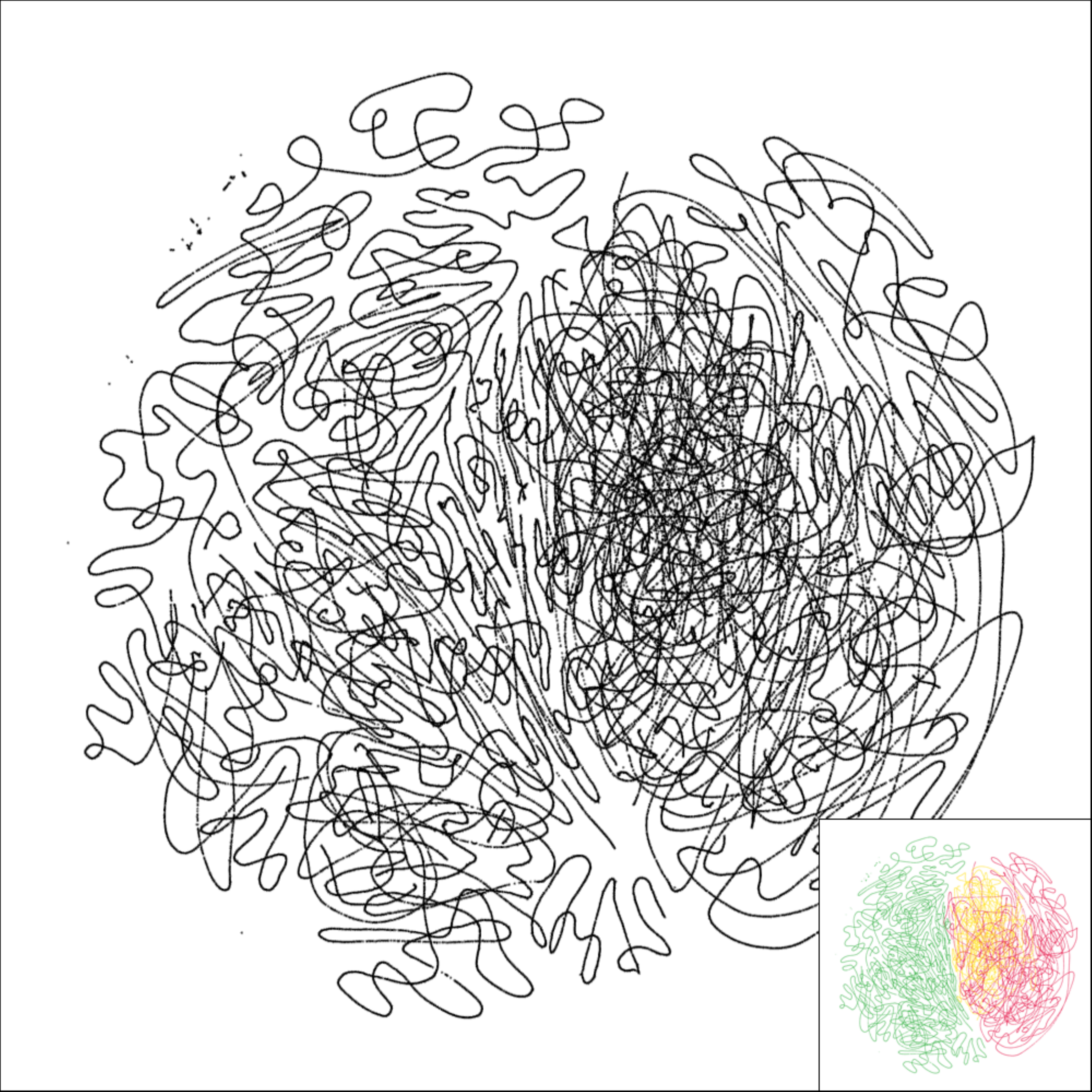}}
	\end{center}
	\caption{Visualizations of the \texttt{TOMORADAR} data set by using the compared methods. The classes are shown by colors in the small sub-figures.}
	\label{fig:vis_tomoradar}
	\vspace{-3mm}
\end{figure}

\setlength{\fboxsep}{0pt}
\begin{figure}[p]
	\begin{center}
		\subcaptionbox{SCE (7713 seconds)}
		{\fbox{\includegraphics[width=\comparedfigwidth,height=\comparedfigwidth]{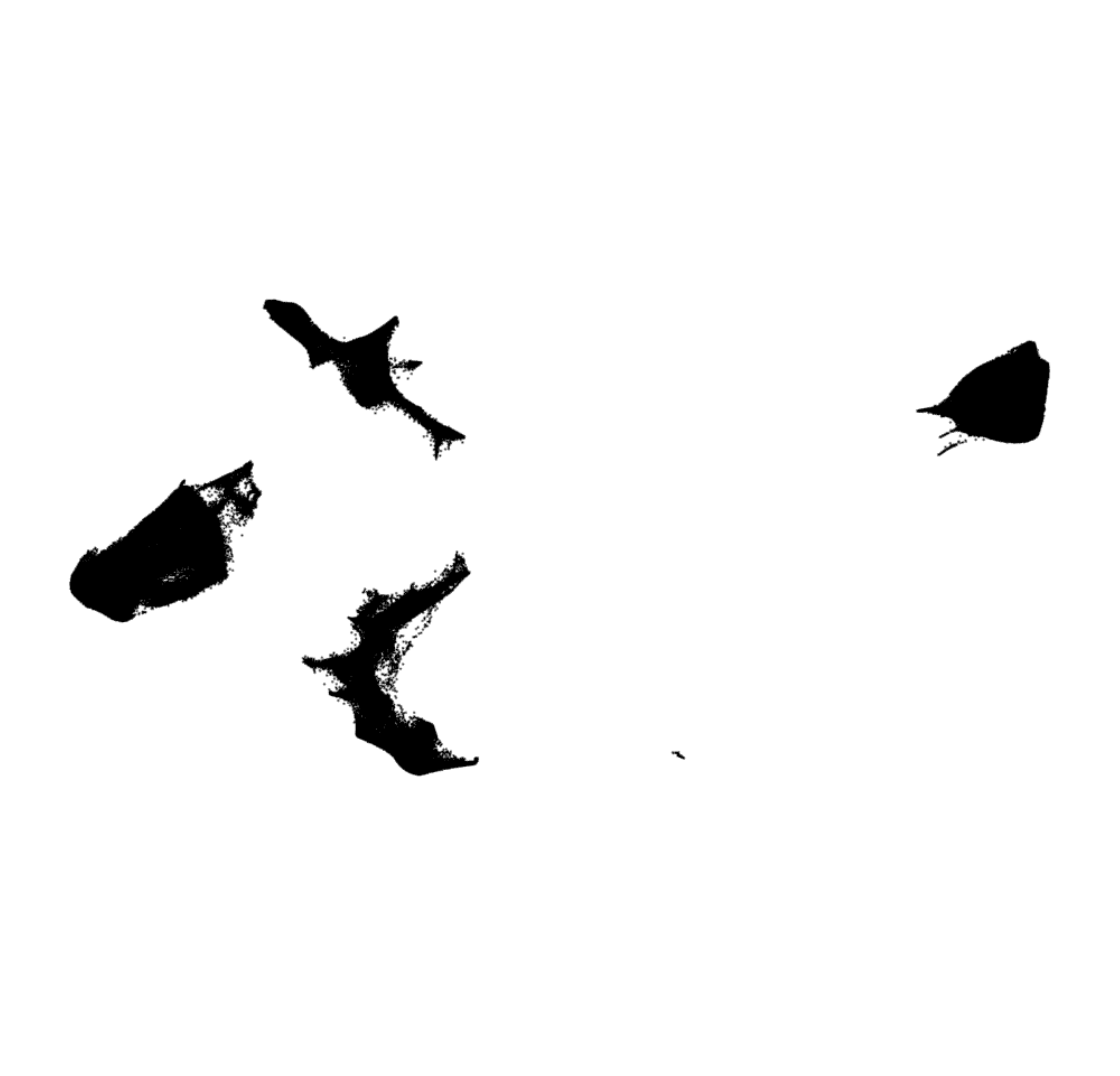}}}
		\subcaptionbox{t-SNE}
		{\fbox{\includegraphics[width=\comparedfigwidth,height=\comparedfigwidth]{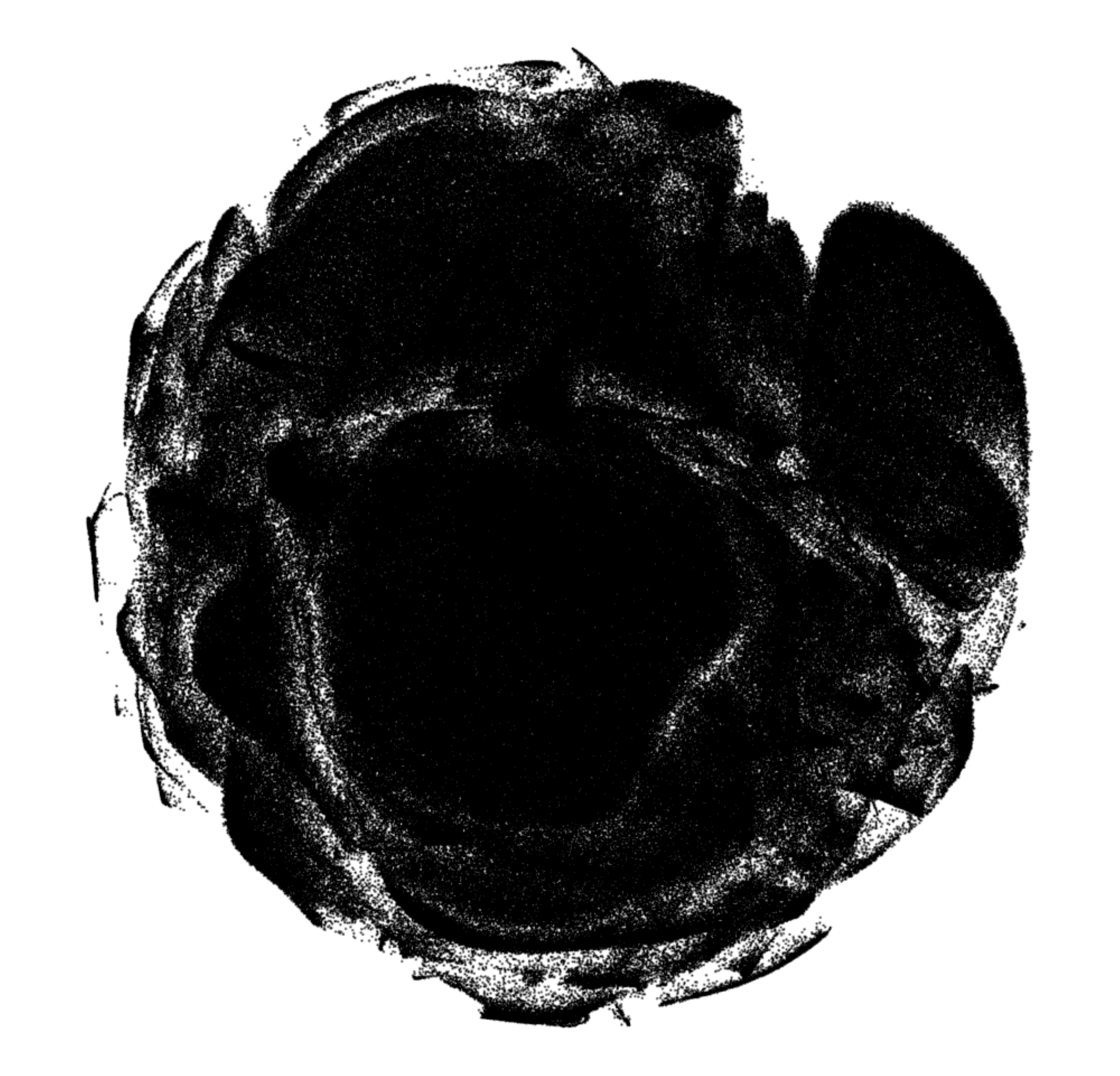}}}
		
		\vspace{\baselineskip}
		
		\subcaptionbox{LargeVis}
		{\fbox{\includegraphics[width=\comparedfigwidth,height=\comparedfigwidth]{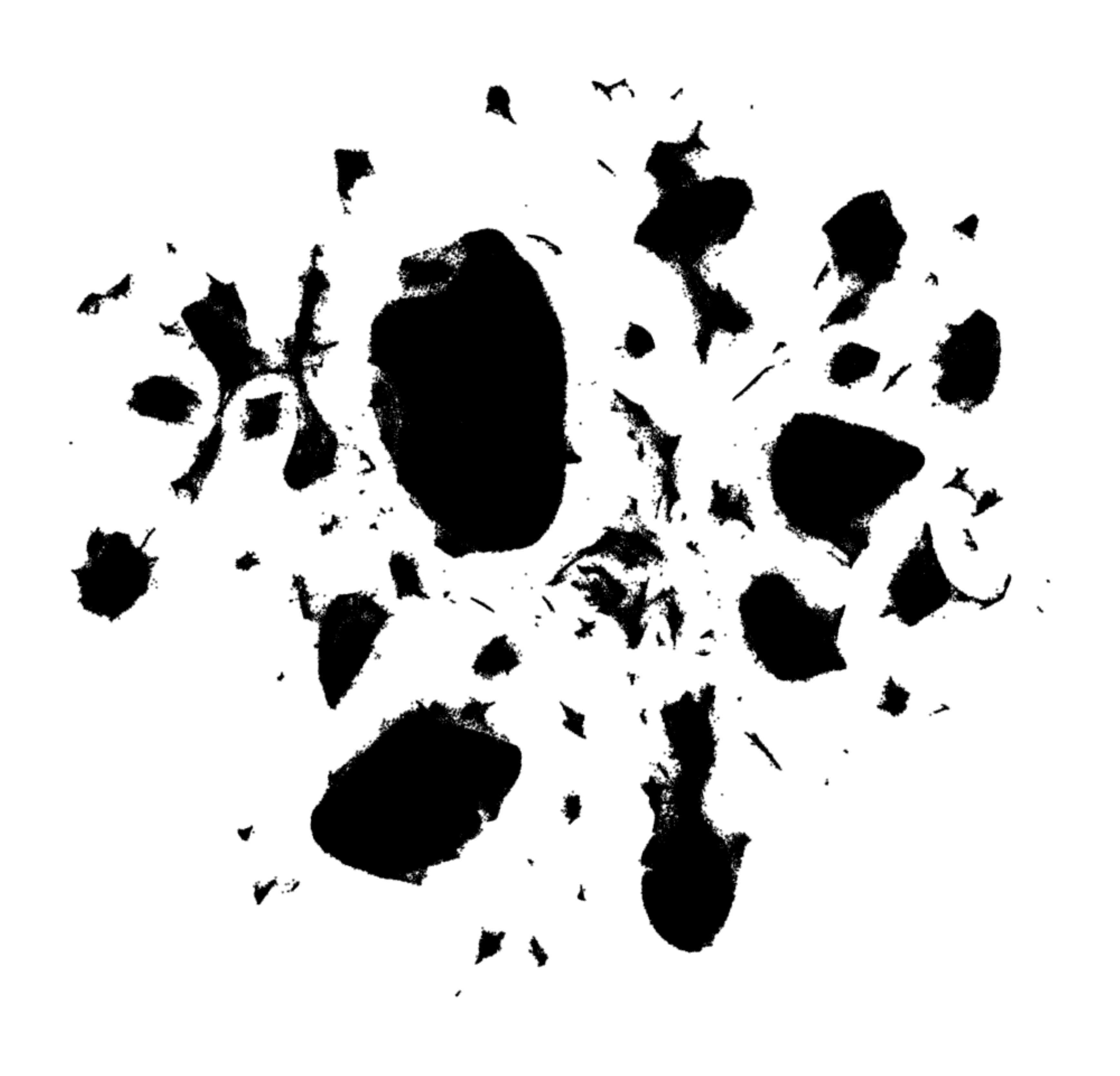}}}
		\subcaptionbox{UMAP}
		{\fbox{\includegraphics[width=\comparedfigwidth,height=\comparedfigwidth]{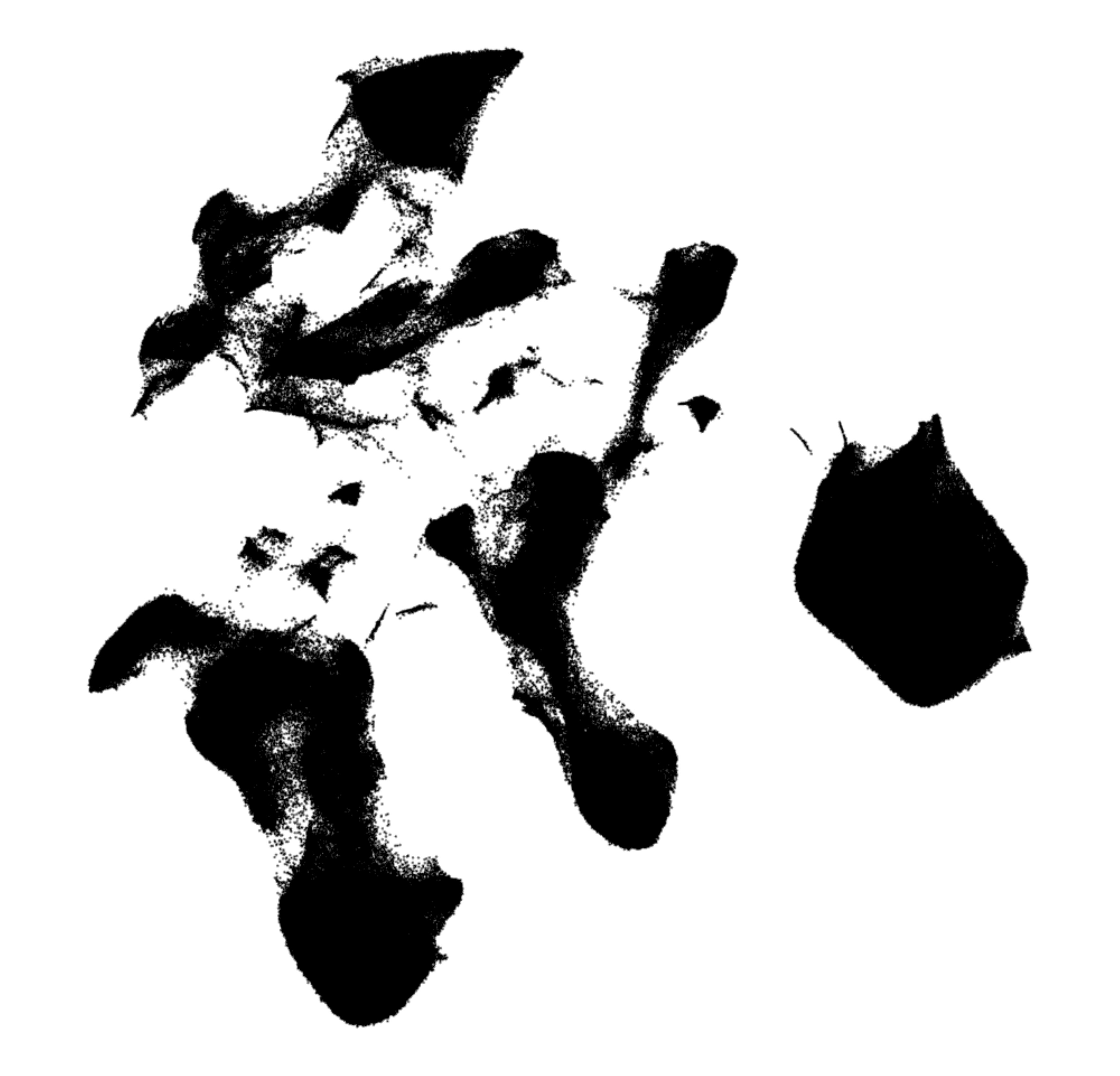}}}
	\end{center}
	\caption{Visualizations of the \texttt{FLOW-CYTOMETRY} data set by using the compared methods.}
	\label{fig:vis_flow_cytometry}
	\vspace{-3mm}
\end{figure}

\begin{figure}[p]
	\begin{center}
		\subcaptionbox{SCE (8969 seconds)}
		{\fbox{\includegraphics[width=\comparedfigwidth,height=\comparedfigwidth]{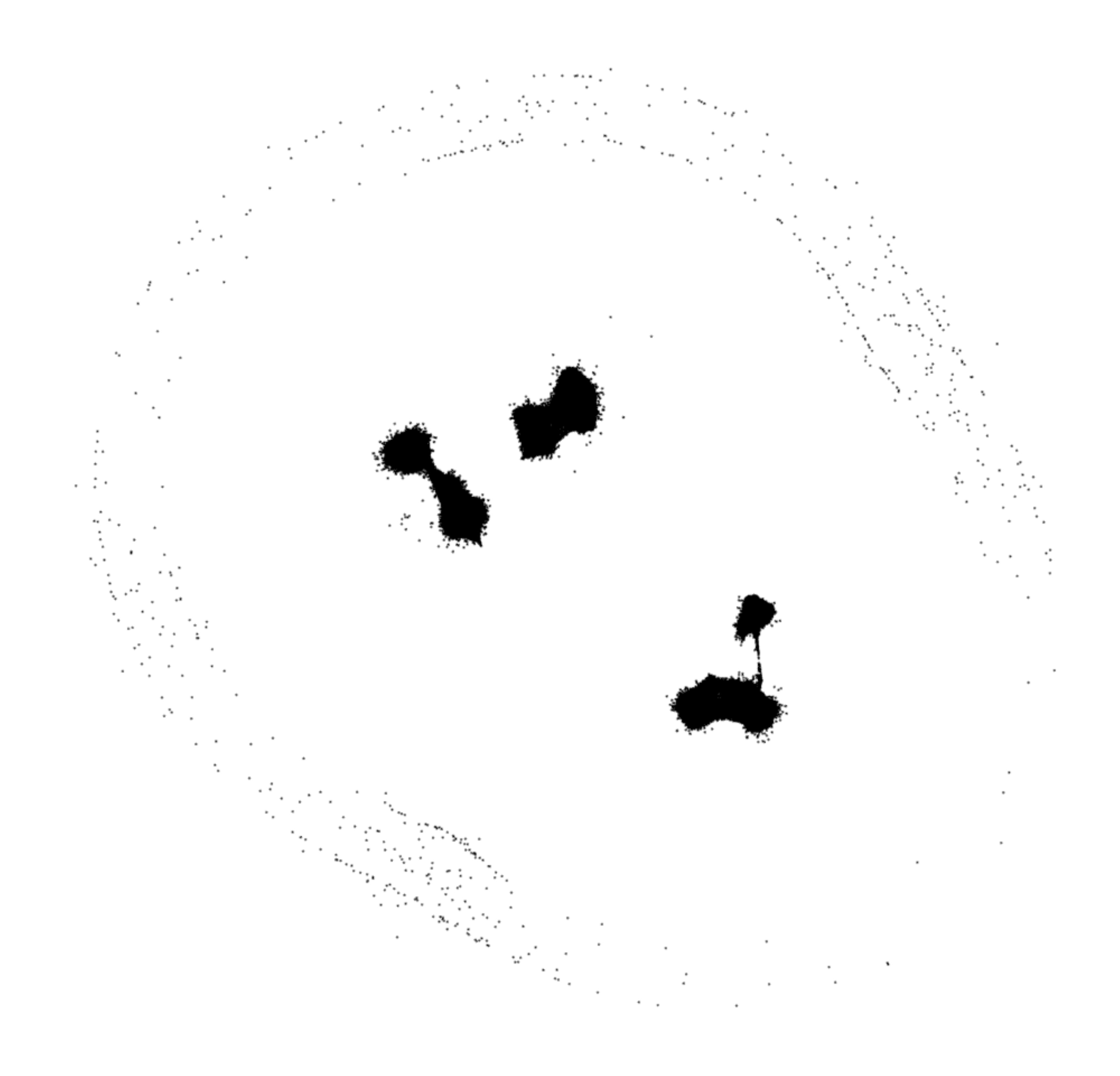}}}
		\subcaptionbox{t-SNE}
		{\fbox{\includegraphics[width=\comparedfigwidth,height=\comparedfigwidth]{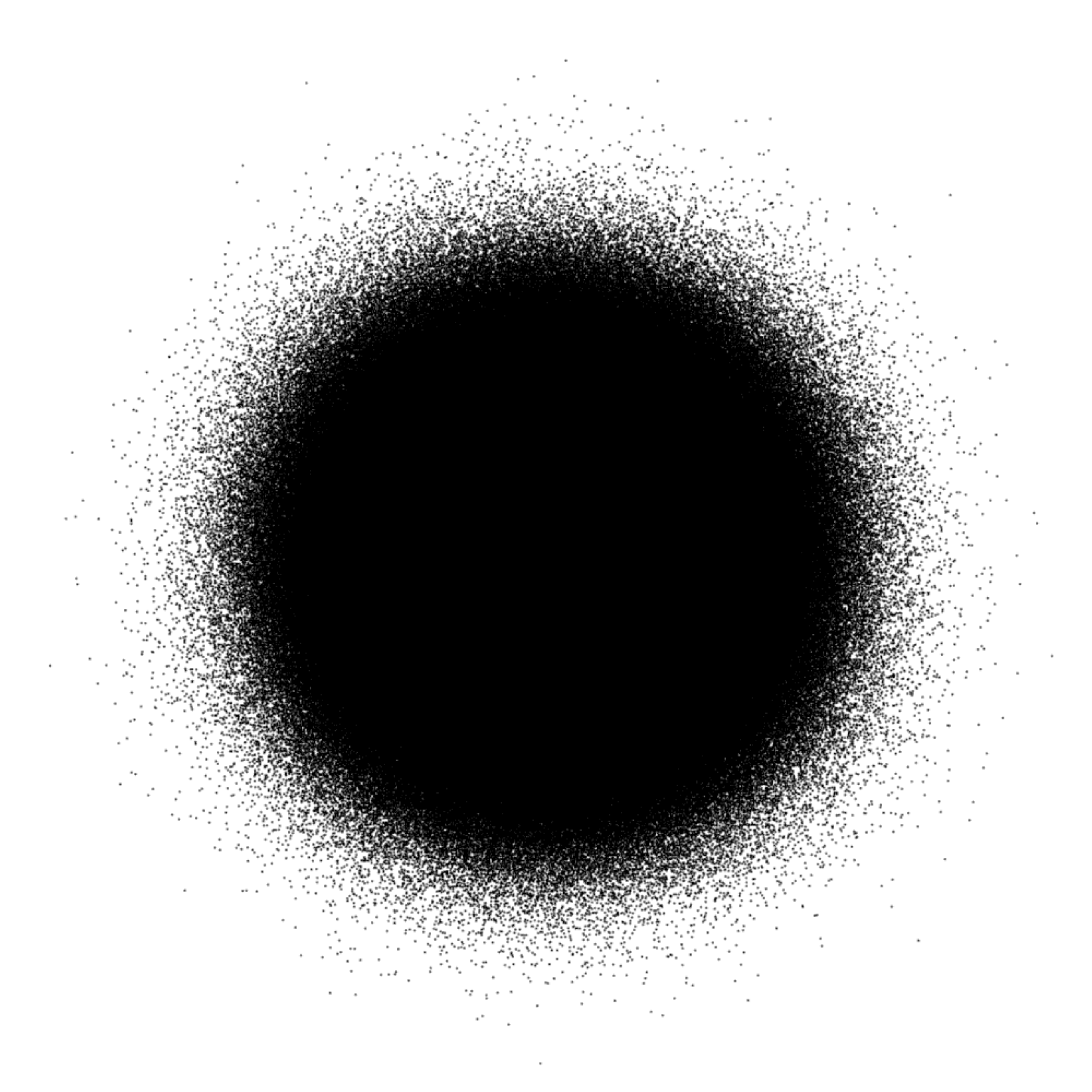}}}
		
		\vspace{\baselineskip}
		
		\subcaptionbox{LargeVis}
		{\fbox{\includegraphics[width=\comparedfigwidth,height=\comparedfigwidth]{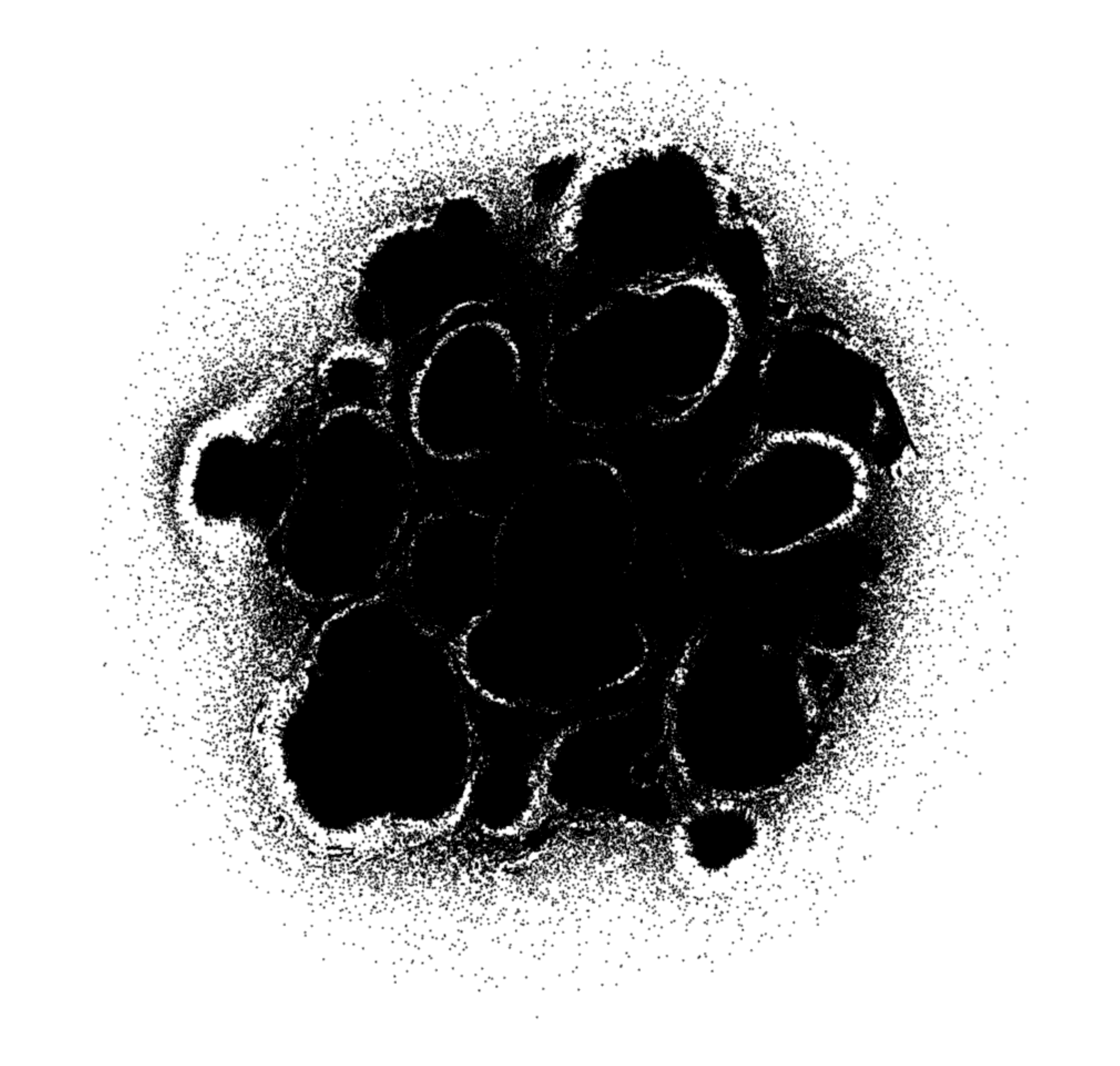}}}
		\subcaptionbox{UMAP}
		{\fbox{\includegraphics[width=\comparedfigwidth,height=\comparedfigwidth]{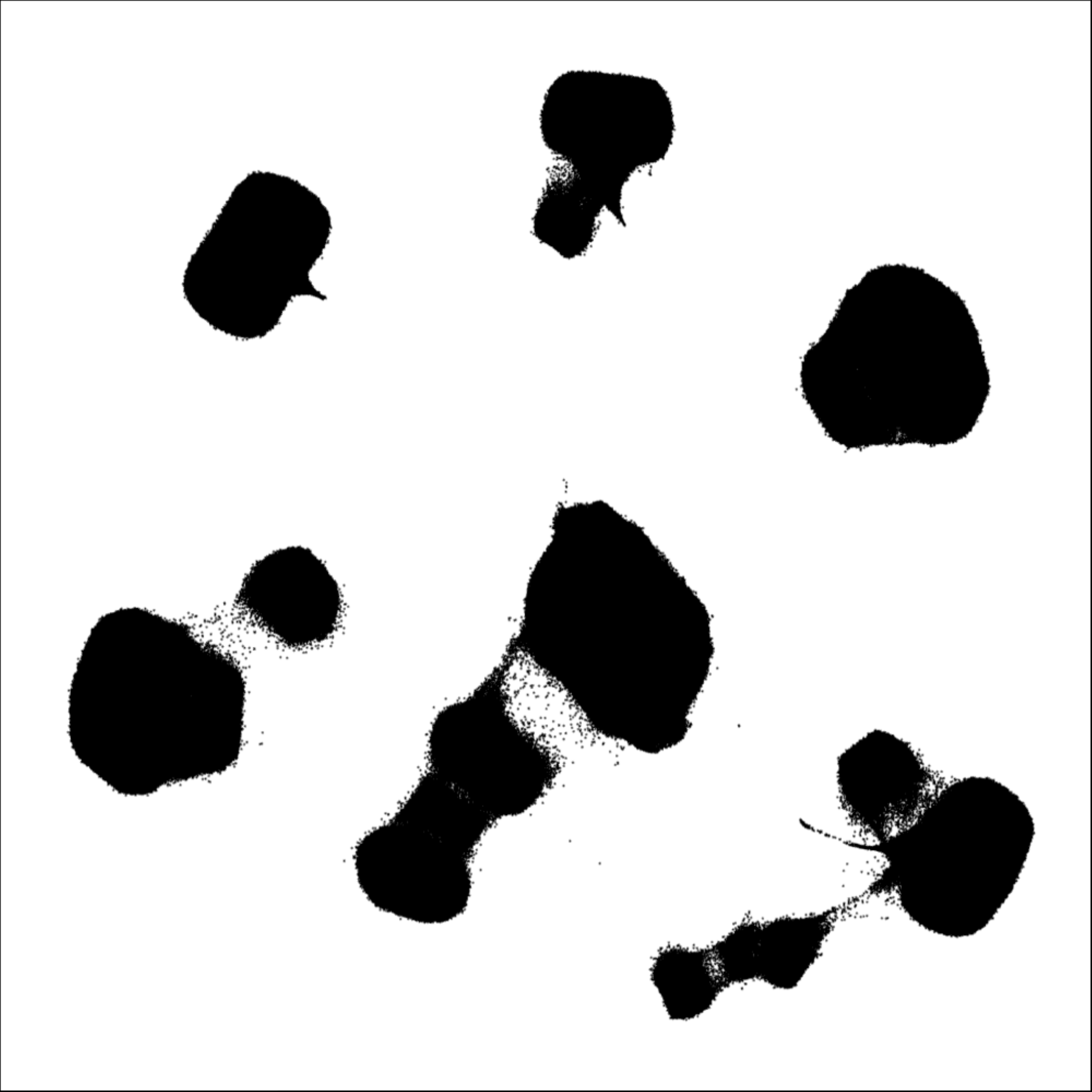}}}
	\end{center}
	\caption{Visualizations of the \texttt{HIGGS} data set by using the compared methods.}
	\label{fig:vis_higgs}
	\vspace{-3mm}
\end{figure}

\subsection{User study}
\label{sec:user_study}
Since visualizations are designed for human use, we have performed a user study about the human choice among GSNE visualizations corresponding to a range of $s$ values for seeing clusters. We can then compare the resulting $s$ values in SCE and t-SNE to see which is closer to the human choices. 

We have used the four smallest data sets \texttt{SHUTTLE}, \texttt{MNIST}, \texttt{IJCNN}, and \texttt{TOMORADAR}. For each data set, we have ran GSNE with $s=10^t\cdot N^{-2}$, where $t\in[-4,6]$ for \texttt{TOMORADAR} and $t\in[0,8]$ for the other data sets. These ranges of $s$ values were set to be wide enough from over-attractive to over-repulsive such that meaningful $s$ choices should take place in between.

The user interface of the study can be found in \url{http://clres.cs.hut.fi/ClAnalysis/webpage.html}. For each data set, the series of visualizations are presented to a user (see the supplemental document Section 2 for a screenshot), where he or she uses a slider to specify the $s$ value and inspects the corresponding precomputed visualization. The user selects a preferred \textit{s} value for cluster visualization and then presses the ``Next'' button. The system records the user choice, and the study proceeds to the next series of visualization.

We first performed a controlled laboratory study, where 40 users came to the test computer room and gave their evaluations. We later conducted a crowdsourcing user study, following the established good practices of crowdsourcing for visualization research \citep{Borgo2017Crowdsourcing}. Using the crowdsourcing platform CrowdFlower\footnote{\url{https://www.crowdflower.com/}, now purchased by appen.com.}, we collected empirical data from a large and diverse population made up of 300 participants. We then combined the data of the controlled and crowdsourcing studies for the analysis, leading to 340 answers for each data set. More details about the user study can be found in the supplemental document Section 2.

The results are shown in Figure \ref{fig:user_study}. The SNE choice of $s$ according to Eq.~\ref{eq:snechoice} is shown by blue dotted-dash lines. We can see that the $s$ chosen by SNE is on the right of the human median (solid green line) for all data sets, which indicates that, in human eyes, GSNE with a smaller $s$ is often better than t-SNE for cluster visualization. In contrast, the SCE choice (red dash lines) are closer to the human median for all four data sets.

\newcommand{\histfigwidth}{7cm}
\begin{figure}[t]
	\begin{center}
		\subcaptionbox{SHUTTLE}
		{\includegraphics[width=\histfigwidth,height=\histfigwidth]{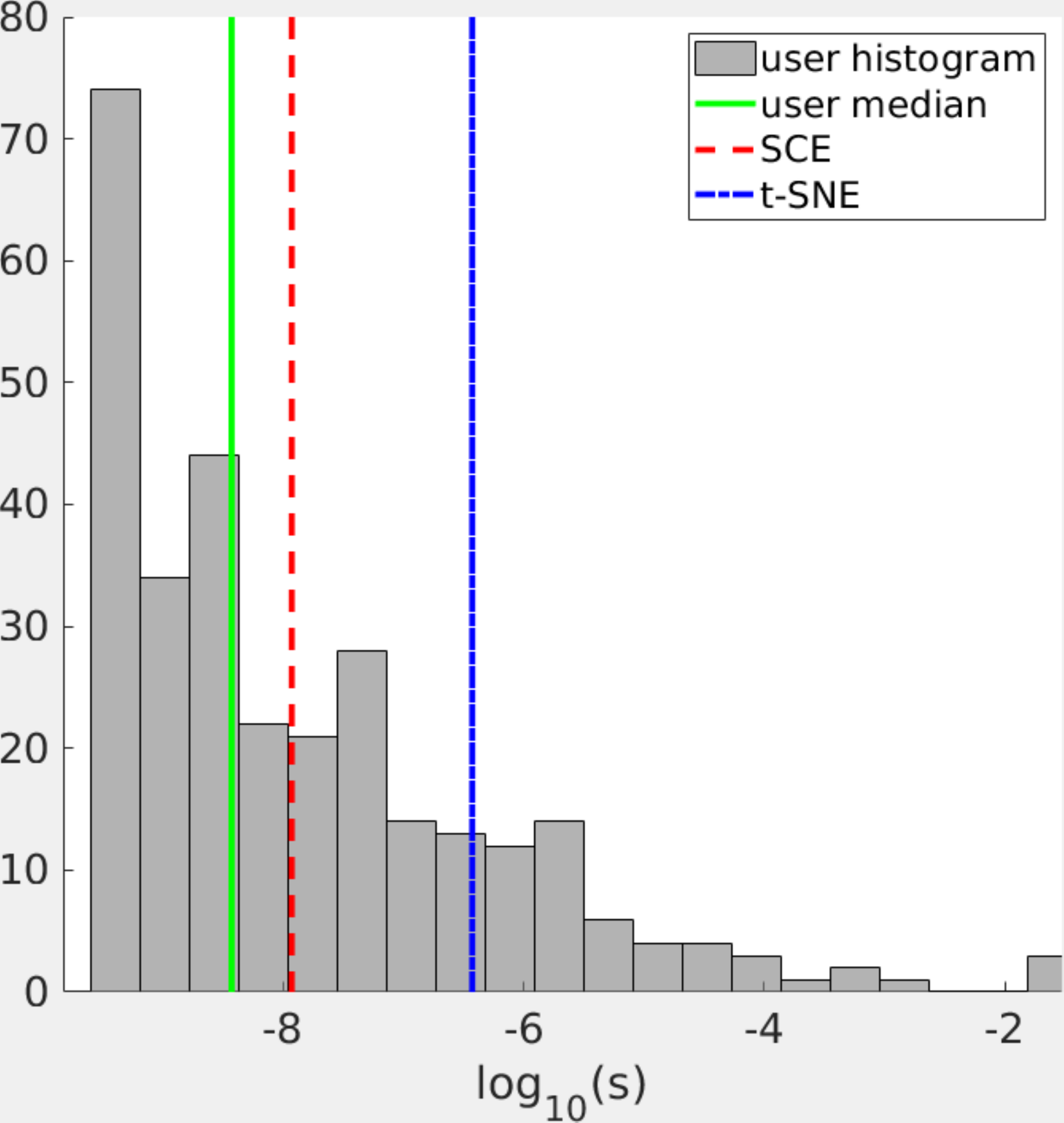}}
		\subcaptionbox{MNIST}
		{\includegraphics[width=\histfigwidth,height=\histfigwidth]{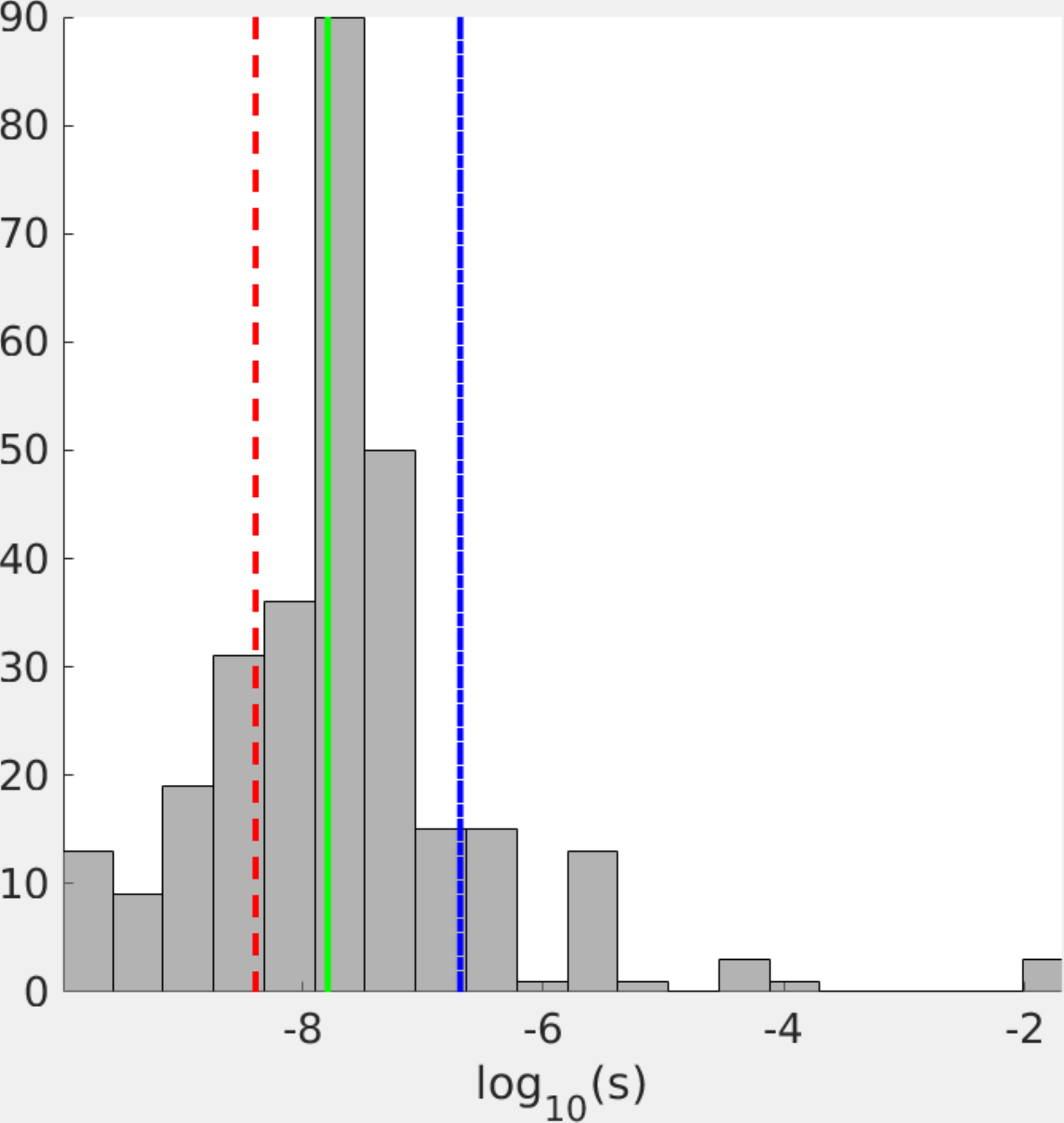}}
		
		\vspace{\baselineskip}
		
		\subcaptionbox{IJCNN}
		{\includegraphics[width=\histfigwidth,height=\histfigwidth]{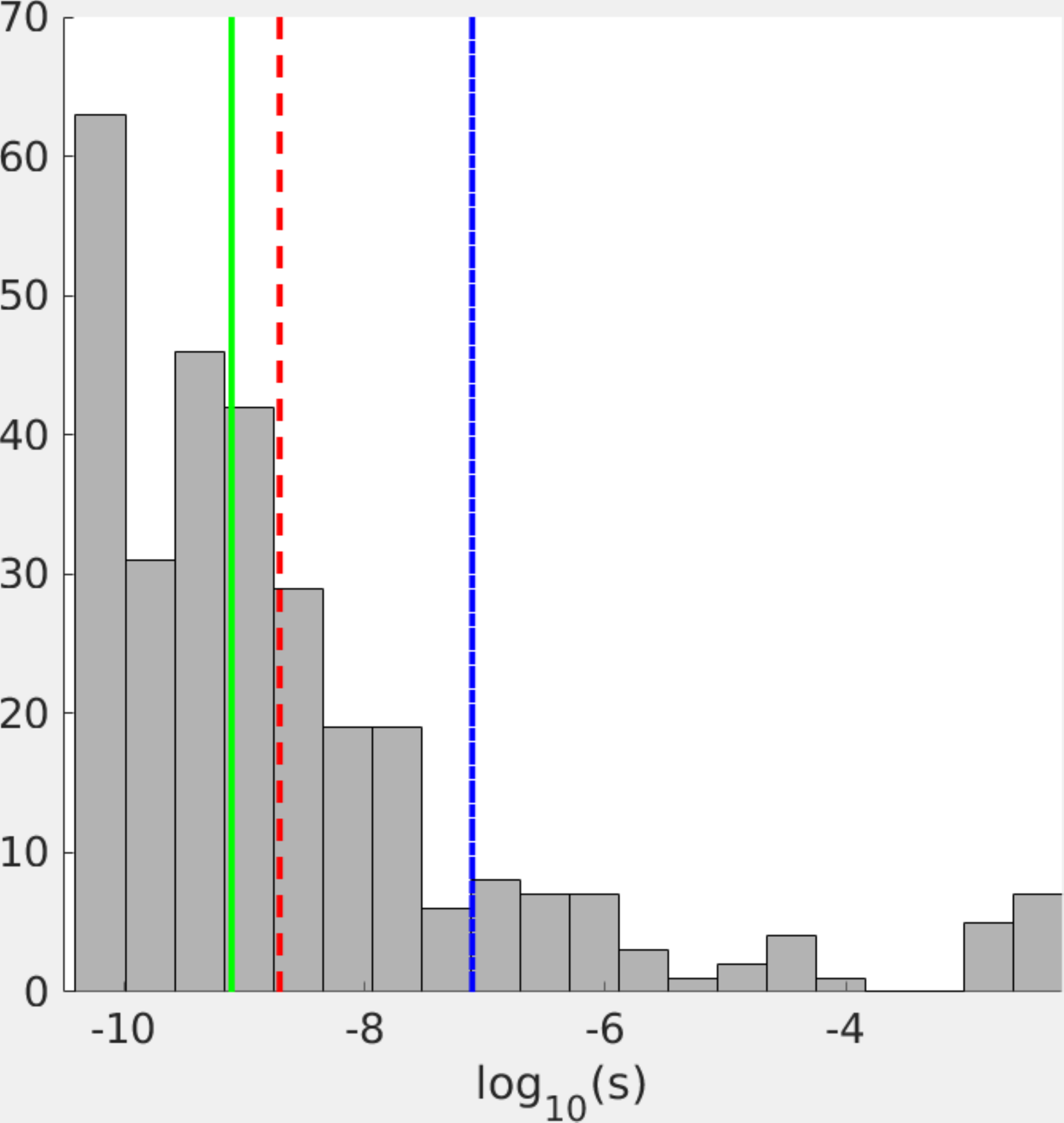}}
		\subcaptionbox{TOMORADAR}
		{\includegraphics[width=\histfigwidth,height=\histfigwidth]{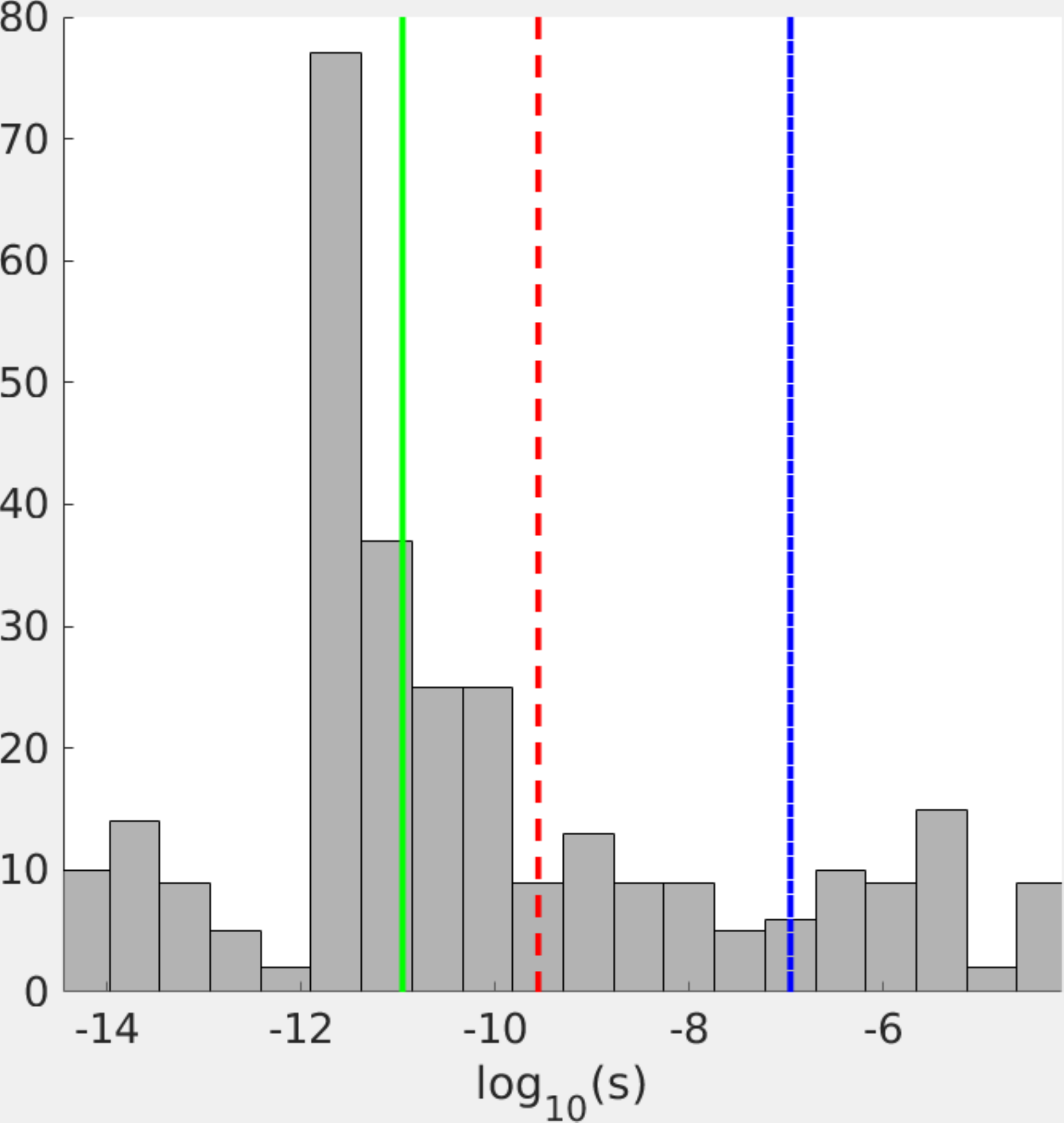}}
	\end{center}
	\caption{The $s$ values for cluster visualization: (gray bars) histogram of human choices, (green solid line) median of human choices, (red dash line) SCE choice, and (blue dash
		dotted line) t-SNE choice.}
	\label{fig:user_study}
	\vspace{-3mm}
\end{figure}

\subsection{Clustering quality}
\label{sec:clustering_quality}
Cluster visualization is an unsupervised task where the supervised labels are not available. Even if class labels are available in some data sets, they are not necessarily aligned with the intrinsic data clusters, for example, in \texttt{SHUTTLE} and \texttt{HIGGS}.

Here we have used an unsupervised approach to verify the SCE clustering quality. We first manually clustered the mapped data points in 2D space. Because most clusters are well separated in the SCE visualizations, the manual clustering is easy with little ambiguity. After the clustering, we reordered the rows and columns of the input $P$ matrix according to the cluster labels and examined the nonzero entries (blue dots) in Figure \ref{fig:blockvis}. We can see blockwise diagonal patterns in all plots, where each block corresponds to a cluster in the visualizations. Moreover, the blue dots within clusters are denser than those between clusters, which means SCE achieves good clustering quality.

\newcommand{\spyfigwidth}{4.5cm}
\begin{figure}[t]
	\begin{center}
		\subcaptionbox{SHUTTLE}
		{\includegraphics[width=\spyfigwidth,height=\spyfigwidth]{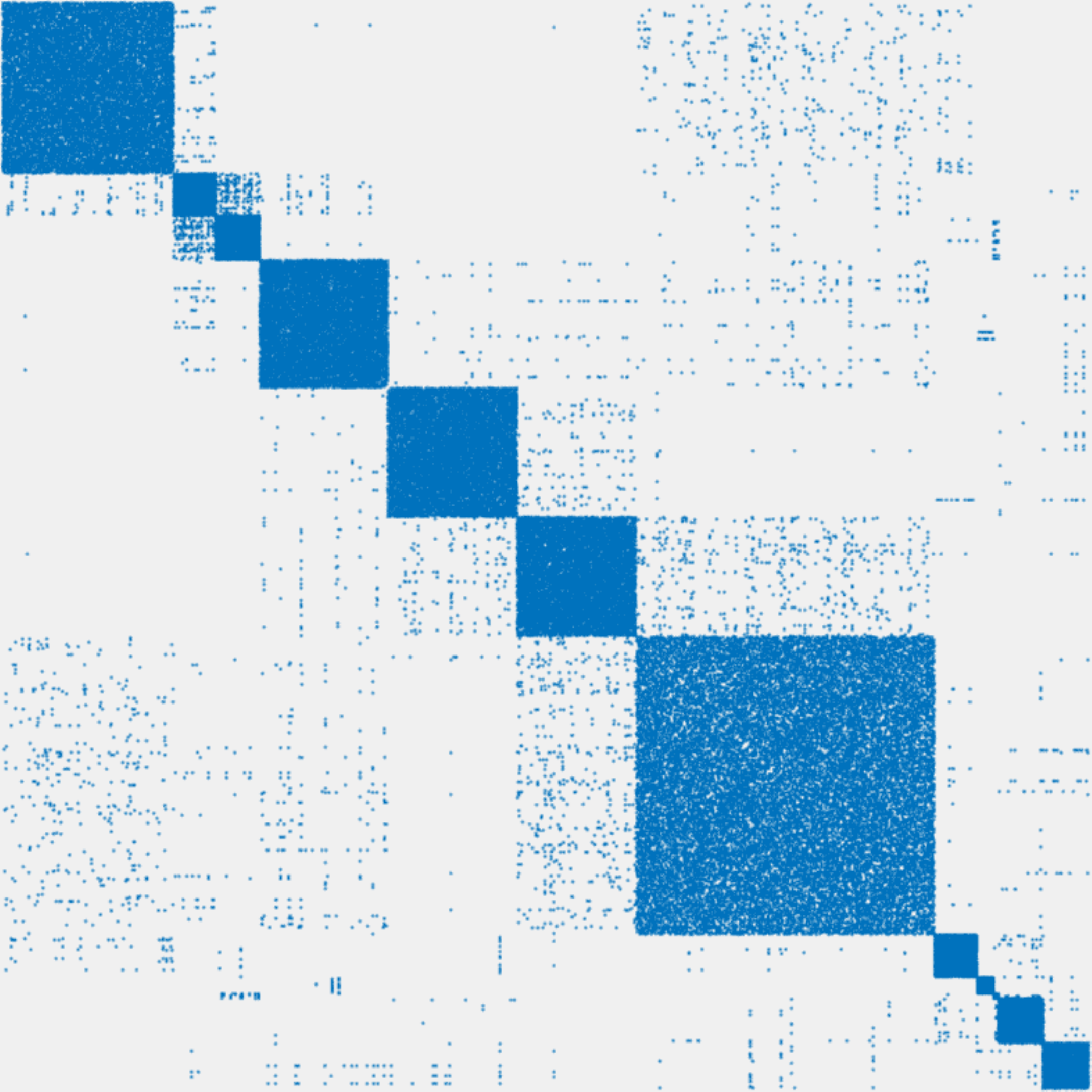}}
		\subcaptionbox{MNIST}
		{\includegraphics[width=\spyfigwidth,height=\spyfigwidth]{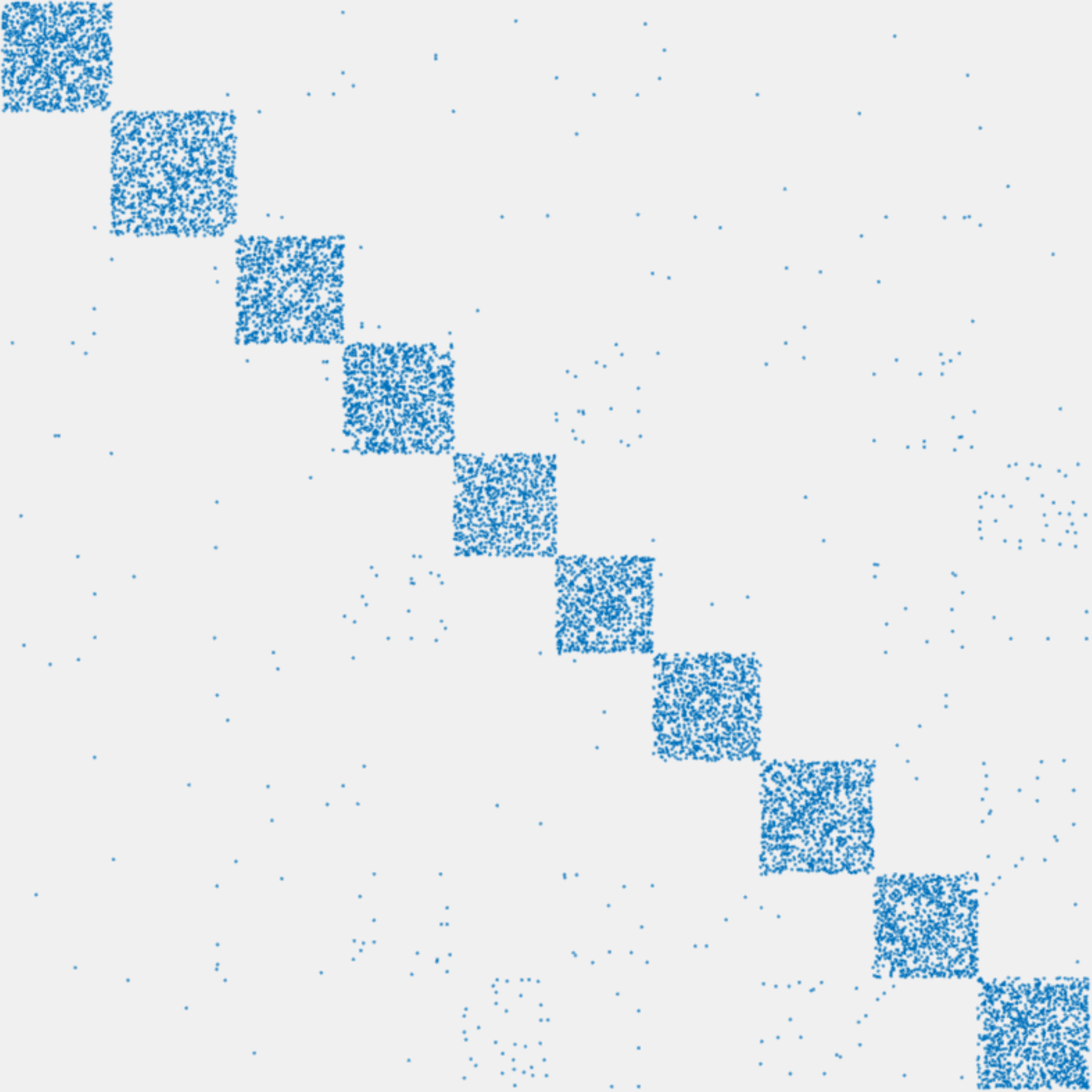}}
		\subcaptionbox{IJCNN}
		{\includegraphics[width=\spyfigwidth,height=\spyfigwidth]{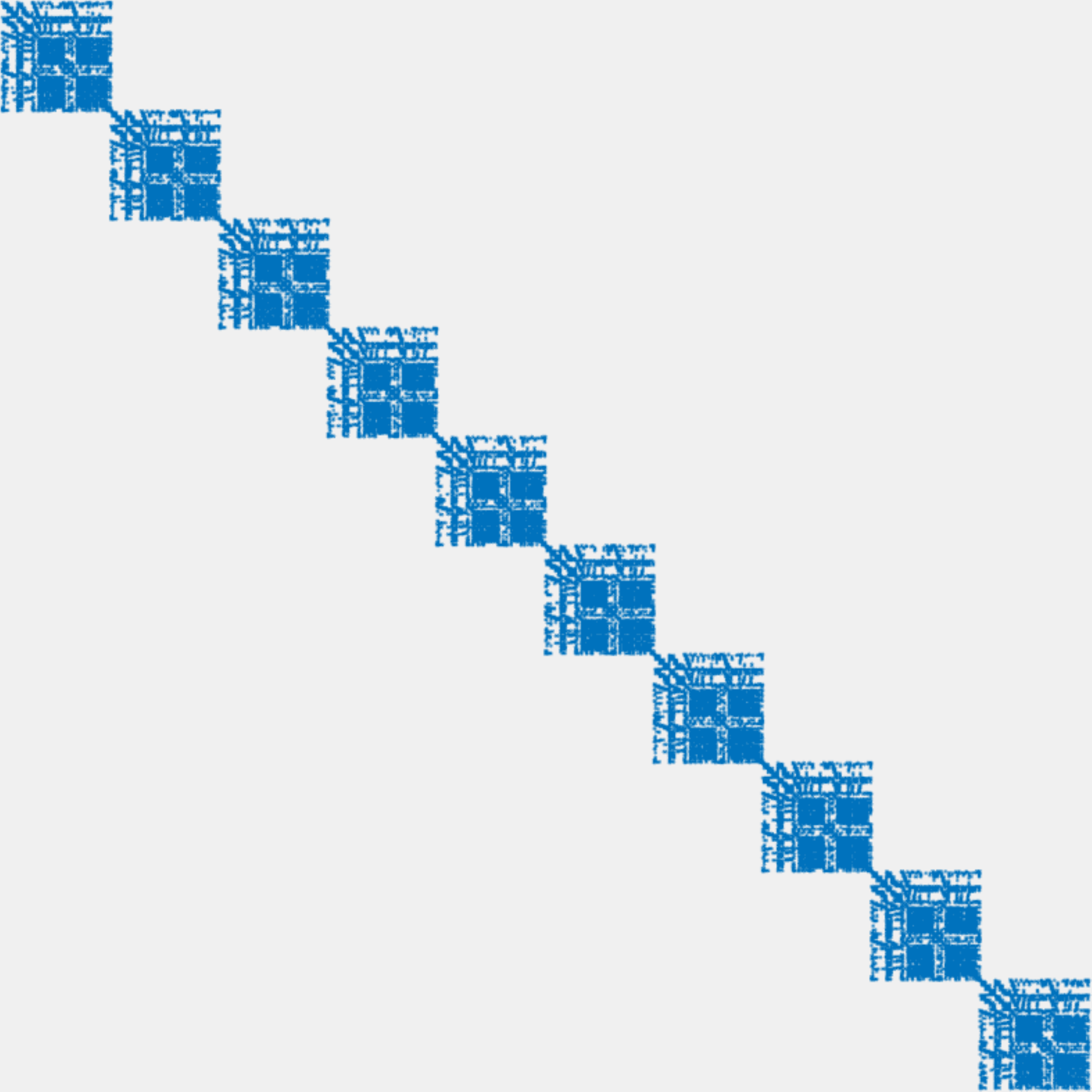}}
		
		\vspace{\baselineskip}
		
		\subcaptionbox{TOMORADAR}
		{\includegraphics[width=\spyfigwidth,height=\spyfigwidth]{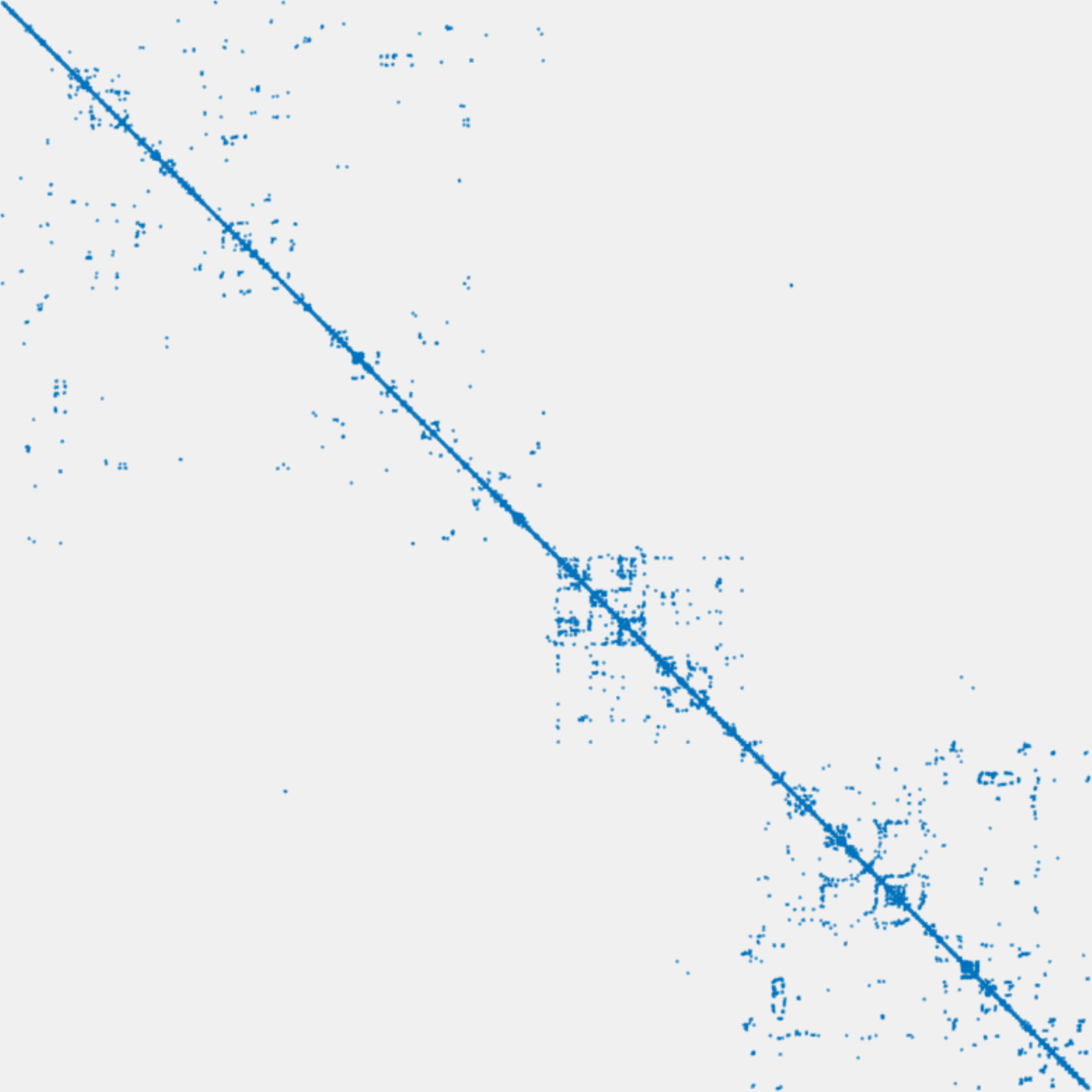}}
		\subcaptionbox{FLOW-CYTOMETRY}
		{\includegraphics[width=\spyfigwidth,height=\spyfigwidth]{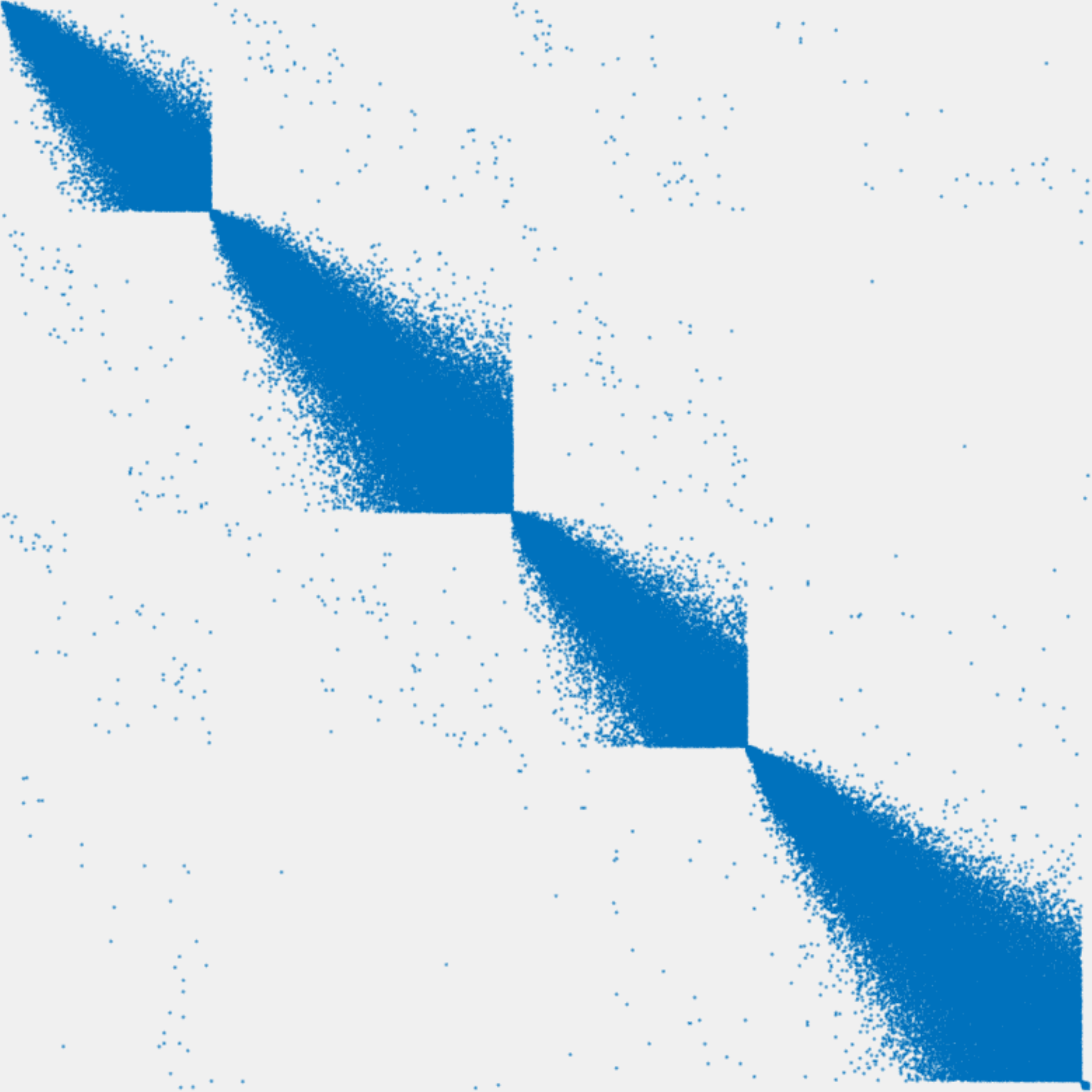}}
		\subcaptionbox{HIGGS}
		{\includegraphics[width=\spyfigwidth,height=\spyfigwidth]{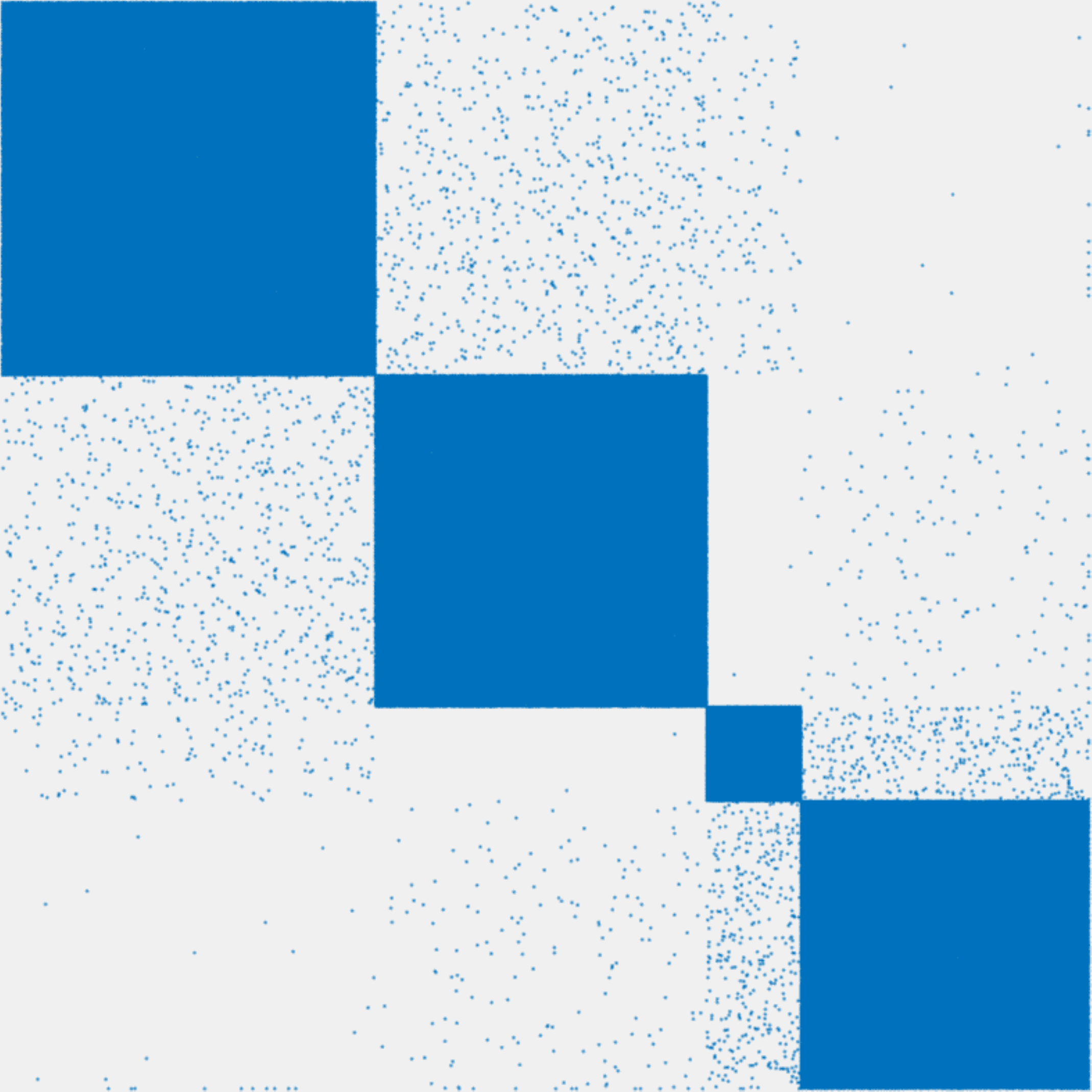}}
	\end{center}
	\caption{Visualization of the similarity matrix $P$ of the experimented data sets using Matlab \texttt{spy} function, where the rows and columns are sorted by the manual cluster labels. Blue dots show the 1's in the matrix and white dots show the 0's. Due to limited resolution, the figures shows a uniform subsample 10\% data points.}
	\label{fig:blockvis}
	\vspace{-3mm}
\end{figure}

\section{Conclusions}
\label{sec:conclusions}
We have presented a new nonlinear dimensionality reduction method called Stochastic Clustering Embedding for better cluster visualization. Our method modifies t-SNE by using an adaptive and effective attraction-repulsion tradeoff. We have tested our method in various real-world data sets and compared it with other modern NLDR methods. The experimental results show that our method can consistently identify the intrinsic clusters. Furthermore, we have contributed a simple and fast optimization algorithm that can easily be implemented in modern parallel computing platforms.

In this work, we have only considered the layout algorithms which produce the embedding coordinates. The visualization quality is also determined by the visual elements such as dot sizes, colors, opacity in the display. One promising area for further research would be to incorporate a cognitive user model, which could potentially improve cluster visualization to a significant degree. Such models could be fitted with Approximate Bayesian Computation as shown by e.g., \citet{kangasraasio2017chi,micallef2017towards} using the efficient implementation of inference algorithms available in the ELFI Python package \citep{Lintusaari2018}.

\clearpage

\appendixtitleon
\begin{appendices}

\section{Optimization equivalence}
\label{sec:optequiv}
The optimization equivalence connection was originally proposed between $\alpha$- and R\'enyi-divergences, and between $\beta$- and $\gamma$-divergences \citep{yang2014optequiv}. Here we make the connection explicit between non-normalized and normalized KL-divergences.
\begin{proof}
	The minimum of $D_\text{I}(P||sq)$ over $s$ appears when $\displaystyle\fracpartial{D_\text{I}(P||sq)}{s}=0$. That is,\newline
	$\displaystyle-s^{-1}\sum_{ij}P_{ij}+\sum_{ij}q_{ij}=0$
	or
	\begin{align}
	s=\frac{1}{\sum_{ij}q_{ij}}.
	\end{align}
	Obviously such $s>0$. By putting it back to $D_\text{I}(P||sq)$, we obtain
	\begin{align}
	D_\text{I}(P||sq)=&
	\sum_{ij}\left[P_{ij}\ln\frac{P_{ij}}{\displaystyle\left(\sum_{ab}q_{ab}\right)^{-1}q_{ij}}
	-P_{ij}+\frac{q_{ij}}{\displaystyle\sum_{ab}q_{ab}}\right]\\
	=&\sum_{ij}P_{ij}\ln\frac{P_{ij}}{Q_{ij}},
	\end{align}
	which is recognized as $D_\text{KL}(P||Q)$.
\end{proof}

\section{User study of GSNE}
\label{sec:ui}
We conducted a user study in both a controlled laboratory and on the crowdsourcing platform CrowdFlower (respectively 40 and 300 participants per data set) to identify a value for $s$ that human users prefer for cluster visualization with GSNE. Thus, in total, we empirically collected 340 answers for each of the 4 data sets.

To be eligible for participation in the user study, the user should have at least a high-school education and be somewhat familiar with scatter plots. In addition, all participants should report that they have not participated in the user study before. These controls are designed to ensure the validity of the collected data.

Each test trial begins with an orientation phase. The user first sees several examples of good and bad cluster visualizations as well as their explanations. Next, the user practices on a few visualizations for synthetic and real-world data sets. We also used these practice questions to screen out robot visits and careless answer clicks \citep{Borgo2017Crowdsourcing}.

\begin{figure*}[p]
	\begin{center}
		\includegraphics[width=16cm]{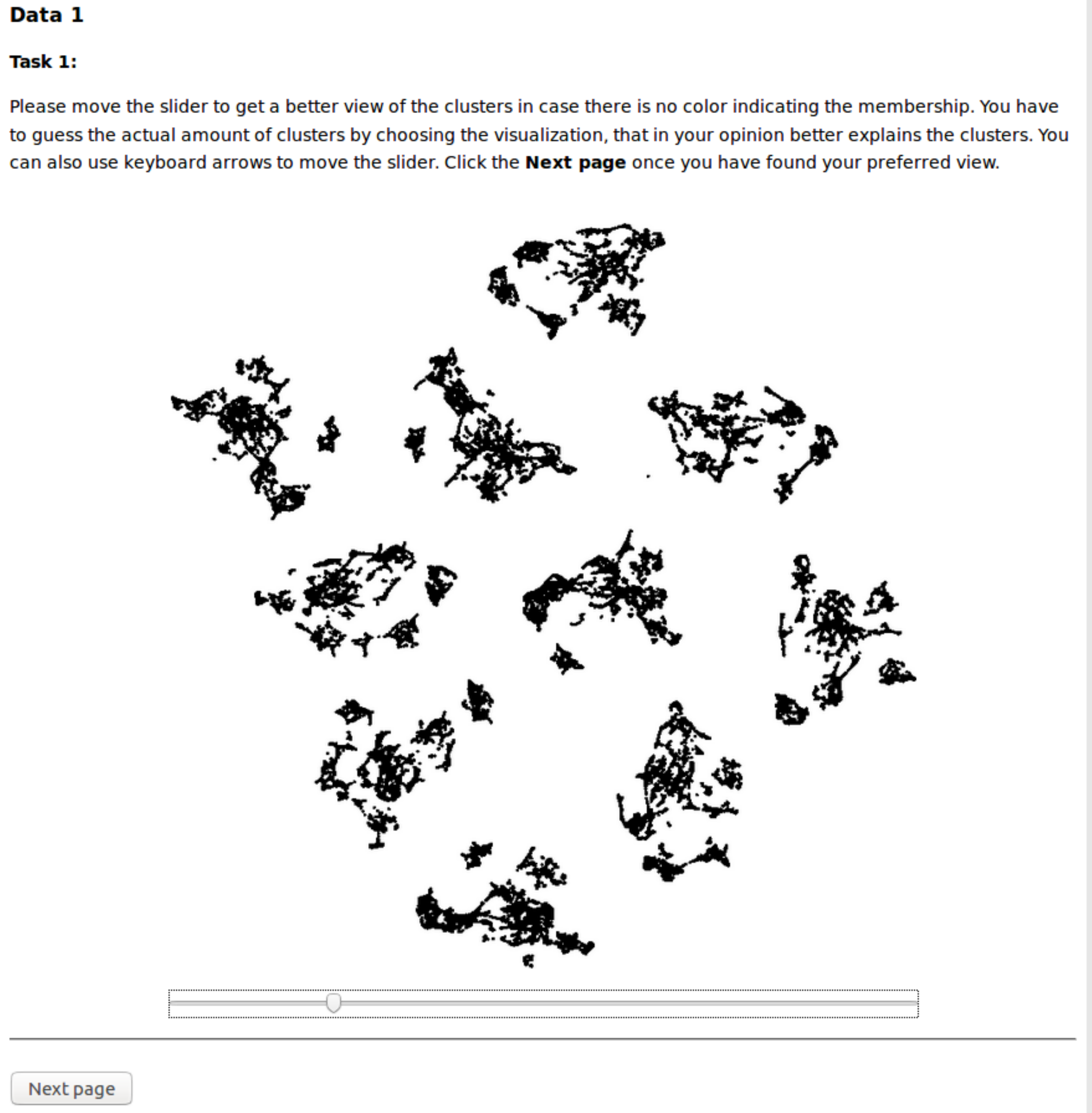}
	\end{center}
	\caption{The user interface of the user study.}
	\label{fig:ui}
\end{figure*}

Then the actual user study starts. The tasks are presented to the user in random order, one at a time. Figure \ref{fig:ui} shows a screenshot of the user interface. At the top is the task description. In the middle is the GSNE visualization for user inspection. The user can move the slider below the visualization to specify the $s$ value. Once $s$ is changed, the program will show the corresponding GSNE visualization in the middle display. We did not use colors to show ground truth classes because cluster visualization is unsupervised. When the user decides the most preferred cluster visualization, he or she clicks the ``Next page'' button at the bottom and proceeds to the next task. At the same time, the system records the user's choice. We removed answers from inattentive participants who did not move the slider for all tasks. 

\section{Comparison with over-exaggerated t-SNE}
\label{sec:vsalllying}
The ``early-exaggeration'' trick in the original t-SNE optimization algorithm also amplifies the attraction force to achieve more compact visualizations. We have tested whether using the trick throughout all iterations can give good cluster visualization for the tested data sets. We call this over-exaggerated t-SNE (with $\beta=12$) and compare it with SCE. The resulting visualizations are shown in Figures \ref{fig:comparedalllying1} and \ref{fig:comparedalllying2}.  

We can see that using over-exaggeration only occasionally yields some clusters, which is inferior to the SCE visualizations on the right. For \texttt{SHUTTLE} and \texttt{IJCNN}, over-exaggerated t-SNE produces some clusters but also shatters others. For \texttt{TOMORADAR}, the shattering is more severe and thus badly fails the over-exaggerated t-SNE for cluster visualization. For \texttt{MNIST}, over-exaggerated t-SNE barely produces ten clusters, though many data points are scattered between and around the clumps.

\renewcommand{\comparedfigwidth}{7cm}
\begin{figure*}[p]
	\begin{center}
		\begin{tabular}{cc}
			\texttt{SHUTTLE}, over-exaggerated t-SNE & \texttt{SHUTTLE}, SCE \\
			\includegraphics[width=\comparedfigwidth]{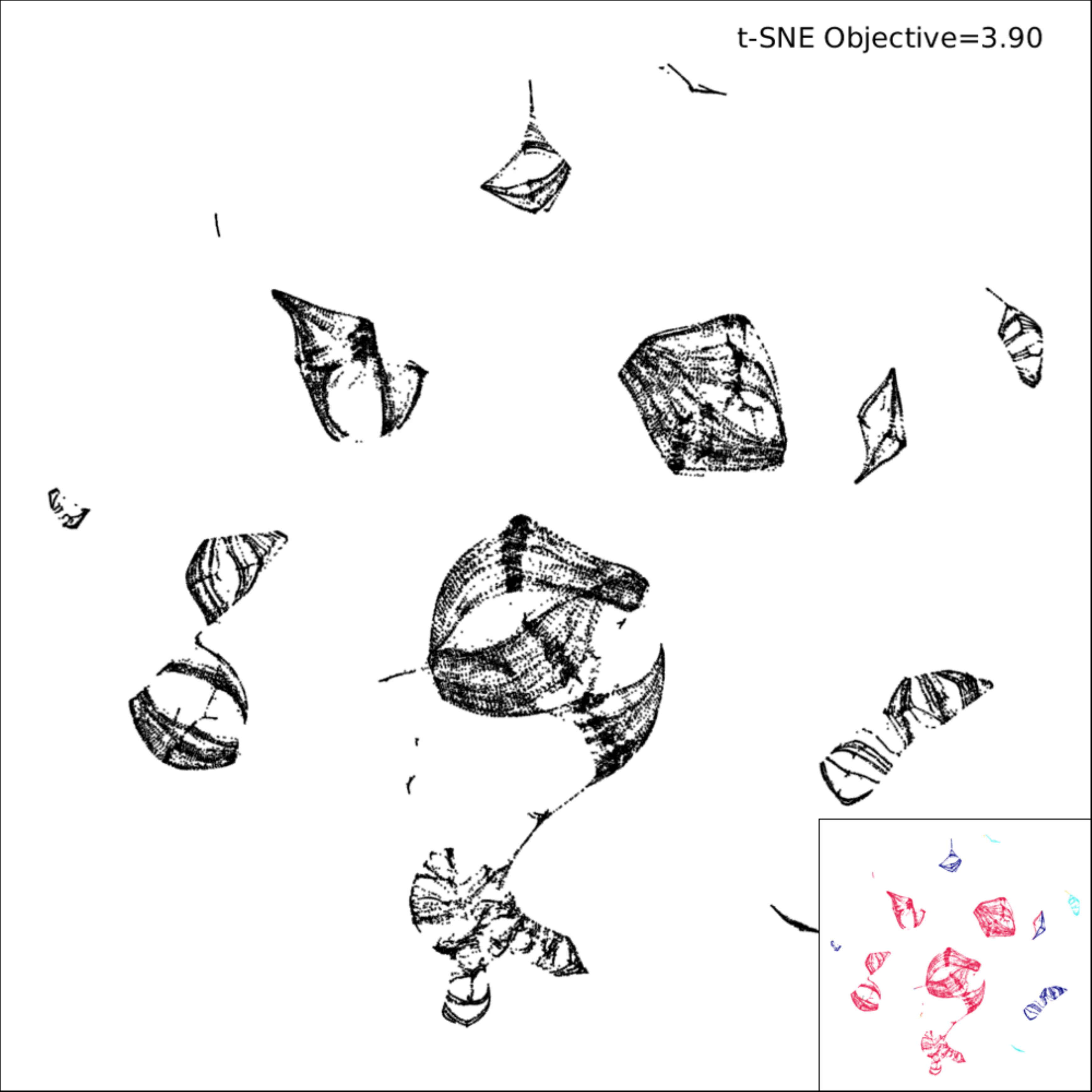} &
			\includegraphics[width=\comparedfigwidth]{figures/shuttle_sce.pdf}\\
			\texttt{MNIST}, over-exaggerated t-SNE & \texttt{MNIST}, SCE \\
			\includegraphics[width=\comparedfigwidth]{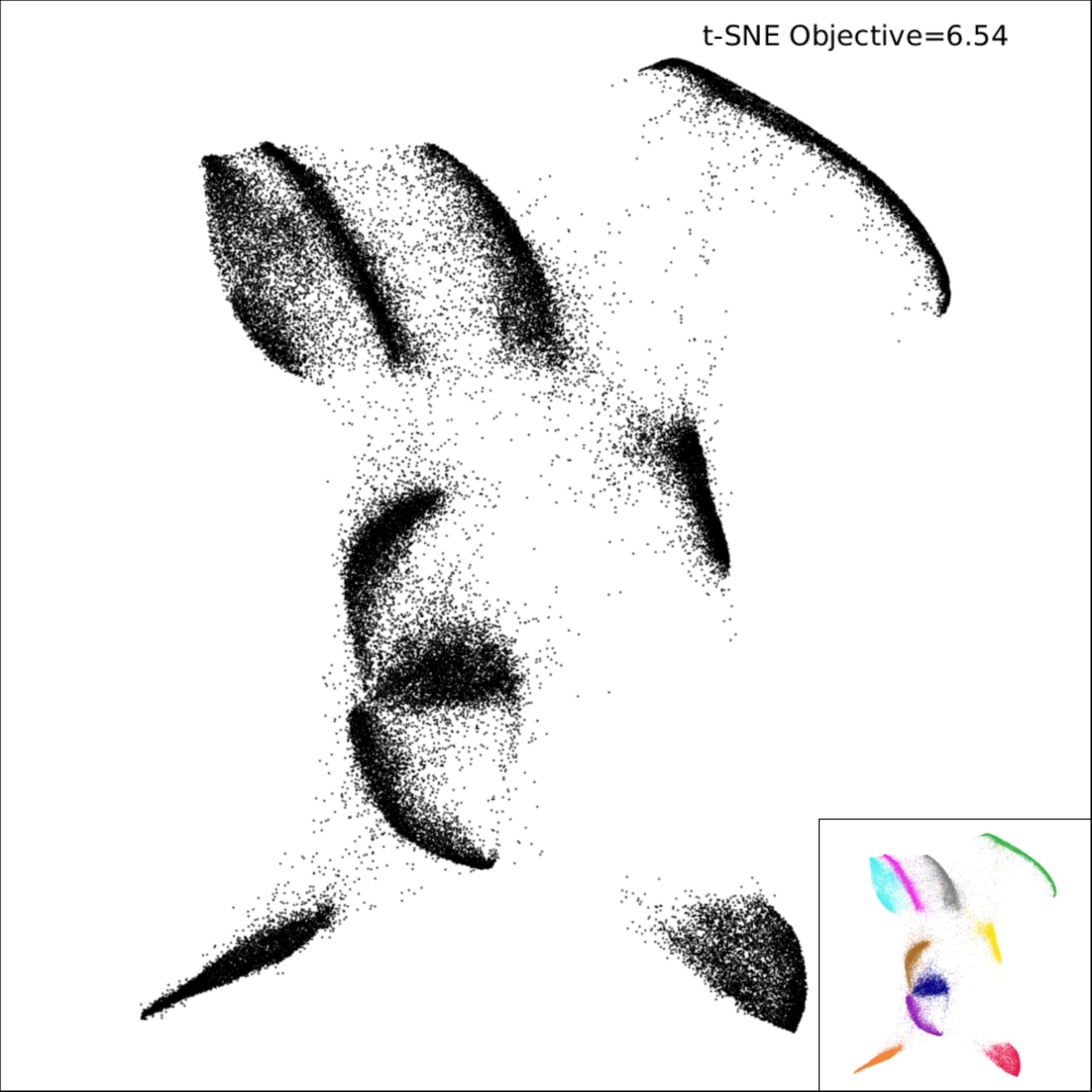} &
			\includegraphics[width=\comparedfigwidth]{figures/mnist_sce.pdf}
		\end{tabular}
	\end{center}
	\caption{Visualizations of \texttt{SHUTTLE} and \texttt{MNIST} data sets by using over-exaggerated t-SNE and SCE. The classes are shown by colors in the small sub-figures.}
	\label{fig:comparedalllying1}
\end{figure*}

\begin{figure*}[p]
	\begin{center}
		\begin{tabular}{cc}
			\texttt{IJCNN}, over-exaggerated t-SNE & \texttt{IJCNN}, SCE \\
			\includegraphics[width=\comparedfigwidth]{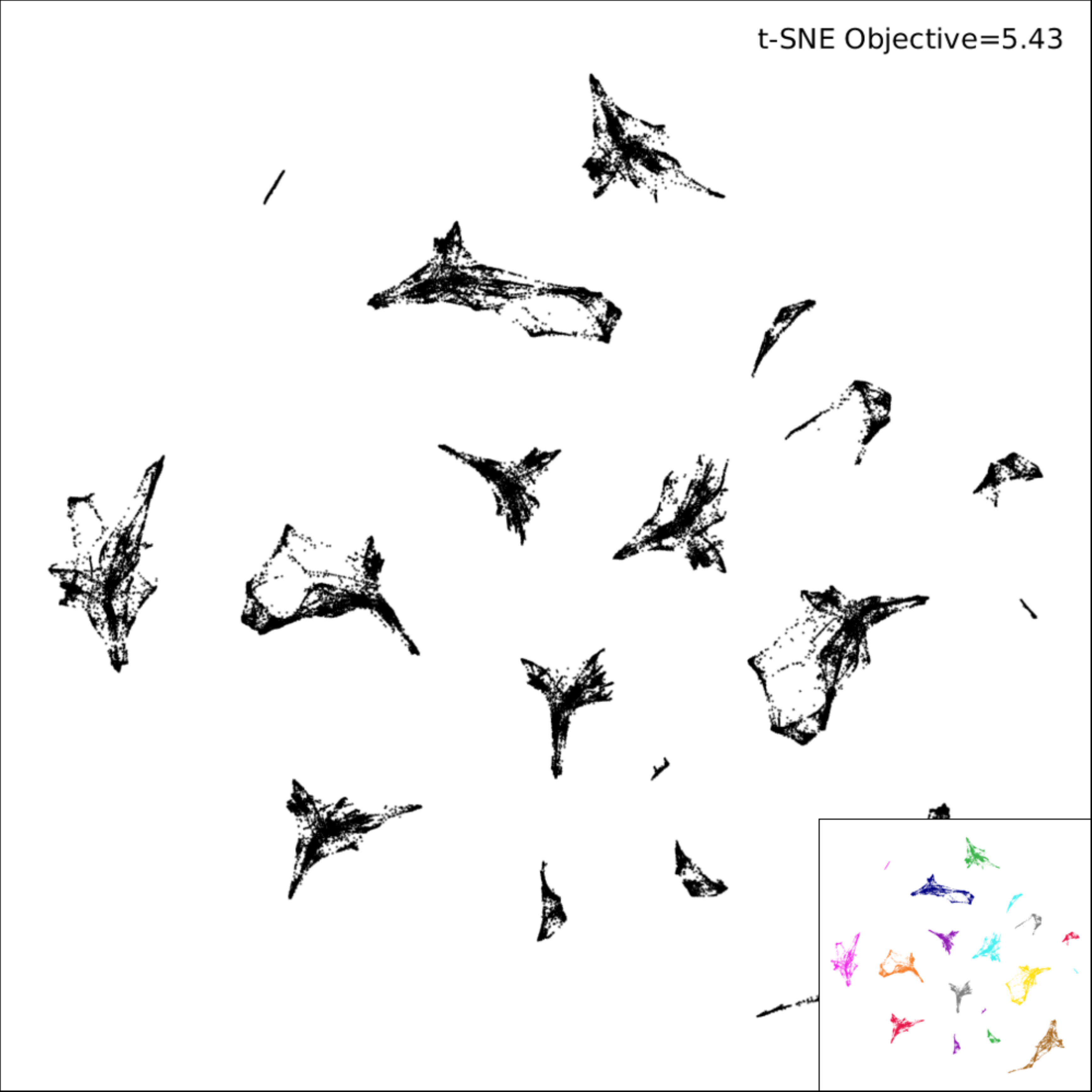} &
			\includegraphics[width=\comparedfigwidth]{figures/ijcnn_sce.pdf}\\
			\texttt{TOMORADAR}, over-exaggerated t-SNE & \texttt{TOMORADAR}, SCE \\
			\includegraphics[width=\comparedfigwidth]{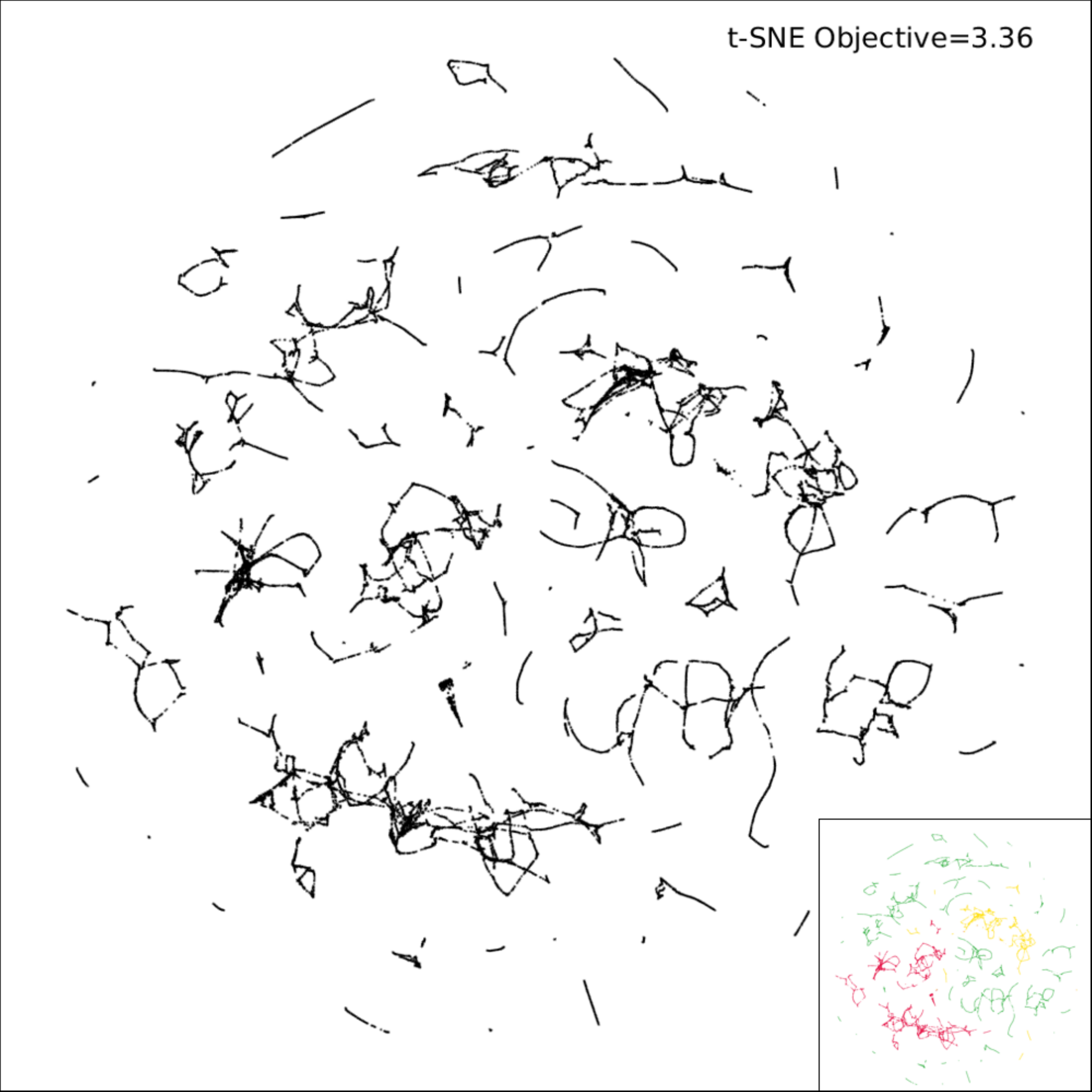} &
			\includegraphics[width=\comparedfigwidth]{figures/tomoradar_sce.pdf}
		\end{tabular}
	\end{center}
	\caption{Visualizations of \texttt{IJCNN} and \texttt{TOMORADAR} data sets by using over-exaggerated t-SNE and SCE. The classes are shown by colors in the small sub-figures.}
	\label{fig:comparedalllying2}
\end{figure*}

\section{Comparison with opt-SNE}
\label{sec:vsoptsne}
Here we provide a comparison with a recent method called optimized t-SNE \citep[opt-SNE;][]{optsne} which also adjusts the attraction-repulsion tradeoff. It automates the selection of three important parameters for the t-SNE run: 1) initial learning rate, 2) the number of iterations spent in early exaggeration, and 3) the number of total iterations. Despite the changes in initialization, opt-SNE still minimizes the original t-SNE objective, i.e.~minimum normalized KL-divergence (KLD) at the end, because the opt-SNE authors believed that lower KLD values indicate superior visualization quality \citep[][Page 5]{optsne}.

Despite the similarity, our method has a major difference from opt-SNE. Our method SCE does not minimize the same cost function in t-SNE throughout the iterations because we find that a lower KLD can correspond to a worse visualization in terms of showing clusters. 

\begin{figure}[p]
	\begin{center}
		\begin{tabular}{cc}
			\texttt{SHUTTLE} & \texttt{MNIST} \\
			\includegraphics[width=\comparedfigwidth]{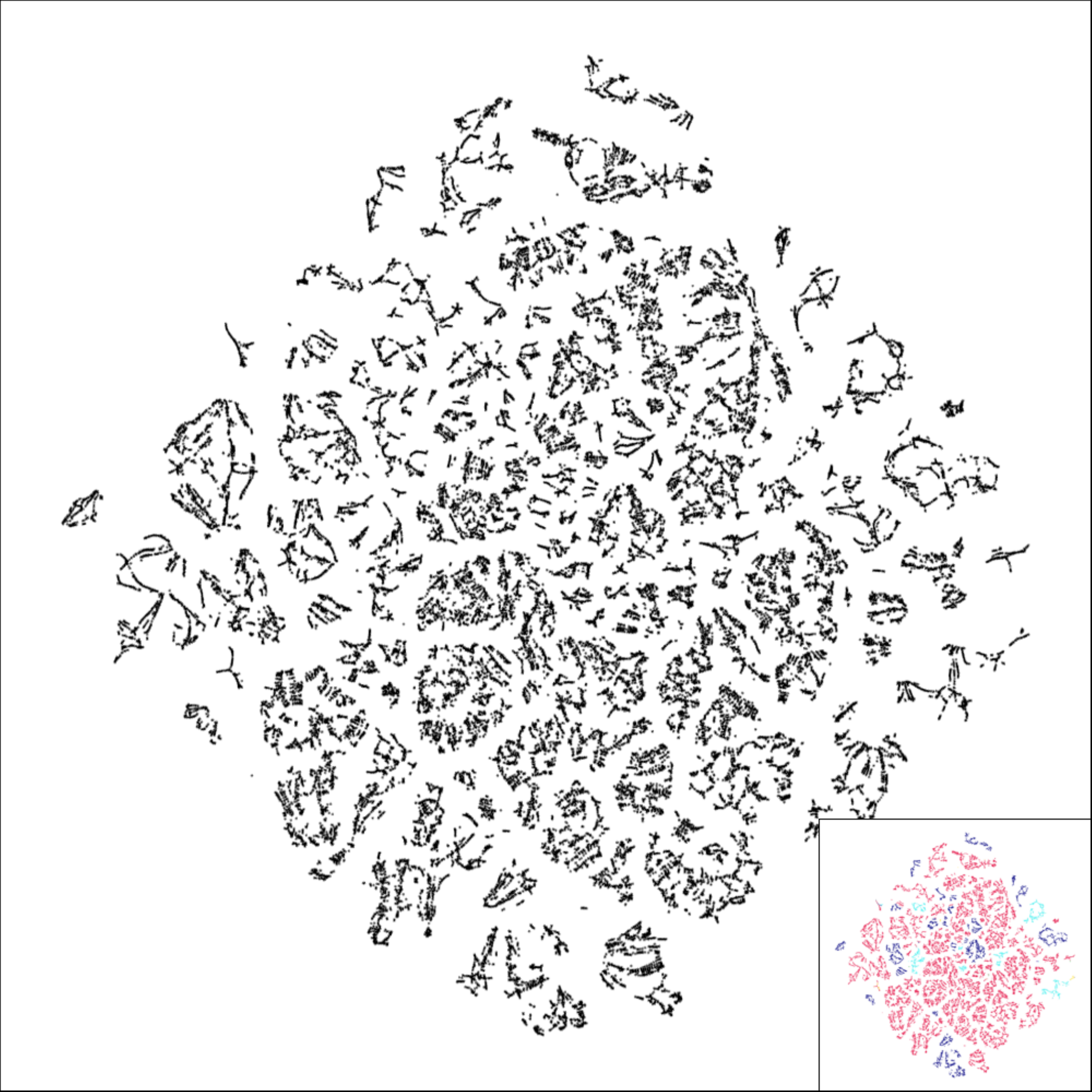} &
			\includegraphics[width=\comparedfigwidth]{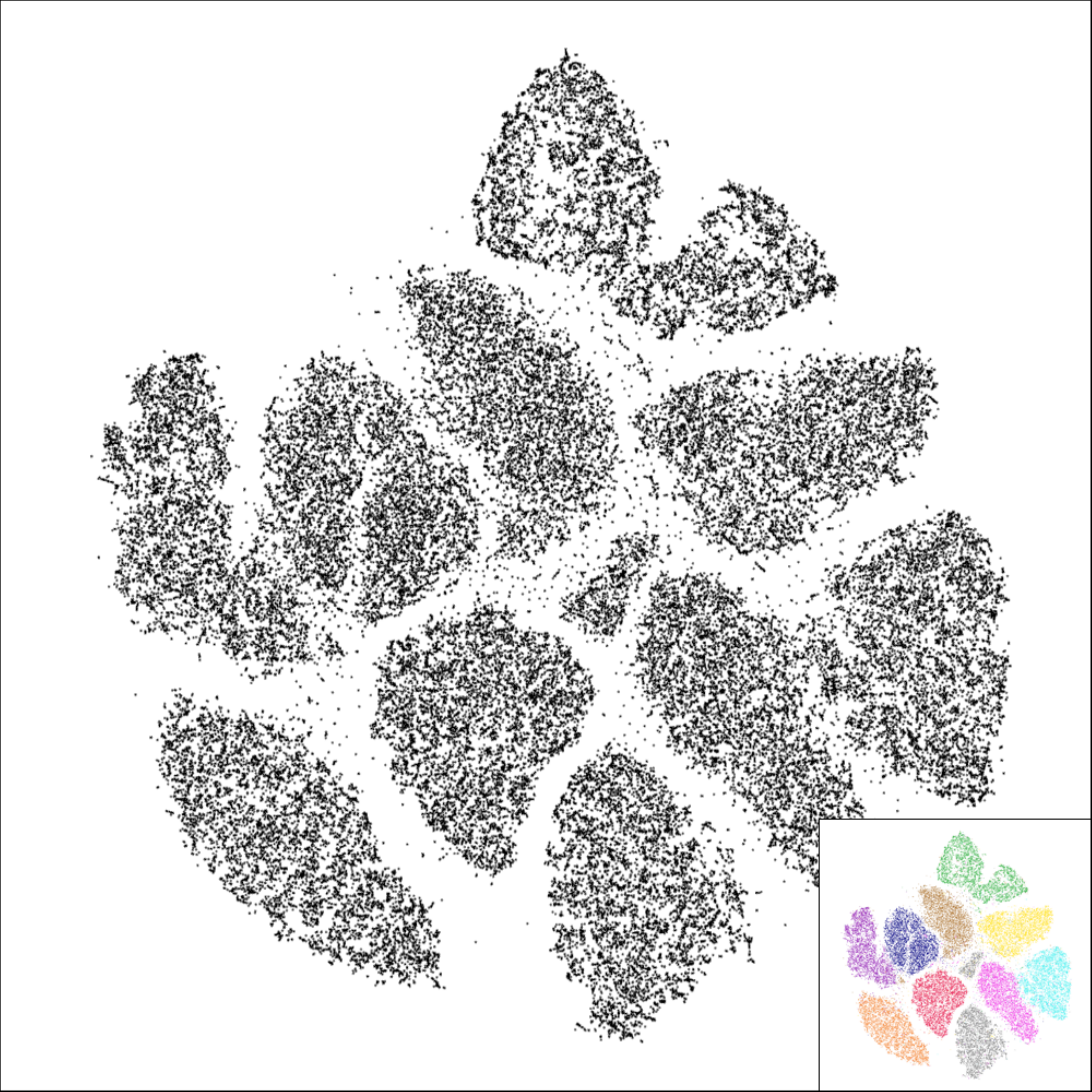}\\
			\texttt{IJCNN} & \texttt{TOMORADAR} \\
			\includegraphics[width=\comparedfigwidth]{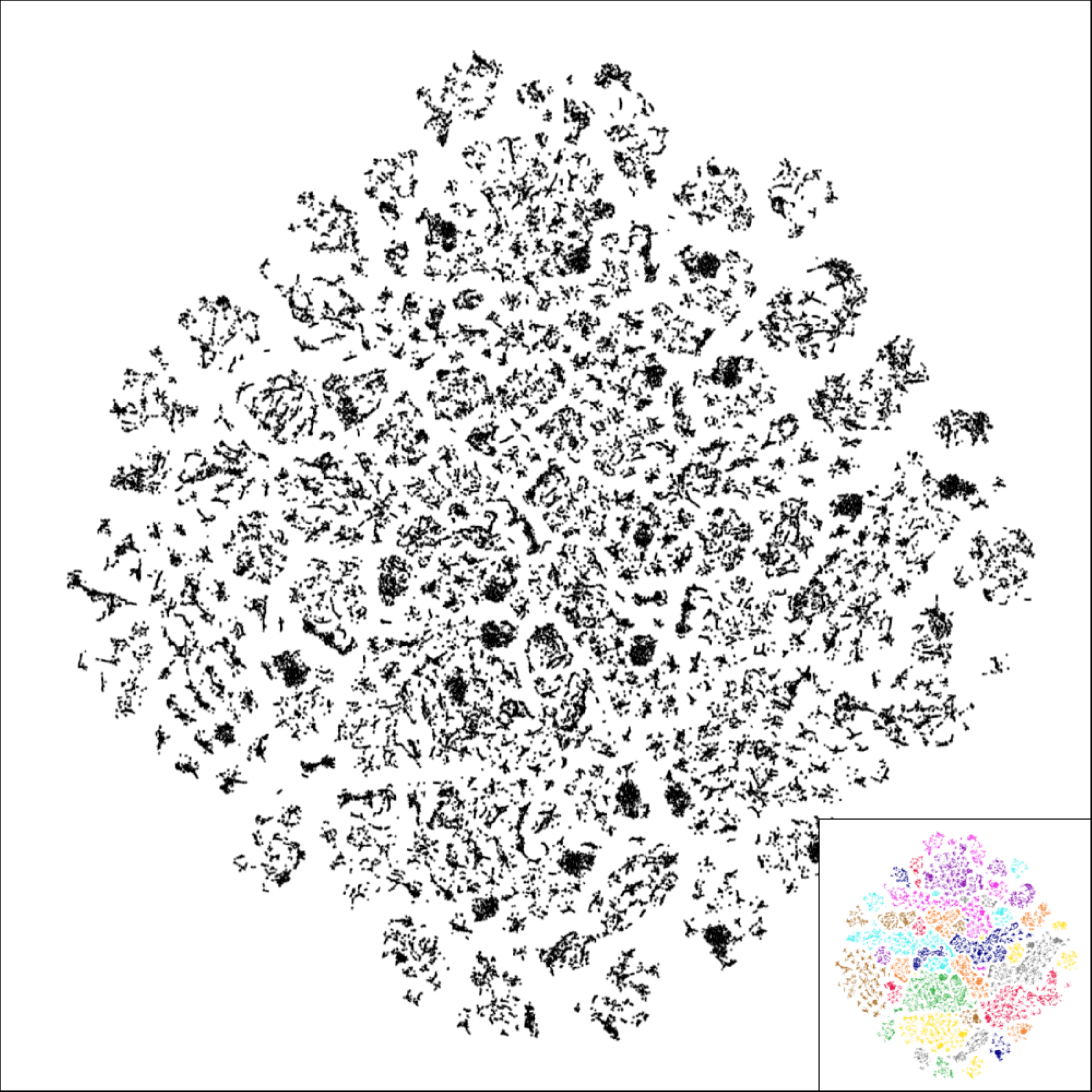} &
			\includegraphics[width=\comparedfigwidth]{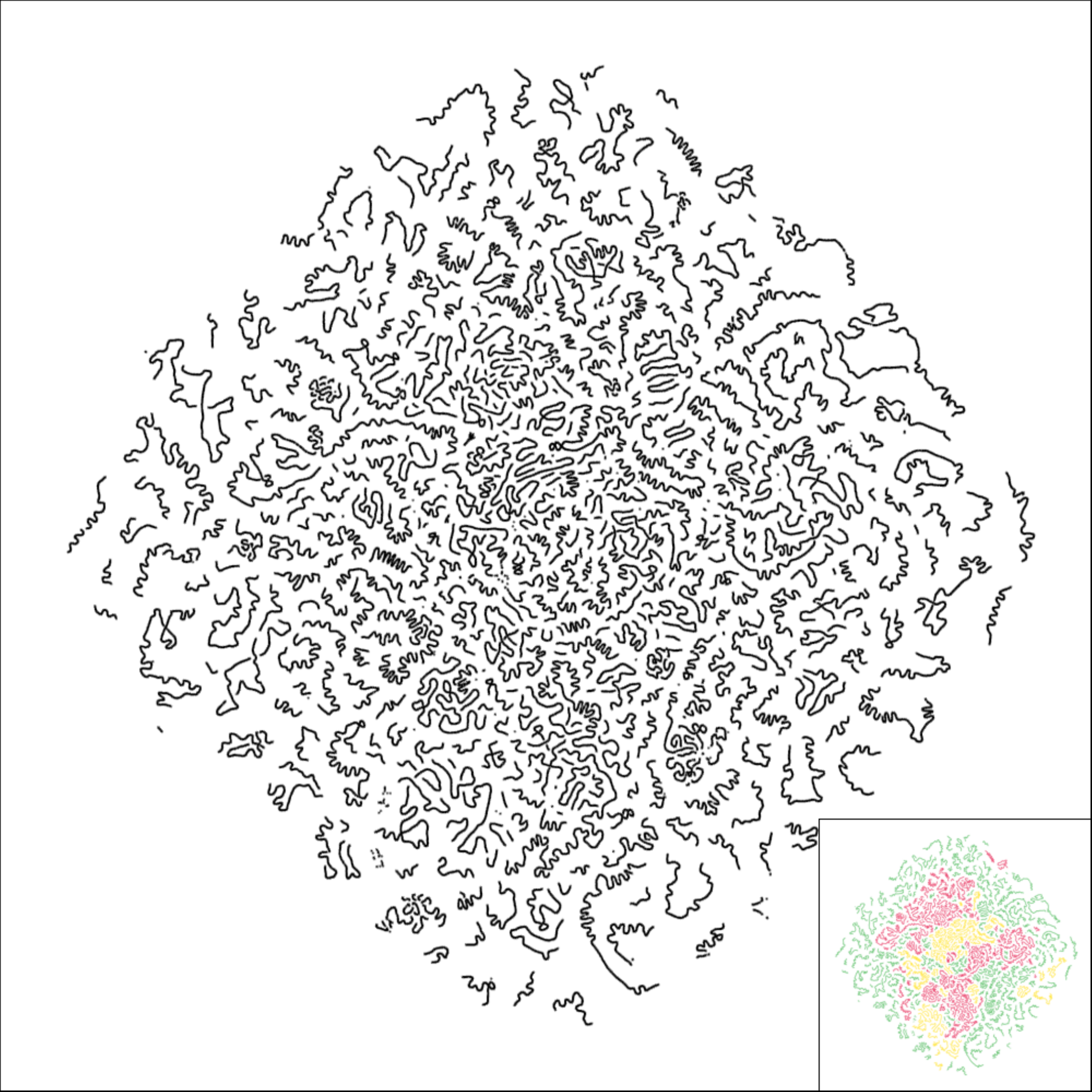}
		\end{tabular}
	\end{center}
	\caption{Visualizations of the \texttt{SHUTTLE}, \texttt{MNIST}, \texttt{IJCNN}, and \texttt{TOMORADAR} data sets by using opt-SNE. The input to opt-SNE are the original vectorial data. The classes are shown by colors in the small sub-figures.}
	\label{fig:optsne}
\end{figure}

\begin{figure}[p]
	\begin{center}
		\includegraphics[width=16cm]{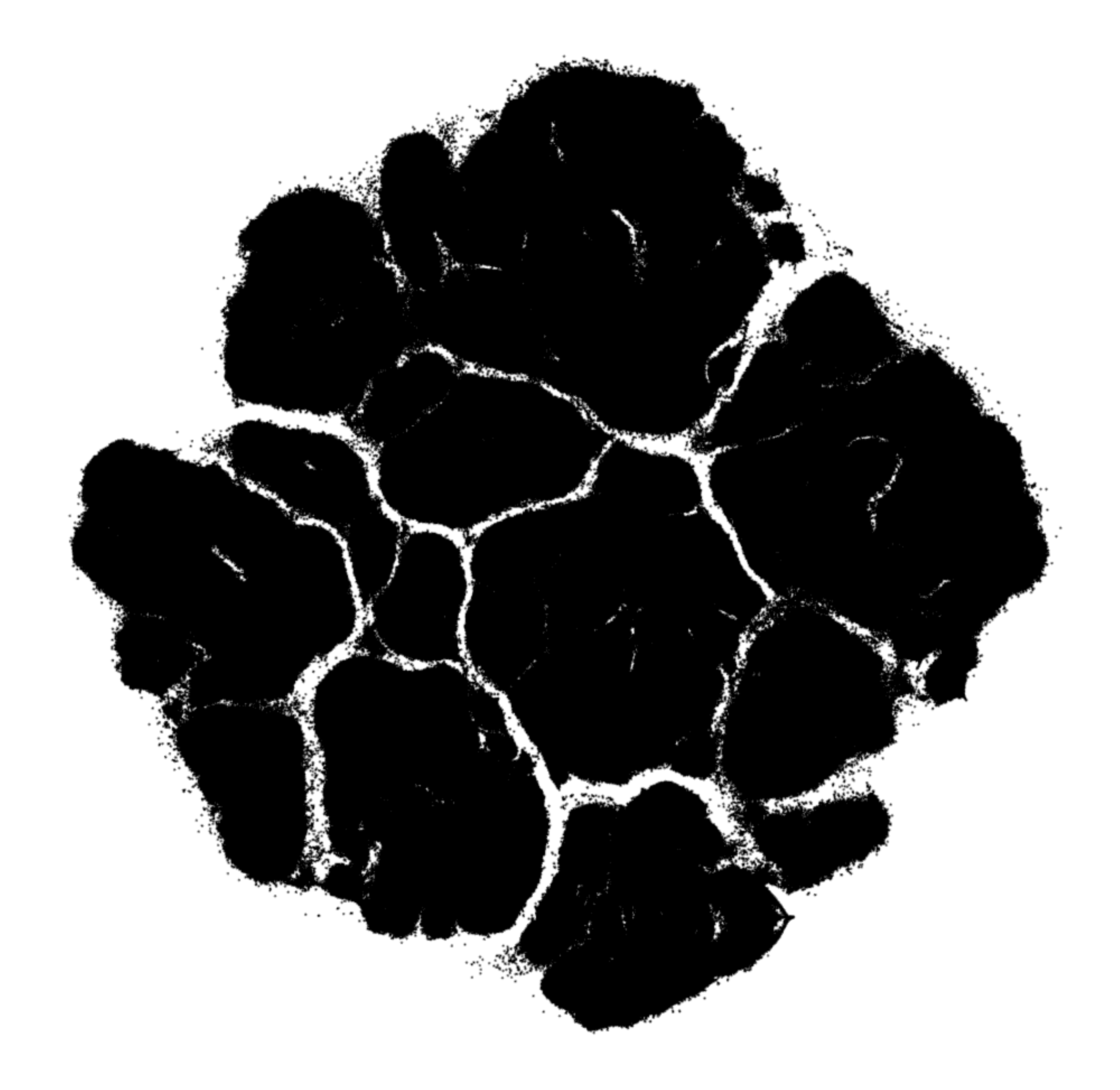}
	\end{center}
	\caption{Visualizations of the \texttt{HIGGS} data set by using opt-SNE.}
	\label{fig:optsnehiggs}
\end{figure}

\begin{figure}[p]
	\begin{center}
		\begin{tabular}{cc}
			\texttt{SHUTTLE} & \texttt{MNIST} \\
			\includegraphics[width=\comparedfigwidth]{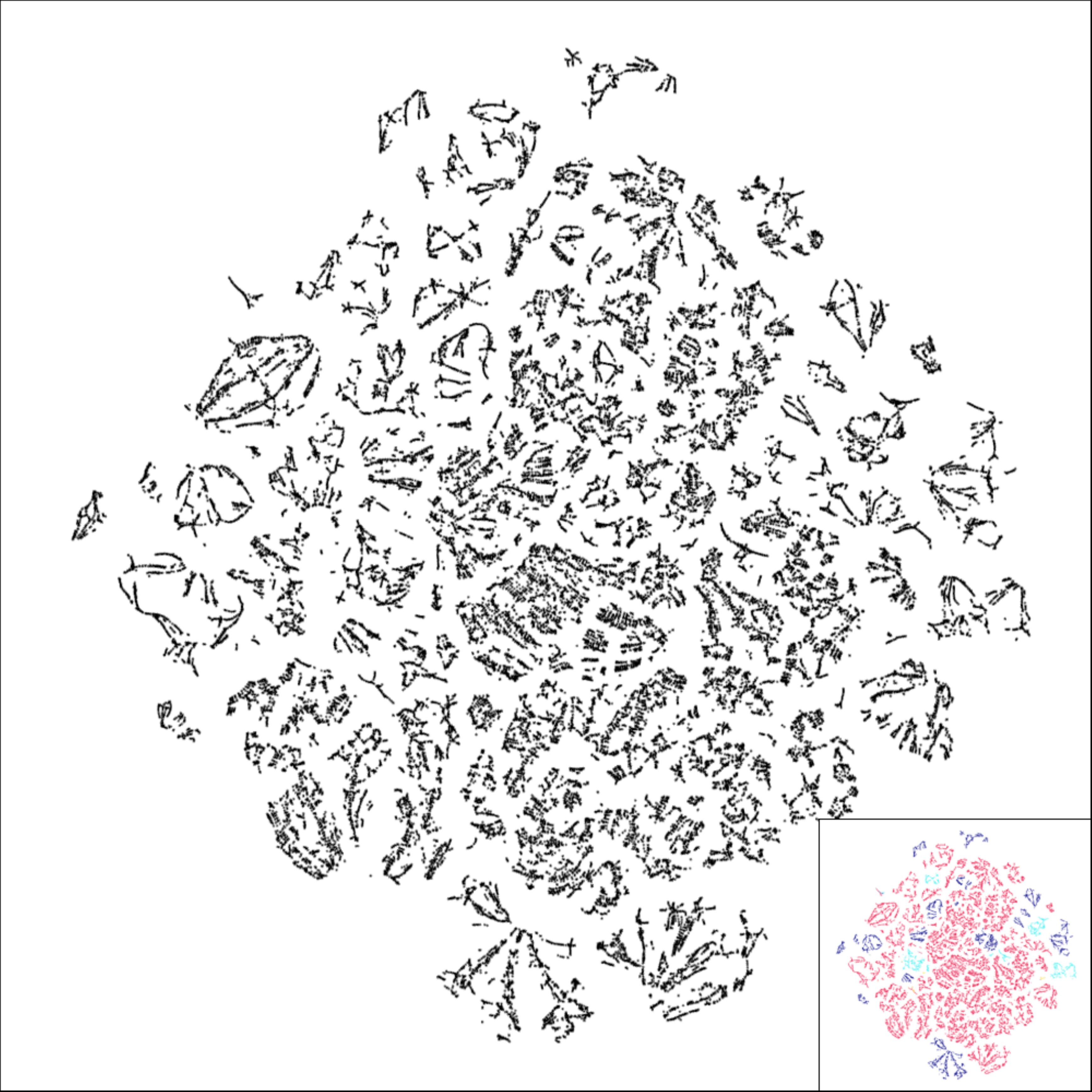} &
			\includegraphics[width=\comparedfigwidth]{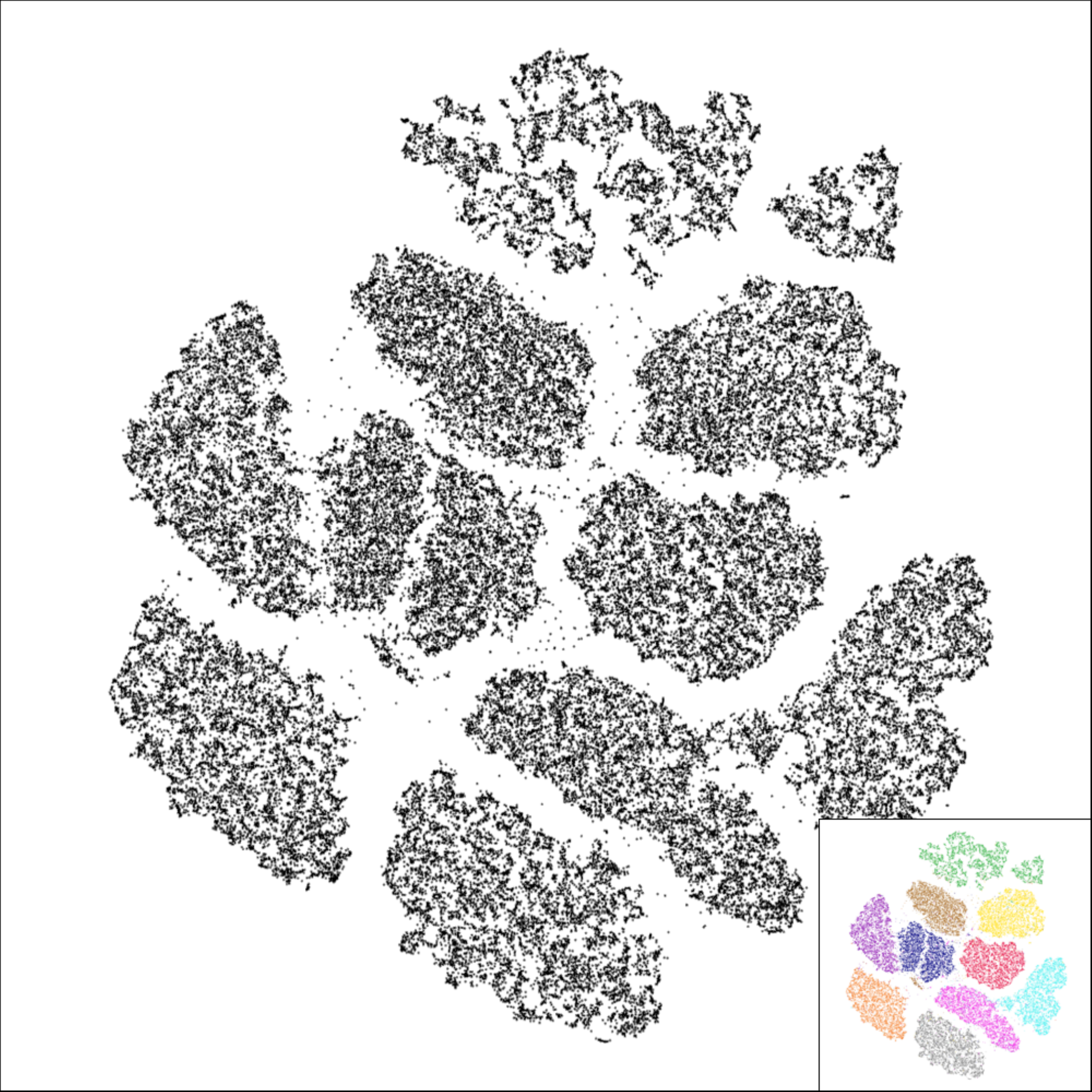}\\
			\texttt{IJCNN} & \texttt{TOMORADAR} \\
			\includegraphics[width=\comparedfigwidth]{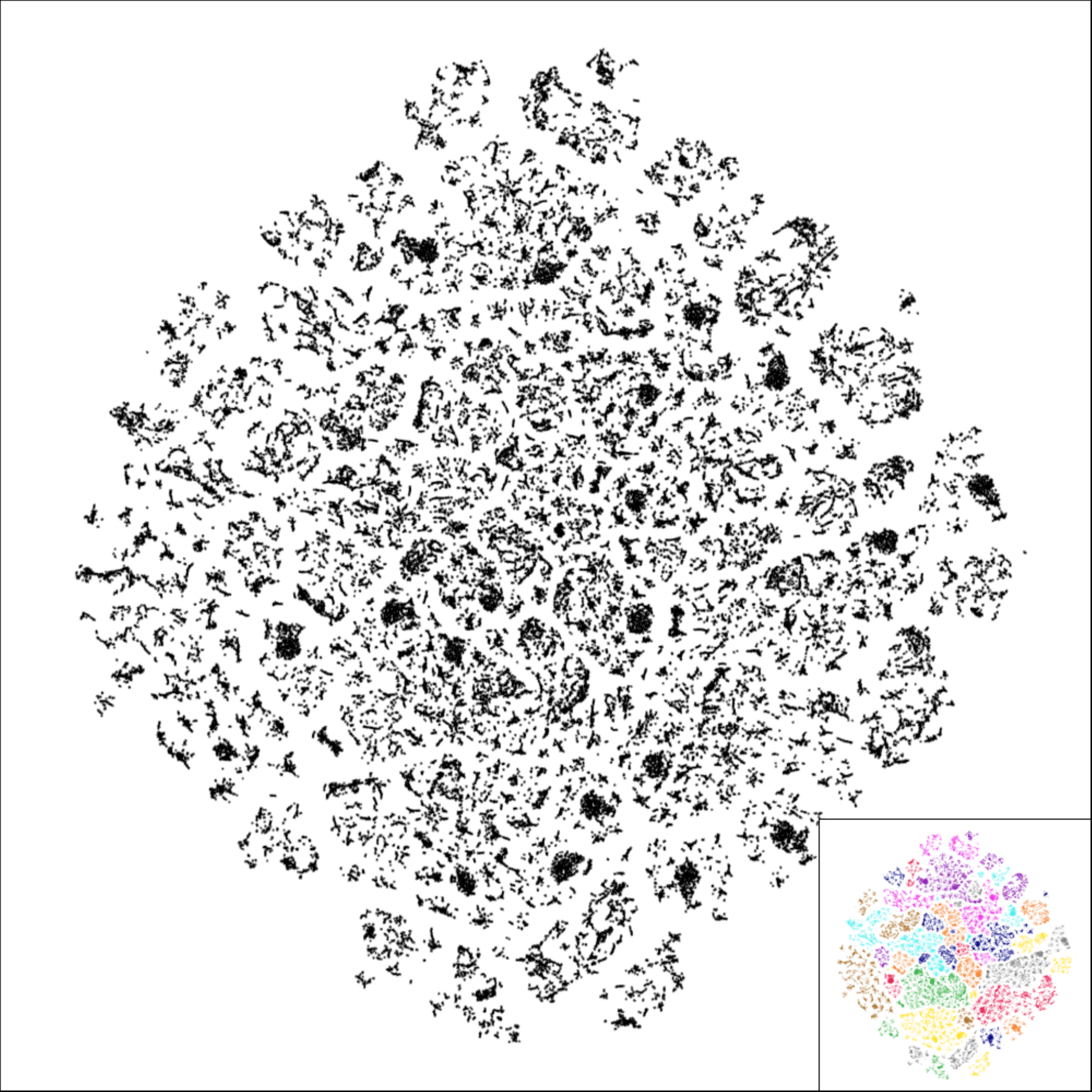} &
			\includegraphics[width=\comparedfigwidth]{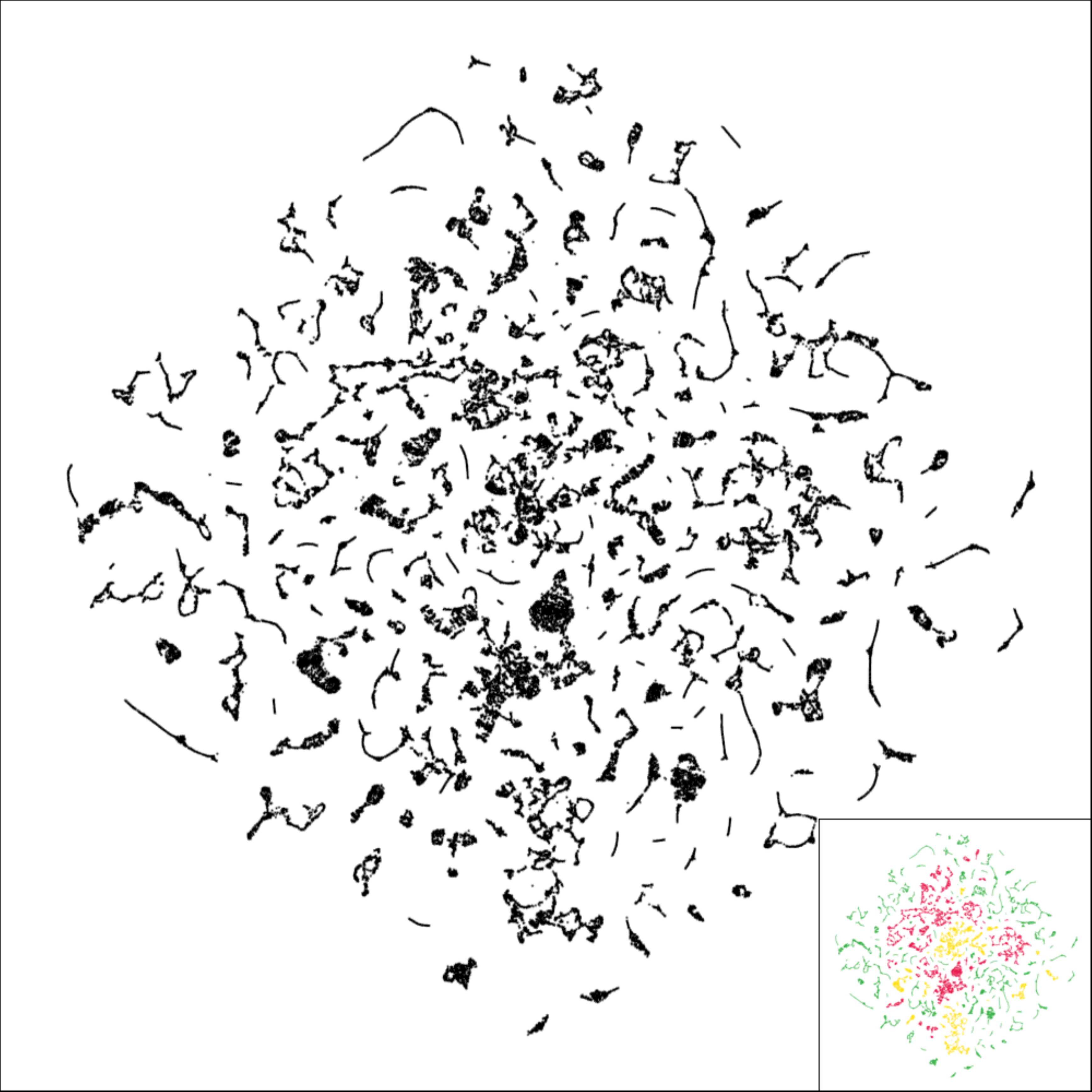}
		\end{tabular}
	\end{center}
	\caption{Visualizations of the \texttt{SHUTTLE}, \texttt{MNIST}, \texttt{IJCNN}, and \texttt{TOMORADAR} data sets by using opt-SNE and the same similarity matrices for SCE.. The classes are shown by colors in the small sub-figures.}
	\label{fig:optsnep}
\end{figure}

\begin{figure}[p]
	\begin{center}
		\includegraphics[width=16cm]{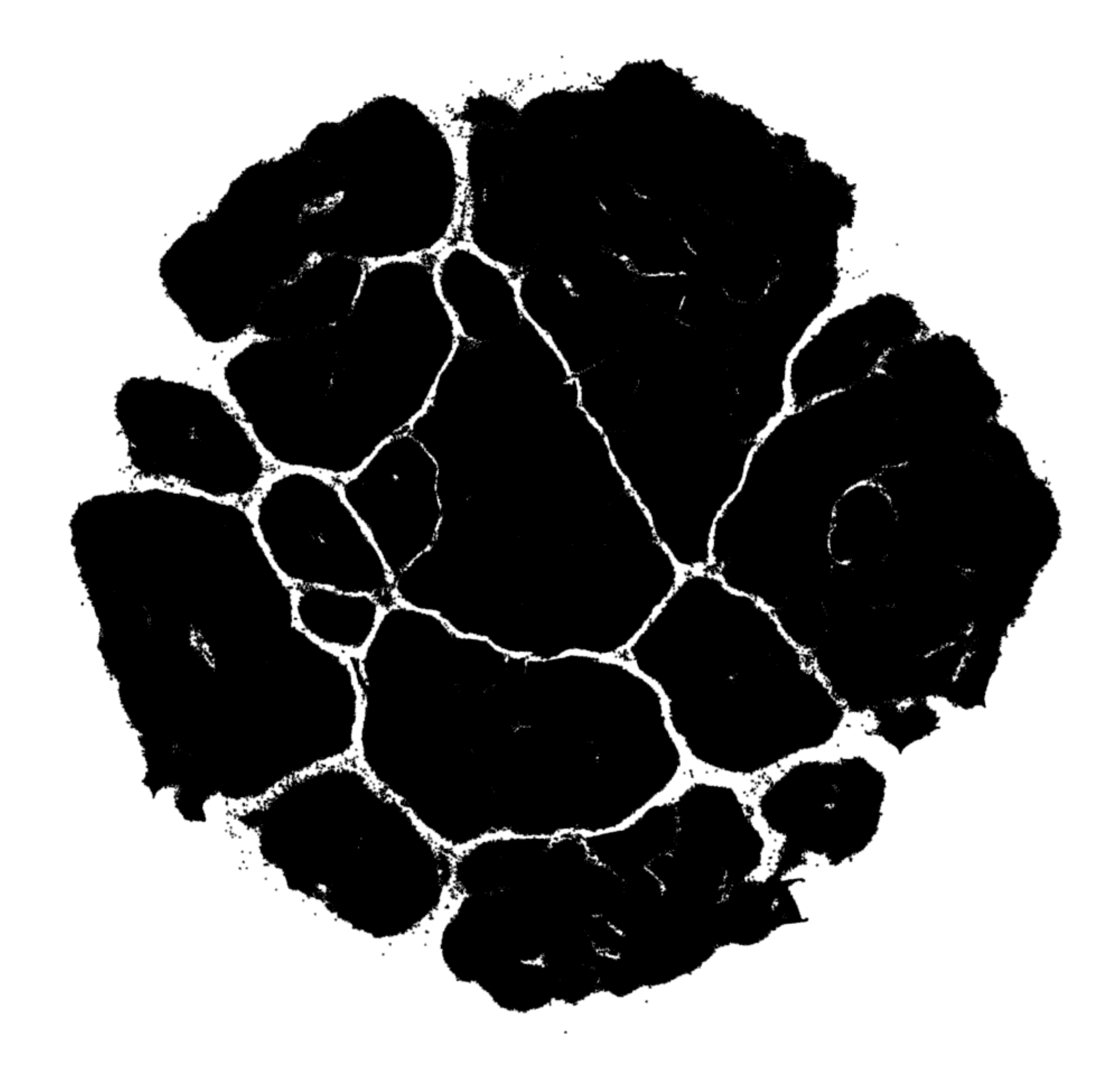}
	\end{center}
	\caption{Visualizations of the \texttt{HIGGS} data set by using opt-SNE and the same similarity matrix for SCE.}
	\label{fig:optsnephiggs}
\end{figure}

We have tried opt-SNE on five tested data sets \texttt{SHUTTLE}, \texttt{IJCNN}, \texttt{MNIST}, \texttt{TOMORADAR} and \texttt{HIGGS}. First we applied opt-SNE on the original vectorial data. That is, opt-SNE will compute the similarity matrix using its built-in function. We used default parameters of opt-SNE (with ``--optsne'' to True).   The resulting visualizations are given in Figures \ref{fig:optsne} and \ref{fig:optsnehiggs}.
Next, for the ablation study, we modified opt-SNE and made it admits the same similarity matrix $P$ we used for SCE and SCE. The resulting visualizations are given in Figures \ref{fig:optsnep} and \ref{fig:optsnephiggs}.

The cluster embedding found by opt-SNE is worse than those by SCE, no matter whether it uses built-in similarity calculation or not. For \texttt{SHUTTLE}, \texttt{IJCNN}, and \texttt{TOMORADAR}, the opt-SNE visualizations show many small pieces and no clear cluster pattern. For \texttt{MNIST} and \texttt{HIGGS}, opt-SNE barely finds the clusters, but the cluster boundaries are unclear compared to the SCE visualizations.

\end{appendices}

\newpage

\bibliographystyle{plainnat}
%\bibliography{sce}

\end{document}